\documentclass[11pt,letterpaper, logo]{mystyle}

\usepackage[utf8]{inputenc} % allow utf-8 input
\usepackage[T1]{fontenc}    % use 8-bit T1 fonts
\usepackage{hyperref}       % hyperlinks
\usepackage{url}            % simple URL typesetting
\usepackage{booktabs}       % professional-quality tables
\usepackage{amsfonts}       % blackboard math symbols
\usepackage{nicefrac}       % compact symbols for 1/2, etc.
\usepackage{microtype}      % microtypography
\usepackage{lipsum}
\usepackage{graphicx}
\usepackage{enumitem}
\usepackage[numbers]{natbib}
\usepackage{fontawesome5}
\usepackage[dvipsnames]{xcolor}
\usepackage{amsmath}
\usepackage{twemojis}
\usepackage{bxcoloremoji}
\usepackage[table,xcdraw]{xcolor}
\usepackage{multirow}

\NewDocumentCommand{\heng}
{ mO{} }{\textcolor{red}{\textsuperscript{\textit{Heng}}\textsf{\textbf{\small[#1]}}}}

\usepackage{booktabs,tabularx}
\newtcbox{\llmchip}{on line, arc=2pt, colback=gray!12, colframe=gray!20,
  boxsep=0.7pt, left=2pt, right=2pt, top=0.2ex, bottom=0.2ex}

\newtcbox{\agentchip}{on line, arc=2pt, colback=blue!8, colframe=blue!25,
  boxsep=0.7pt, left=2pt, right=2pt, top=0.2ex, bottom=0.2ex}

\definecolor{AgentTeal}{HTML}{0AA5A5}
\definecolor{AgentTealLight}{HTML}{D9F3F3}

\definecolor{AgentIndigo}{HTML}{3F51B5}
\definecolor{AgentIndigoLight}{HTML}{E8EAF6}

\definecolor{AgentAmber}{HTML}{FF8F00}
\definecolor{AgentAmberLight}{HTML}{FFF3E0}

\definecolor{AgentViolet}{HTML}{8E24AA}
\definecolor{AgentVioletLight}{HTML}{F3E5F5}

\tcbset{
  agentdef/.style={
    colback=AgentTealLight,
    colframe=AgentTeal,
    colbacktitle=AgentTeal!20!white,
    coltitle=black,
    boxrule=0.9pt, arc=2mm,
    left=3mm, right=3mm, top=2mm, bottom=2mm,
    fonttitle=\bfseries,
    title=Definition of Agentic Reasoning
  },
  agentscope/.style={
    colback=AgentIndigoLight,
    colframe=AgentIndigo,
    colbacktitle=AgentIndigo!20!white,
    coltitle=black,
    boxrule=0.9pt, arc=2mm,
    left=3mm, right=3mm, top=2mm, bottom=2mm,
    fonttitle=\bfseries,
    title=Survey Scope
  },
  agentcontrib/.style={
    colback=AgentAmberLight,
    colframe=AgentAmber,
    colbacktitle=AgentAmber!20!white,
    coltitle=black,
    boxrule=0.9pt, arc=2mm,
    left=3mm, right=3mm, top=2mm, bottom=2mm,
    fonttitle=\bfseries,
    title=Contributions
  },
   agentstruct/.style={
      colback=LARGBlue!12!white,
      colframe=LARGBlue,
      colbacktitle=LARGBlue!20!white,
      coltitle=LARGBlue,
    boxrule=0.9pt, arc=2mm,
    left=3mm, right=3mm, top=2mm, bottom=2mm,
    fonttitle=\bfseries,
    title=Survey Structure
  }
}

\newtcolorbox{takeawaybox}[1]{
  colback=white!98!black,
  colframe=white!86!black,
  title={\textcolor{black}{#1}},
  boxrule=0.8pt,
  arc=2pt,
  left=6pt,
  right=6pt,
  top=0pt,
  bottom=0pt,
  before skip=5pt,
}

\graphicspath{ {./images/} }

\setlist[itemize]{leftmargin=12pt}

\runningtitle{Agentic Reasoning for Large Language Models}

\title{%
\textbf{Agentic Reasoning for Large Language Models}\\
\vspace{0.3em}
{\fontsize{11.5}{13.5}\selectfont\scshape\color{LARGBlue!85} $\lozenge$~ Foundations ~$\cdot$~ Evolution ~$\cdot$~ Collaboration ~$\lozenge$}
\vspace{-0.8em}
}

\author{}

\date{\vspace{-3ex}}

\begin{document}

\author{
   \normalfont 
   Tianxin Wei$^{\textcolor{Maroon}{1}\dag}$ \quad 
   Ting-Wei Li$^{\textcolor{Maroon}{1}\dag}$ \quad 
   Zhining Liu$^{\textcolor{Maroon}{1}\dag}$ \quad 
   Xuying Ning$^{\textcolor{Maroon}{1}}$ \quad 
   Ze Yang$^{\textcolor{Maroon}{2}}$ \quad
   Jiaru Zou$^{\textcolor{Maroon}{1}}$ \quad \\
   Zhichen Zeng$^{\textcolor{Maroon}{1}}$ \quad 
   Ruizhong Qiu$^{\textcolor{Maroon}{1}}$ \quad 
   Xiao Lin$^{\textcolor{Maroon}{1}}$ \quad
   Dongqi Fu$^{\textcolor{Maroon}{2}}$ \quad
   Zihao Li$^{\textcolor{Maroon}{1}}$ \quad 
   Mengting Ai$^{\textcolor{Maroon}{1}}$ \quad
   Duo Zhou$^{\textcolor{Maroon}{1}}$ \quad
   Wenxuan Bao$^{\textcolor{Maroon}{1}}$ \quad
   Yunzhe Li$^{\textcolor{Maroon}{1}}$ \quad
   Gaotang Li$^{\textcolor{Maroon}{1}}$ \quad
   Cheng Qian$^{\textcolor{Maroon}{1}}$ \quad
   Yu Wang$^{\textcolor{Maroon}{5}}$ \quad
   Xiangru Tang$^{\textcolor{Maroon}{6}}$ \quad
   Yin Xiao$^{\textcolor{Maroon}{1}}$ \quad
   Liri Fang$^{\textcolor{Maroon}{1}}$ \quad
   Hui Liu$^{\textcolor{Maroon}{3}}$ \quad
   Xianfeng Tang$^{\textcolor{Maroon}{3}}$ \quad
   Yuji Zhang$^{\textcolor{Maroon}{1}}$ \quad
   Chi Wang$^{\textcolor{Maroon}{4}}$ \quad
   Jiaxuan You$^{\textcolor{Maroon}{1}}$ \quad
   Heng Ji$^{\textcolor{Maroon}{1}}$ \quad
   Hanghang Tong$^{\textcolor{Maroon}{1}}$$^{\coloremojicode{2709}}$ \quad
   Jingrui He$^{\textcolor{Maroon}{1}}$$^{\coloremojicode{2709}}$ \quad
   \\
   \vspace{12pt}
   \small

   \raisebox{1.0ex}{\includegraphics[height=1.1ex]{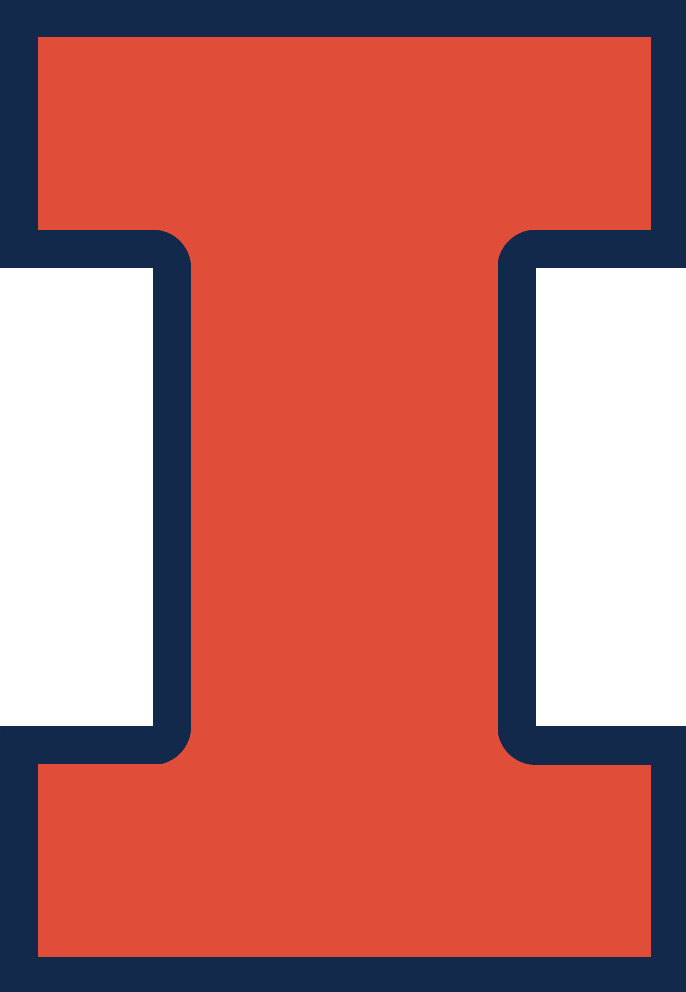}}$^{\textcolor{Maroon}{1}}$University of Illinois Urbana-Champaign \quad
   \raisebox{0.9ex}{\includegraphics[height=1.1ex]{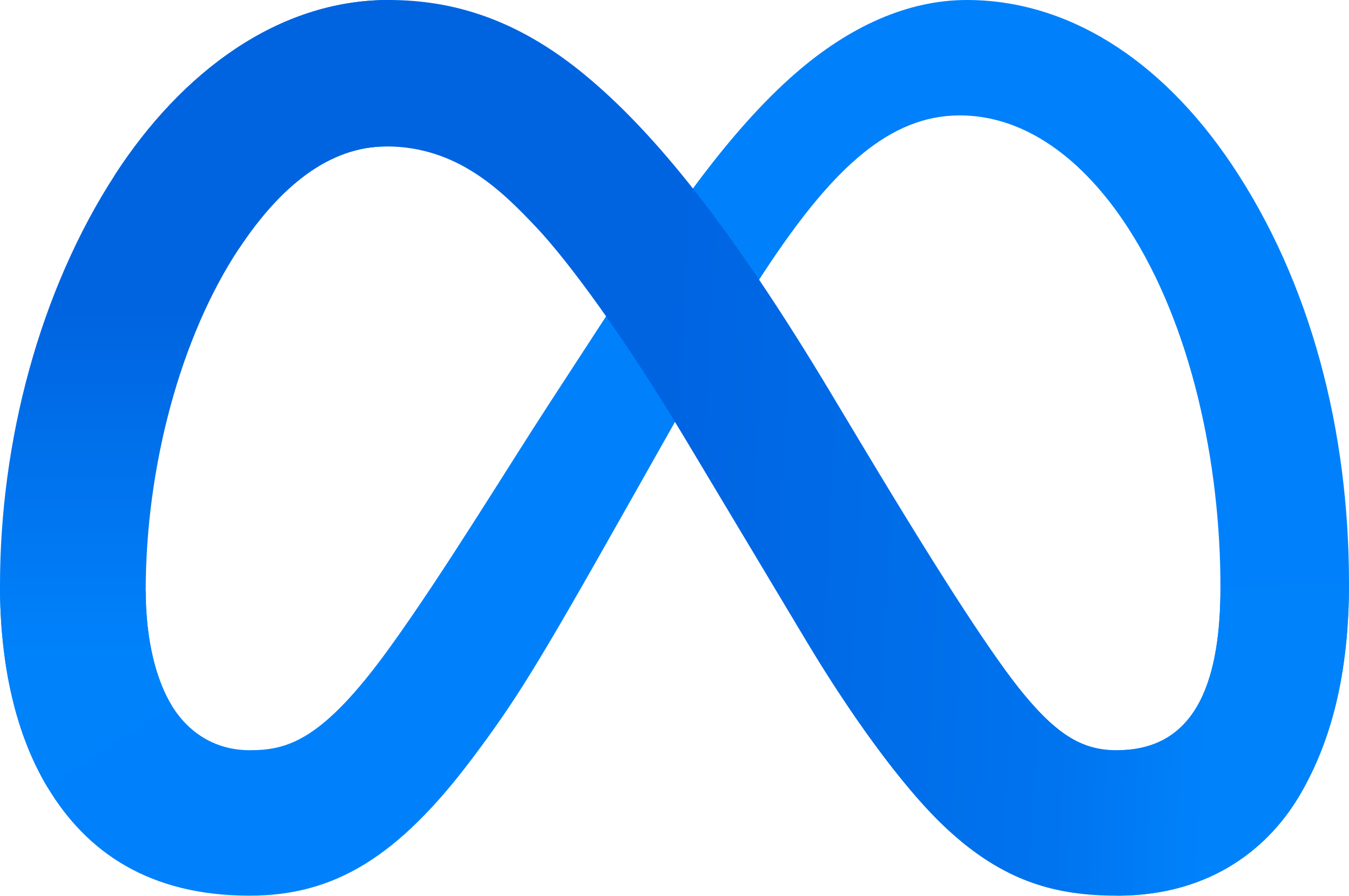}} $^{\textcolor{Maroon}{2}}$Meta \quad
   \raisebox{0.75ex}{\includegraphics[height=1.4ex]{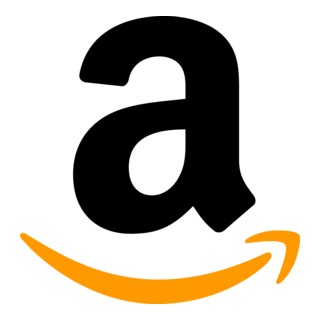}}
   $^{\textcolor{Maroon}{3}}$Amazon \quad
   \raisebox{0.9ex}{\includegraphics[height=1.2ex]{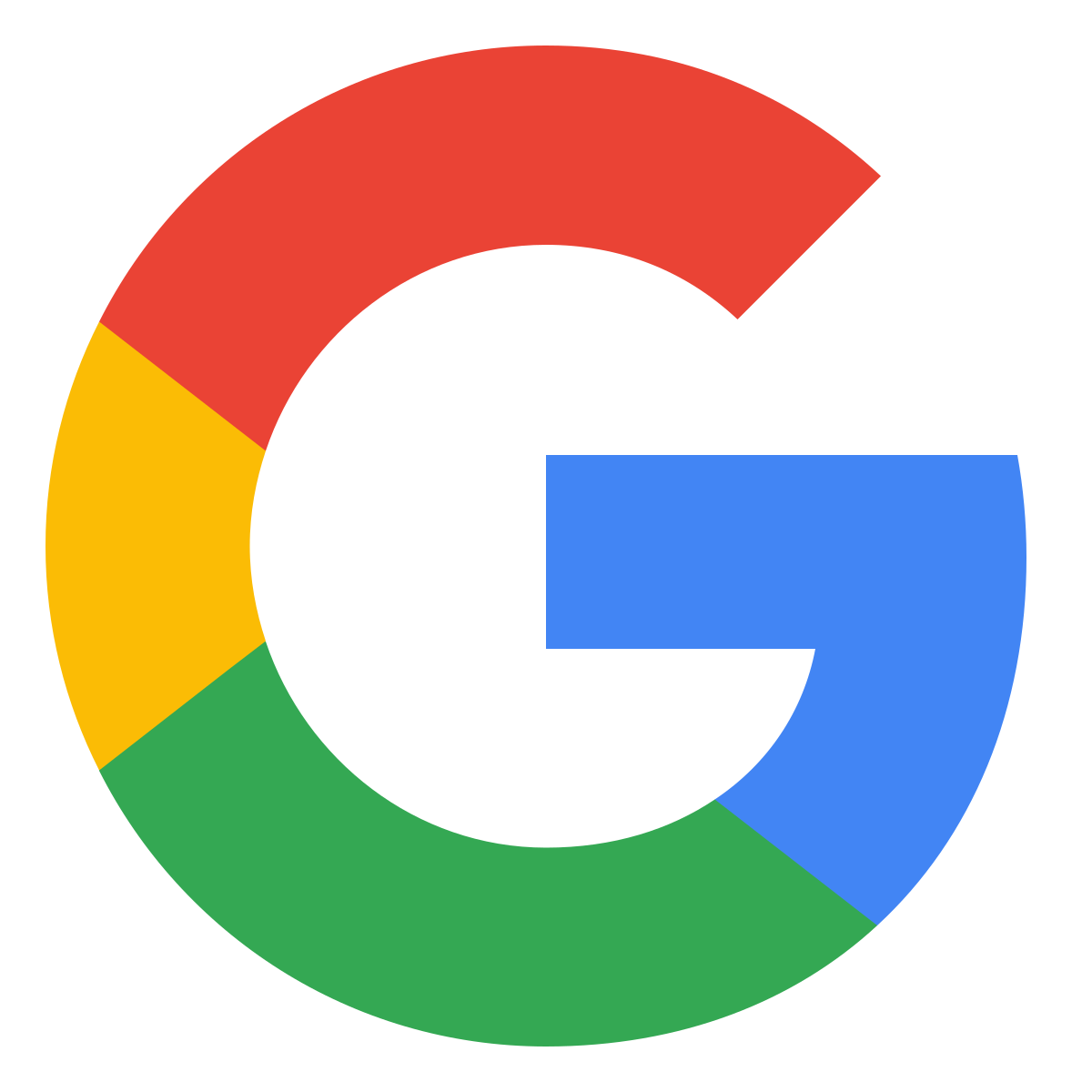}} $^{\textcolor{Maroon}{4}}$Google Deepmind \quad\\
   \raisebox{0.9ex}{\includegraphics[height=1.4ex]{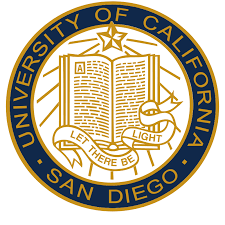}} $^{\textcolor{Maroon}{5}}$UCSD \quad
   \raisebox{0.9ex}{\includegraphics[height=1.2ex]{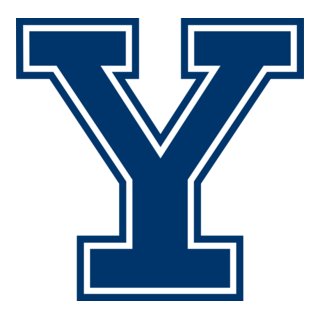}} $^{\textcolor{Maroon}{6}}$Yale \quad
   \\
   $^{\dag}$ \textit{Equal contribution}, \quad  $^{\coloremojicode{2709}}$ \textit{Corresponding Author}
}

\begin{abstract}

 \textbf{\large Abstract:} Reasoning is a fundamental cognitive process underlying inference, problem-solving, and decision-making. While large language models (LLMs) demonstrate strong reasoning capabilities in closed-world settings, exemplified by standard benchmarks in mathematics and code, they struggle in open-ended and dynamic environments. The emergence of \emph{agentic reasoning} marks a paradigm shift, bridging thought and action by reframing LLMs as autonomous agents that plan, act, and learn through continual interaction. In this survey, we provide a systematic roadmap by organizing agentic reasoning along three complementary dimensions. First, we characterize environmental dynamics through three layers: \emph{foundational agentic reasoning} establishes core single-agent capabilities, including planning, tool use, and search, that operate in stable environments; \emph{self-evolving agentic reasoning} examines how agents refine these capabilities through feedback, memory, and adaptation in evolving settings; and \emph{collective multi-agent reasoning} extends intelligence to collaborative scenarios where multiple agents coordinate roles, share knowledge, and pursue shared goals. Across all layers, we analyze system constraints and optimization settings by distinguishing \emph{in-context reasoning}, which scales test-time interaction through structured orchestration and adaptive workflow design, from \emph{post-training reasoning}, which optimizes behaviors through reinforcement learning and supervised fine-tuning. We further review and contextualize agentic reasoning frameworks in real-world applications and benchmarks spanning science, robotics, healthcare, autonomous research, and math, illustrating how different reasoning mechanisms are instantiated and evaluated across domains. This survey synthesizes agentic reasoning methods into a unified roadmap that bridges thoughts and actions, offering actionable guidance for agentic systems across environmental dynamics, optimization settings, and agent interaction settings. Finally, we outline open challenges and future directions, situating how agentic reasoning has developed while identifying what remains ahead: personalization, long-horizon interaction, world modeling, scalable multi-agent training, and governance frameworks for real-world deployment.

\vspace{5mm}
  \coloremojicode{1F3AF}~\textbf{Keywords}: Agentic AI, LLM Agent, Agentic Reasoning, Self-evolving

\faGithub~\textbf{Github}: \href{https://github.com/weitianxin/Awesome-Agentic-Reasoning}{https://github.com/weitianxin/Awesome-Agentic-Reasoning}
\end{abstract}
\maketitle

\begin{figure}[ht]
  \centering
  \includegraphics[width=\linewidth]{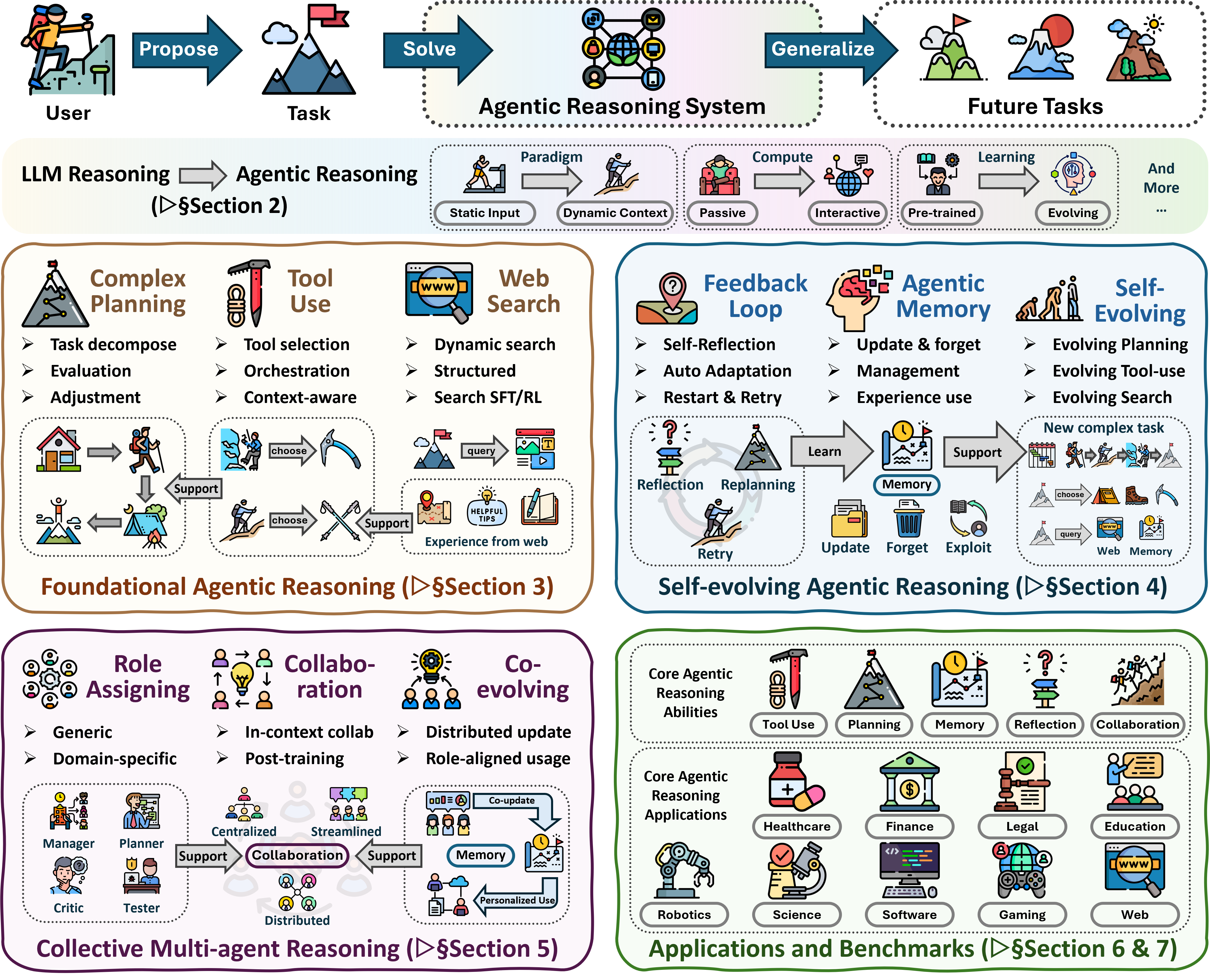}
  \caption{
  An overview of agentic reasoning.
  }
  \label{fig:concept}
  \vspace{-0.6cm}
\end{figure}

\section{Introduction}
\label{sec:intro}

Reasoning lies at the core of intelligence, enabling logical inference, problem-solving, and decision-making across interactive and dynamic settings. Large language models (LLMs) have achieved remarkable gains in closed-world domains such as mathematical problem solving and code generation. Empirically, techniques that explicitize intermediate reasoning, such as Chain-of-Thought prompting, decomposition, and program-aided solving, have significantly bolstered inference performance \cite{wei2022chain,zhou2022least,gao2023pal,yao2023tree}. Yet, these approaches often assume static contexts and short-horizon reasoning. Conventional LLMs lack mechanisms to act, adapt, or improve in open-ended environments where information evolves over time.

In this survey, we systematize this evolution under the framework of \emph{Agentic Reasoning}: rather than passively generating sequences, LLMs are reframed as autonomous reasoning agents that plan, act, and learn through continual interaction with their environment. This reframing unifies \emph{reasoning} with \emph{acting}, positioning reasoning as the organizing principle for perception, planning, decision, and verification. Systems such as ReAct~\cite{yao2023react} interleave deliberation with environment interaction, tool-use frameworks enable self-directed API calling, and workflow-based agents dynamically orchestrate sub-tasks and verifiable actions~\cite{yao2023react,schick2023toolformer,shen2023hugginggpt}. Conceptually, this parallels the shift from static, one-shot inference to sequential decision-making under uncertainty. Unlike simple input-output mapping, this paradigm requires agents to plan over long horizons, navigate partial observability, and actively improve through feedback~\cite{wang2024survey,singh2025agentic,huang2024survey}.

\begin{tcolorbox}[
  agentdef,
  float,
  floatplacement=t
]
\textbf{Agentic reasoning} positions reasoning as the central mechanism of intelligent agents, spanning \emph{foundational capabilities} (planning, tool use, and search), \emph{self-evolving adaptation} (feedback, and memory-driven adaptation), and \emph{collective coordination} (multi-agent collaboration), realizable through either \emph{in-context} orchestration or \emph{post-training} optimization.
\end{tcolorbox}

To systematically characterize the environmental dynamics, we structure our survey around three complementary scopes of agentic reasoning: foundational capabilities, self-evolution, and collective intelligence, spanning diverse interactive and dynamic settings. \emph{\textbf{Foundational Agentic Reasoning}} establishes the bedrock of core single-agent capabilities, including planning, tool use, and search, that enable operations within stable, albeit complex, environments. Here, agents act by decomposing goals, invoking external tools, and verifying results through executable actions. For instance, program-aided reasoning~\cite{gao2023pal} grounds logical derivations in code execution; repository-level systems such as OpenHands~\cite{wang2024openhands} integrate reasoning, planning, and testing into unified loops; and structured memory modules~\cite{chhikara2025mem0,li2025memos} transform factual recall into procedural competence by persisting intermediate reasoning traces for reuse.

Building upon these foundations, \emph{\textbf{Self-Evolving Agentic Reasoning}} enables agents to improve continually through cumulative experience. Encompassing task-specific \emph{self-improvement} (e.g., via iterative critique), this paradigm extends adaptation to include persistent updates of internal states like memory and policy. Rather than following fixed reasoning paths, agents develop mechanisms for feedback integration and memory-driven adaptation to navigate evolving environments. Reflection-based frameworks such as Reflexion~\cite{shinn2023reflexion} allow agents to critique and refine their own reasoning processes, while reinforcement formulations such as RL-for-memory~\cite{yan2025memoryr1} formalize memory writing and retrieval as policy optimization. Through these mechanisms, agents dynamically integrate inference-time reasoning with learning, progressively updating internal representations and decision policies without full retraining. This continual adaptation links reasoning with learning, enabling models to accumulate competence, and generalize across tasks.

Finally, \emph{\textbf{Collective Multi-Agent Reasoning}} scales intelligence from isolated solvers to collaborative ecosystems. Rather than operating in isolation, multiple agents coordinate to achieve shared goals through explicit role assignment (e.g., manager–worker–critic), communication protocols, and shared memory systems~\cite{chen2023autoagents,hong2024metagpt}. As agents specialize in subtasks and refine each other’s outputs, collaboration amplifies reasoning diversity, enabling systems to debate, resolve disagreements, and achieve consistency through natural language-based multi-turn interactions~\cite{unleashing2024,wang2024battleagentbench}. However, this complexity also introduces challenges in stability, communication efficiency, and trustworthiness, necessitating structured coordination frameworks and rigorous evaluation standards~\cite{xiao2023agentbench,zhu2025multiagentbench}.

Across all layers, we analyze system constraints and optimization settings by distinguishing two complementary modes, corresponding to inference-time orchestration \cite{yao2023react,shinn2023reflexion,ni2024tree,li2025search,xu2025mem,wei2025evo} and training-based capability optimization \cite{ma2024coevolving,jin2025search,wei2025webagent,yan2025memoryr1}. \emph{\textbf{In-context Reasoning}} focuses on scaling inference-time compute: through structured orchestration, search-based planning, and adaptive workflow design, it enables agents to navigate complex problem spaces dynamically without modifying model parameters. Conversely, \emph{\textbf{Post-training Reasoning}} targets capability internalization: it consolidates successful reasoning patterns or tool-use strategies into the model's weights via reinforcement learning and fine-tuning. Together, they provide an actionable roadmap for designing agents.

\begin{tcolorbox}[
  agentscope,
  float,
  floatplacement=t
]
This survey reviews \emph{reasoning-empowered agentic systems} where reasoning drives adaptive behavior. We analyze these systems through two complementary optimization modes:
\begin{itemize}
    \item \textbf{In-context Reasoning}: scales inference-time interaction through structured orchestration and planning without parameter updates.
    \item \textbf{Post-training Reasoning}: internalizes reasoning strategies into model parameters via reinforcement learning and fine-tuning.
\end{itemize}
Our scope covers methodologies embedding these modes into planning, memory, and self-improvement across single-agent and multi-agent contexts. This survey summarizes progress up to 2025.
\end{tcolorbox}

Building on the three-layer taxonomy, agentic reasoning has begun to underpin a wide range of practical applications, from mathematical exploration \cite{trinh2024solving,romera2024mathematical} and vibe coding \cite{wang2024openhands,sapkota2025vibe,karpathy2025vibecoding} to scientific discovery \cite{bran2023chemcrowaugmentinglargelanguagemodels,bousetouane2025physicalaiagentsintegrating,ding2024matexpertdecomposingmaterialsdiscovery}, embodied robotics \cite{wang2023voyager,meghan2024embodiedrag,zhao2025embodied}, healthcare \cite{li2024mmedagent,huang2025biomni}, and autonomous web exploration~\cite{li2025websailor,zheng2025skillweaver}.
These applications expose distinct reasoning demands shaped by domain-specific data modalities, interaction constraints, and feedback loops, motivating diverse system designs \cite{sapkota2025ai,liu2024dynamic} that integrate planning, tool use, search, reflection, memory mechanisms, and multi-agent coordination.
On the other hand, the benchmark landscape has emerged to evaluate agentic reasoning, ranging from targeted tests that isolate individual agentic capabilities to application-specific benchmarks that assess end-to-end behavior in domain-specific environments and scenarios~\cite{shuyan2023webarena,jing2024visualwebarena,lawrence2024videowebarena,shridhar2020alfworld,xiao2023agentbench,zhu2025multiagentbench,deng2023mind2web,gou2025mind2web2}.

Together, this survey synthesizes agentic reasoning methods into a unified roadmap that bridges reasoning and acting. We systematically characterize these methods across the complementary scopes of foundational, self-evolving, and collective reasoning, while distinguishing between in-context and post-training optimization modes. We further contextualize this roadmap through representative applications and evaluation benchmarks, illustrating how different agentic reasoning mechanisms are instantiated and assessed across realistic domains and task settings. Finally, we outline open challenges and future directions, identifying key frontiers such as personalization, long-horizon interaction, world modeling, scalable multi-agent training, and governance frameworks for real-world deployment.

\begin{tcolorbox}[agentcontrib]
This survey makes the following contributions:
\begin{itemize}
\item \textbf{Conceptual framing}: We formalize the paradigm of \emph{Agentic Reasoning}, spanning foundational, self-evolving, and collective reasoning layers.
\item \textbf{Systematic review}: We analyze single-agent, adaptive, and multi-agent systems, emphasizing reasoning-centered workflow orchestration across in-context and post-training dimensions.

\item \textbf{Applications and evaluation}: We review real-world applications and benchmarks to illustrate the instantiation and evaluation of agentic reasoning mechanisms.
\item \textbf{Future agenda}: We identify emerging challenges in robustness, trustworthiness, and efficiency, outlining directions for the next generation of adaptive and collaborative agents.
\end{itemize}
\end{tcolorbox}

\newpage
\addtocontents{toc}{\protect\setcounter{tocdepth}{3}}
\tableofcontents

\begin{tcolorbox}[agentstruct,title=\textbf{Survey Structure}]
This survey is organized as follows:
\begin{itemize}
  \item \textbf{Sec. \ref{sec:from}: \emph{Preliminaries}.} Key background on LLM and Agentic reasoning.
  \item \textbf{Sec. \ref{sec:foundational}: \emph{Foundational Agentic Reasoning}.} Core single-agent capabilities including planning, tool use, and search.
  \item \textbf{Sec. \ref{sec:selfevolve}: \emph{Self-evolving Reasoning}.} Feedback, memory, and continual adaptation mechanisms that enhance reasoning over time.
  \item \textbf{Sec. \ref{sec:collective}: \emph{Collective Multi-agent Reasoning}.} Coordination, communication, and shared-memory strategies for collaboration.
  \item \textbf{Sec. \ref{sec:applications}: \emph{Applications}.} Reasoning-empowered applications across science, robotics, healthcare, autonomous research and math/code.
  \item \textbf{Sec. \ref{sec:benchmarks}: \emph{Benchmarks}.} Datasets, metrics, and evaluation protocols for assessing reasoning and agentic abilities.
  \item \textbf{Sec. \ref{sec:openproblems}: \emph{Open Problems}.} Challenges and future directions for AI Agent reasoning.
\end{itemize}
\end{tcolorbox}

\section{From LLM Reasoning to Agentic Reasoning}
\label{sec:from}

Traditional reasoning with large language models (LLMs) is typically formulated as a one-shot or few-shot prediction task over static inputs. These models rely on scaling \textbf{test-time computation}, improving accuracy by increasing model size or inference budget, but without the ability to interact, remember, or adapt to changing goals. Methods such as prompt engineering, in-context learning, and chain-of-thought prompting have made reasoning more explicit, yet conventional LLMs remain passive sequence predictors that operate within fixed prompts.

\emph{Agentic reasoning}, in contrast, emphasizes \textbf{scaling test-time interaction}. Instead of depending solely on internal parameters, agentic systems reason through action: invoking tools, exploring alternatives, updating memory, and integrating feedback. This transforms inference into an iterative process that includes decision steps, reflection, and learning from experience. Reasoning becomes a dynamic loop that connects the model, memory, and environment.

\begin{table}[h]
\centering
\caption{Contrasting capabilities of \textbf{LLM reasoning} and \textbf{agentic reasoning}.}
\label{tab:llm-vs-agentic}
\renewcommand{\arraystretch}{1.25}
\begin{tabular}{
  >{\centering\arraybackslash\bfseries}m{3.2cm}
  @{\hspace{3pt}}
  >{\centering\arraybackslash}m{4.5cm}
  @{\hspace{2pt}}
  >{\centering\arraybackslash}m{0.5cm}
  @{\hspace{2pt}}
  >{\centering\arraybackslash}m{4.5cm}
}
\toprule
\textbf{Dimension}
& \llmchip{\textbf{LLM Reasoning}}
& $\leftrightarrow$
& \agentchip{\textbf{Agentic Reasoning}} \\
\midrule
\multirow{2}{3.2cm}{\centering Paradigm}
  & \llmchip{passive}
  & $\leftrightarrow$
  & \agentchip{interactive} \\
  & \llmchip{static input}
  & $\leftrightarrow$
  & \agentchip{dynamic context} \\
\midrule
\multirow{2}{3.2cm}{\centering Computation}
  & \llmchip{single pass}
  & $\leftrightarrow$
  & \agentchip{multi step} \\
  & \llmchip{internal compute}
  & $\leftrightarrow$
  & \agentchip{with feedback} \\
\midrule
\multirow{2}{3.2cm}{\centering Statefulness}
  & \llmchip{context window}
  & $\leftrightarrow$
  & \agentchip{external memory} \\
  & \llmchip{no persistence}
  & $\leftrightarrow$
  & \agentchip{state tracking} \\
\midrule
\multirow{2}{3.2cm}{\centering Learning}
  & \llmchip{offline pretraining}
  & $\leftrightarrow$
  & \agentchip{continual improvement} \\
  & \llmchip{fixed knowledge}
  & $\leftrightarrow$
  & \agentchip{self evolving} \\
\midrule
\multirow{2}{3.2cm}{\centering Goal Orientation}
  & \llmchip{prompt based}
  & $\leftrightarrow$
  & \agentchip{explicit goal} \\
  & \llmchip{reactive}
  & $\leftrightarrow$
  & \agentchip{planning} \\
\bottomrule
\end{tabular}
\end{table}

This transition marks a conceptual shift: reasoning no longer scales through static capacity, but through structured interaction that enables planning, adaptation, and collaboration across time and tasks.

\subsection{Positioning Our Survey}

While several recent surveys have examined LLM reasoning or agent architectures \citep{huang2022towards, chen2025towards, xu2025towards, ke2025survey, zhang2025survey, zhang2025landscape, lin2025comprehensive, fang2025comprehensive, gao2025survey}, our work focuses specifically on \textbf{agentic reasoning} as a unified paradigm for understanding reasoning as interaction. We position this survey at the intersection of model-centric reasoning and system-level intelligence, aiming to bridge prior discussions on reasoning mechanisms and agent architectures.

\textbf{Relation to LLM Reasoning Surveys.}
Existing surveys on LLM reasoning mainly investigate how to elicit or enhance reasoning within a model’s internal computation process.
For example, \citet{huang2022towards, chen2025towards, xu2025towards, ke2025survey} summarize prompting and scaling techniques such as chain-of-thought, reinforcement post-training, and long-context reasoning, emphasizing how LLMs can learn to reason better through inference-time supervision or post-training alignment.
These works improve the internal expressiveness of reasoning traces but typically remain within static inference settings, where reasoning unfolds in a single forward pass without external interaction.
In contrast, our survey examines how reasoning extends \emph{beyond} text generation, encompassing dynamic planning, adaptive memory, and feedback-driven behavior during deployment.

\textbf{Relation to AI Agent Surveys.}
Several contemporary surveys have begun to explore LLM-based agents from architectural or system perspectives \citep{zhang2025landscape, lin2025comprehensive, fang2025comprehensive, gao2025survey}.
These works analyze how agents employ reinforcement learning, planning, and tool-use modules to operate in complex environments.
For instance, \citet{zhang2025landscape, lin2025comprehensive} focus on reinforcement learning for agentic search and decision-making, while \citet{fang2025comprehensive, gao2025survey} emphasize self-evolving and lifelong agentic systems that continuously learn from interaction.
Our focus complements these perspectives by centering on the \textbf{reasoning process} that these architectures enable, specifically how interaction, feedback, and collaboration transform static inference into adaptive reasoning.
Rather than viewing reasoning as an implicit by-product of architectural design, we treat it as the unifying mechanism that links single-agent reinforcement, multi-agent coordination, and self-evolving intelligence.

In summary, our survey provides a reasoning-centric lens on intelligent agency.
We examine how foundational reasoning mechanisms, post-training adaptation, and long-term self-evolution jointly constitute the basis of \emph{agentic reasoning}, illustrating the transition from static prediction to interactive, adaptive, and continually improving intelligence.

\subsection{Preliminaries}
\label{sec:prelims}

This subsection formalizes the transition from static language modeling to agentic reasoning. To align with the \textbf{three-layered dimensions} (Foundational, Self-Evolving, Collaboration) outlined in the introduction, we unify these capabilities under a single {control-theoretic framework}.

\paragraph{Formalizing Agentic Reasoning: A Latent-Space View.}
Standard approaches often conflate the agent's context with the environment state. We model the environment as a \textbf{Partially Observable Markov Decision Process (POMDP)} and introduce an internal \emph{reasoning variable} to expose the ``think--act'' structure of agentic policies. Concretely, we consider the tuple
$\langle \mathcal{X}, \mathcal{O}, \mathcal{A}, \mathcal{Z}, \mathcal{M}, \mathcal{T}, \Omega, \mathcal{R}, \gamma \rangle$, where
$\mathcal{X}$ is the latent \emph{environment state} space (unobservable to the agent),
$\mathcal{O}$ is the observation space (e.g., user queries, API returns),
$\mathcal{A}$ is the external action space (e.g., tool invocation, final answer),
$\mathcal{Z}$ is a \emph{reasoning trace} space (e.g., latent plans, optionally verbalized as chain-of-thought),
and $\mathcal{M}$ is the agent's \emph{internal memory/context} space (e.g., a sufficient statistic of interaction history).
$\mathcal{T}$ and $\Omega$ denote the transition and observation kernels, $\mathcal{R}$ the reward, and $\gamma\in(0,1)$ the discount factor.

At timestep $t$, the agent conditions on a history
$h_t = (o_{\le t}, z_{<t}, a_{<t})$ (i.e., $o_t$ is observed before generating $z_t$ and then $a_t$).
Equivalently, the history can be summarized by an internal memory state $m_t\in\mathcal{M}$.
Crucially, we distinguish external actions from internal reasoning. We factorize the policy as
\begin{equation}
\label{eq:policy_decomp}
\pi_\theta(z_t, a_t \mid h_t)
= \underbrace{\pi_{\text{reason}}(z_t \mid h_t)}_{\text{Internal Thought}}
\cdot \underbrace{\pi_{\text{exec}}(a_t \mid h_t, z_t)}_{\text{External Action}}.
\end{equation}
This decomposition highlights the core shift in agentic systems: performing computation in $\mathcal{Z}$ (thinking) before committing to $\mathcal{A}$ (acting). The objective remains maximizing the expected return
$J(\theta)=\mathbb{E}_{\tau}\!\left[\sum_{t\ge 0}\gamma^t r_t\right]$.

\paragraph{In-Context Reasoning: Inference-Time Search.}
In this regime, model parameters $\theta$ are frozen. The agent optimizes the reasoning trajectory by searching over $\mathcal{Z}$ to maximize a heuristic value function $\hat{v}(h_t, z)$.
We model inference as selecting a trajectory $\tau = (h_0, z_0, a_0, h_1, z_1, a_1, \ldots)$.
Methods like ReAct \citep{yao2023react} perform greedy decoding over alternating thoughts $z$ and actions $a$.
Tree-of-Thoughts (ToT \citep{yao2023tree}) and related MCTS-style approaches treat partial thoughts as nodes $u \in \mathcal{U}$ (e.g., a representation derived from $(h_t,z_t)$) and search for an optimal path:
\begin{equation}
\tau^\star \in \arg\max_{\tau} \sum_{t} \hat{v}_\phi(u_t),
\end{equation}
where $\hat{v}_\phi$ is a heuristic evaluator or verifier. This corresponds to planning in $\mathcal{Z}$ without updating the policy parameters.

\paragraph{Post-Training: Policy Optimization.}
This paradigm optimizes $\theta$ to align the policy with long-horizon rewards $r_t$ (e.g., correctness, safety), including reasoning models (e.g., DeepSeek-R1 \citep{guo2025deepseekr1}) and learning-to-search systems (e.g., Search-R1 \citep{jin2025search}, DeepRetrieval \citep{jiang2025deepretrieval}) that train multi-turn reasoning or tool use with RL.
While PPO \citep{schulman2017proximal} is standard, \textbf{Group Relative Policy Optimization (GRPO)} \citep{shao2024deepseekmath}-based methods are widely used for reasoning tasks. GRPO eliminates the value network by constructing advantages from group-relative rewards.
For a group of $G$ sampled outputs $\{y_i\}_{i=1}^G$ from the same prompt $q$, a common GRPO objective is:
\begin{equation}
\mathcal{L}^{\text{GRPO}}(\theta)
= \mathbb{E}_{q \sim P(Q)} \left[
\frac{1}{G} \sum_{i=1}^G
\left(
\min\!\left( \rho_i \hat{A}_i,\; \mathrm{clip}(\rho_i, 1-\epsilon, 1+\epsilon)\hat{A}_i \right)
- \beta\, \mathbb{D}_{KL}(\pi_\theta \,\|\, \pi_{\text{ref}})
\right)
\right],
\end{equation}
where $\rho_i = \frac{\pi_\theta(y_i \mid q)}{\pi_{\theta_{\text{old}}}(y_i \mid q)}$ and the group-normalized advantage is
\begin{equation}
\hat{A}_i
= \frac{r_i - \mu}{\sigma + \delta},\quad
\mu=\frac{1}{G}\sum_{j=1}^G r_j,\quad
\sigma=\sqrt{\frac{1}{G}\sum_{j=1}^G (r_j-\mu)^2},
\end{equation}
with $\delta>0$ a small constant for numerical stability.
Advanced methods such as ARPO \citep{lu2025arpo} and DAPO \citep{yu2025dapo} extend this framework to handle sparse rewards and improve stability in complex tool-use environments (e.g., via replay/rollout strategies and decoupled clipping).

\paragraph{Collective Intelligence: Multi-Agent Reasoning.}
We extend the single-agent formulation to a \emph{decentralized} partially observable multi-agent setting, commonly formalized as a \emph{Dec-POMDP}. The core distinction lies in expanding each agent's observation to include a \textbf{communication channel} $\mathcal{C}$. For a system of $N$ agents, the joint policy $\boldsymbol{\pi}$ is composed of individual policies $\pi^i$, where agent $i$'s observation $o^i_t$ explicitly includes communicative messages $c^{-i}_{t-1}$ generated by peers.
Crucially, in agentic MARL, communication is not merely signal transmission but an extension of the reasoning process: one agent's external action can act as a prompt that triggers another agent's internal reasoning chain.
Existing frameworks like AutoGen \citep{wu2024autogen} and CAMEL \citep{li2023camel} represent static role-playing with fixed policies. Recent agentic RL advances (e.g., GPTSwarm~\cite{zhuge2024gptswarm}, MaAS, agents trained via PPO/GRPO~\cite{hong2025multi}) aim to \emph{optimize} this joint reasoning distribution. The challenge shifts from single-agent planning to \textbf{mechanism design}: optimizing the communication topology and incentive structures to align decentralized reasoning processes $\pi^i_{\text{reason}}$ toward a coherent global objective, often utilizing Centralized-Training/Decentralized-Execution (CTDE) paradigms to stabilize the emergence of cooperative behaviors.

\paragraph{Self-Evolving Agents: The Meta-Learning Loop.}
While foundational agents optimize reasoning $z$ within an episode, self-evolving agents optimize the agent system itself across episodes $k=1, \dots, K$. Let $\mathcal{S}_k$ denote the evolvable system state (e.g., explicit memories, tool libraries, or code). A generic meta-update rule is
\begin{equation}
\mathcal{S}_{k+1} \leftarrow U(\mathcal{S}_k, \tau_k, \mathcal{F}_k),
\end{equation}
where $\mathcal{F}_k$ represents environmental feedback (rewards, execution errors) and $\mathcal{S}_k$ represents the evolvable state. We categorize self-evolution by the nature of $\mathcal{S}$:
\begin{itemize}[leftmargin=*]
    \item \textbf{Verbal Evolution:} $\mathcal{S}$ consists of textual reflections or guidelines. Methods like Reflexion \citep{shinn2023reflexion} update $\mathcal{S}$ by synthesizing error logs into linguistic cues that condition future reasoning policies.
    \item \textbf{Procedural Evolution:} $\mathcal{S}$ consists of a library of executable tools or skills. Agents like Voyager \citep{wang2023voyager} evolve by synthesizing new code-based skills, expanding the action space $\mathcal{A}$ permanently.
    \item \textbf{Structural Evolution:} $\mathcal{S}$ consists of the agent's source code or architecture itself. Advanced methods like AlphaEvolve \citep{novikov2025alphaevolve} treat the agent's code as a hypothesis space, using an LLM as a mutation operator to search for superior reasoning algorithms.
\end{itemize}

This framework unifies these diverse approaches as gradient-free or gradient-based optimization steps over the agent's explicit memories and artifacts (and optionally parameters), closing the loop between experience and competence.

\section{Foundational Agentic Reasoning}
\label{sec:foundational}

Agentic reasoning originates from the behavior of a single agent. 
Before discussing adaptation and collaboration, we focus on how an individual agent translates reasoning into structured action through three core components: \emph{planning}, \emph{search}, and \emph{tool use}. 
In this setting, the agent is not a passive text generator but an autonomous problem solver that formulates plans, explores alternatives through retrieval or environment search, and leverages tools to execute grounded operations. 
Together, these mechanisms establish the foundation of agentic reasoning, linking abstract deliberation with verifiable action.

A canonical foundational workflow can be viewed as an iterative cycle that interleaves 
\textbf{planning} (goal decomposition and task formulation),  \textbf{tool use} (invoking external systems or APIs to act on the world) and \textbf{search} (retrieval and exploration for decision support), 
Reasoning serves as the organizing principle across these stages, determining when to plan, what to retrieve, and how to act, transforming static inference into interactive decision-making.

By analyzing these components, we clarify how structured reasoning elevates a static LLM into an autonomous, goal-driven agent. 
The next section introduces \textbf{self-evolving reasoning}, where \emph{feedback} and \emph{memory} enable continual adaptation and extension of these foundational capabilities. 
Subsequently, we examine \textbf{collective reasoning}, in which multiple agents coordinate through roles, communication, and shared memory to achieve objectives beyond individuals.

\subsection{Planning Reasoning} % (ToT, ReWOO, hierarchical)

Planning is a central component of intelligent behavior, enabling agents to decompose problems, sequence decisions, and navigate complex environments with foresight. Recent research has increasingly explored planning in the context of large language models (LLMs), either as autonomous agents or as components in broader systems. In this section, we categorize existing work in agent planning for reasoning into six methodological styles, where each category highlights a distinct planning strategy that supports complex agentic reasoning. 

\begin{figure}
    \centering
    \includegraphics[width=\linewidth]{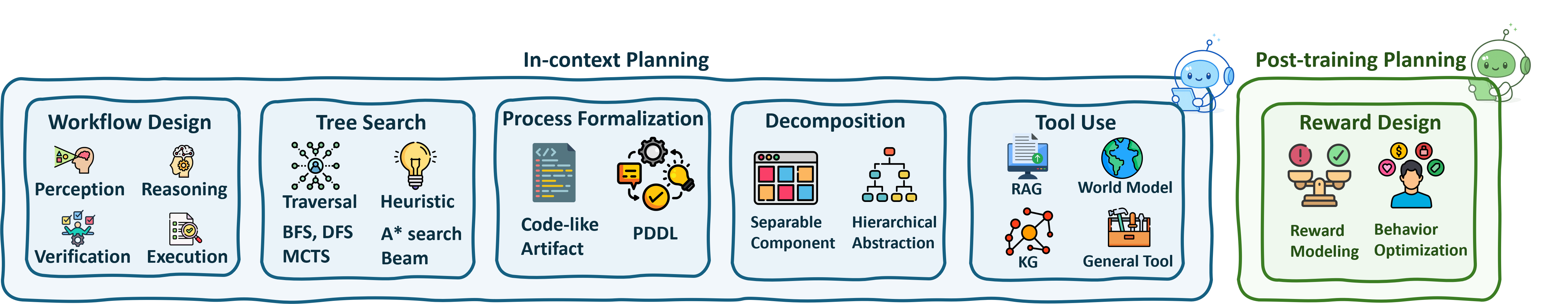}
    \caption{Overview of \textbf{Planning Reasoning} in LLM agents, categorized into in-context planning and post-training planning.}
    \label{fig:planning}
\end{figure}

\begin{table*}[!t]
\centering
\caption{Representative \textbf{Agentic Planning} systems categorized by \textit{Modality}, \textit{Structure}, \textit{Format}, and \textit{Tool}.}
\renewcommand{\arraystretch}{1.15}
\setlength{\tabcolsep}{6pt}
\begin{tabular}{p{3cm}>{\centering\arraybackslash}p{3cm}>{\centering\arraybackslash}p{4cm}>{\centering\arraybackslash}p{5.5cm}}
\toprule
\textbf{Method} & \textbf{Structure} & \textbf{Format} & \textbf{Tool} \\
\midrule
\rowcolor[HTML]{F5F5F5}
\multicolumn{4}{l}{\textbf{Modality I: Language Agents (e.g., Search Agents, Code Agents)}} \\
ReWOO~\cite{xu2023rewoo} & Decomposed & Natural Language & None \\
Reflexion~\cite{shinn2023reflexion} & Sequential & Natural Language & None \\
LLM+P~\cite{liu2023llmp} & Sequential & Formal Language & None \\
IPC~\cite{valmeekam2023planning} & Sequential & Formal Language & None \\
ToT~\cite{yao2023tree} & Tree & Natural Language & None \\
GoT~\cite{besta2024graph} & Graph & Natural Language & None \\
AoT~\cite{sel2023algorithm} & Graph & Natural Language & None \\
HTP~\cite{gui2025hypertree} & Hypertree & Natural Language & Retrieval \\
RefPlan~\cite{jeong2025reflect} & Tree & Constrained Space & None \\
Gorilla~\cite{patil2024gorilla} & Sequential & Programming Language & Retrieval, API \\
CodeNav~\cite{gupta2024codenav} & Sequential & Programming Language & Code Indexer, Code Search \\
PoG~\cite{chen2024plan} & Graph & Natural Language & Knowledge Graph \\
Tool-Planner~\cite{liu2025toolplanner} & Sequential & Natural Language & Tool Cluster \\
\midrule
\rowcolor[HTML]{F5F5F5}
\multicolumn{4}{l}{\textbf{Modality II: Visual/Multimodal Agents (e.g., GUI Agents, Embodied Agents)}} \\
VisualPredictor~\cite{liang2024visualpredicator} & Tree & Formal Language & None \\
LLM-Planner~\cite{song2023llmplannerfewshotgroundedplanning} & Sequential & Formal Language & Object Detector, KNN \\
Agent-E~\cite{abuelsaad2024agent} & Sequential & Formal Language & DOM Grounder, Screenshot \\
Agent S~\cite{agashe2024agent} & Hierarchical & Natural Language & API, Search, Memory \\
ExRAP~\cite{yoo2024exploratory} & Sequential & Natural Language & Memory \\
AESOP~\cite{sinha2024real} & Reactive & Natural Language & Anomaly Detector \\
HRV~\cite{cornelio2025hierarchical} & Hierarchical & Formal Language & Symbolic Verifier \\
BehaviorGPT~\cite{zhou2024behaviorgpt} & Sequential & Visual Features & World Model \\
Dino-WM~\cite{zhou2024dino} & Tree & Visual Features & World Model \\
FLIP~\cite{gao2024flip} & Sequential & Visual Features & Language Model \\
\bottomrule
\end{tabular}
\label{tab:agentic_planning_methods}
\end{table*}

\subsubsection{In-context Planning}
\paragraph{Workflow Design.} Workflow-based approaches often emphasize structuring the overall planning process into distinct stages (e.g., perception, reasoning, execution, verification), which are either explicitly scaffolded or learned implicitly. For example, \cite{liu2023llmp,valmeekam2023planning,xu2023rewoo,hao2024llm} design planning pipelines that decompose task solving into subtasks, often leveraging a deliberate plan-and-act framework. Similarly, \cite{zhou2022least,wang2023plan,sel2023algorithm,shen2023hugginggpt} rely on structured prompting to sequentialize tasks and guide reasoning progression. Methods like \cite{liu2023plan} use structured transitions between diverse ``X-of-Thought'' strategies. PERIA \cite{ni2024peria} combines perception, imagination, and action in a unified multimodal workflow. Others such as \cite{erdogan2025plan} explicitly target long-horizon planning through structured sequencing, while \cite{wen2024codeplan} build workflows for code-related planning.

These workflows are then grounded by a reactive controller that iteratively consumes the current state and interleaves reasoning with actions: in web automation, agents follow inspect-reason-act-observe loops \cite{yao2023react,deng2023mind2web}, with robustness improved by dynamically adapting in-context examples \cite{lutz2024wilbur}; in code, agents decide immediate executions/API calls, read outputs or errors, and refine step-by-step \cite{wang2024executable,patil2024gorilla,shinn2023reflexion,gupta2024codenav,rahman2025marco,he2024enhancing,rawat2025pre,aksitov2023rest,jiang2024self}; in robotics, monitors trigger on-the-fly safety interventions and VLM-guided subgoal execution with real-time adjustment \cite{sinha2024real,shah2023lm}. This \emph{reactive workflow} view unifies scripted stage design with online adaptation: the workflow provides interpretable structure and interfaces (what is done when), while the reactive loop supplies closed-loop grounding and error recovery (how it is done in context). The approach is broadly effective yet can accumulate errors over long horizons, motivating incremental verification and memory within the workflow to stabilize execution.

\paragraph{Tree Search / Algorithm Simulation.} Tree-based search strategies, especially BFS, DFS, A*, MCTS, and beam search, have become prominent as interpretable and effective planning scaffolds. Several works simulate tree traversal algorithms to mimic deliberative processes: \cite{yao2023tree,markowitz2024tree,long2023large,koh2024tree} apply breadth- or depth-first strategies to explore structured thought trees. A*-like guided expansions appear in \cite{wang2024qstar,meng2024llmastar,liu2024multimodal}, providing heuristic-driven planning with state evaluation. Besides that, MCTS is heavily explored in agentic research: \cite{hao2023reasoning,putta2024agent,sprueill2023monte,yu2023prompt,zhao2023large,ding2023everything,chen2024tree,kong2024latent,feng2023alphazero,yoon2025monte,schultz2024mastering,chen2025broaden} use MCTS or its variations for controlled exploration and improved reasoning fidelity. Beam search is leveraged in \cite{xie2023self,golovneva2023pathfinder,qian2025discriminator} to prune and prioritize reasoning trajectories efficiently. Other tree-search-inspired works include \cite{gandhi2024stream} which uses learned search policies and \cite{saha2024system} which differentiates between fast (reactive) and slow (deliberative) planning. These methods mirror traditional algorithmic planning, grounding LLMs' search processes in classical decision-making frameworks.

This search-over-hierarchy view maps cleanly onto domain systems. In the web setting, planner-executor architectures generate high-level subtask trees in natural language and bind leaves to DOM-grounded actions, often with memory to persist context \cite{abuelsaad2024agent,guan2023intelligent,agashe2024agent}. For code agents, hierarchical task trees and pseudo-code plans recursively break problems into compilable/editable units, while structured pipelines embed hierarchical RL or MCTS within the tree to choose promising edits and verification paths \cite{gui2025hypertree,ni2024tree,chen2025enhancing,hu2025divide,antoniades2024swe}. In robotics, behavior trees and high-level goal decomposition translate language instructions into subgoal sequences executed by low-level controllers and skills \cite{lykov2024llm,cao2023robot,izzo2024btgenbot,ahn2022icanisay,huang2022inner}. 

Taken together, hierarchical tree-search couples \emph{plan synthesis} (node expansion, heuristic/evidence-based selection) with \emph{plan realization} (leaf grounding and feedback), yielding interpretable, long-horizon agents that can backtrack, refine, and verify before committing to irreversible actions, while remaining flexible enough to incorporate learned policies and memory for efficiency and robustness.

\paragraph{Process Formalization.} Formalizing planning through symbolic representations, programming languages, or logic frameworks ensures compositionality, interpretability, and generalization. Several works encode plans as code-like artifacts or PDDL programs: \cite{guan2023leveraging,mahdavi2024leveraging,katz2024thought,wen2024codeplan,hao2024planning,vyas2024llm} incorporate symbolic logic or procedural programming into LLM prompting or output generation. These representations enable downstream tool execution and interface more cleanly with classical planners or robot controllers. PDDL-based formulations explicitly bridge LLM planning with well-established planning ecosystems, as in \cite{mahdavi2024leveraging,katz2024thought}. CodePlan \cite{wen2024codeplan} highlights the use of program synthesis to scaffold long-horizon reasoning. Such formalization provides structural scaffolds for agent behavior and often enhances explainability and robustness of the generated plans.

\paragraph{Decoupling / Decomposition.} Decoupling strategies aim to modularize complex planning into separable components such as goal recognition, memory retrieval, and plan refinement. Notably, ReWOO \cite{xu2023rewoo} explicitly separates observation and reasoning modules to optimize for efficiency. Similarly, works like \cite{zhang2025atomic, dong2024diffuserlite,lo2024goal,li2024agent,wang2023goplan,vyas2024llm,kang2025retrointext} break reasoning into reusable or hierarchical abstractions. \cite{gui2025hypertree} promotes hierarchical thinking through hypertrees, while \cite{liang2024visualpredicator} abstracts the world with symbolic predicates to reduce planning burden. Others, such as \cite{ye2024beyond} and \cite{kong2024latent}, decompose via latent variables or state spaces. These decompositions not only enhance tractability, but also align with neural-symbolic hybrid frameworks. They are especially common in long-horizon or multi-agent planning scenarios, such as \cite{zheng2024planagent,nayak2024long}.\nocite{wei2024robust,bao2025latte,chen2024wapiti,liu2025breaking,liu2024logic,liu2024class,liu2024aim,liu2023topological,zeng2025pave,zeng2024graph,lin2025moralise,lin2024backtime,qiu2025saffron,qiu2025ask,qiu2025efficient,qiu2024tucket,qiu2023reconstructing,qiu2022dimes,xu2024discrete,li2025model,zou2025transformer,qiu2024gradient,yoo2025embracing,yoo2025generalizable,yoo2024ensuring,chan2024group,wu2024fair,he2024sensitivity,wang2023networked}

\paragraph{External Aid / Tool Use.} Many systems leverage external structures or tools to aid planning, including retrieval-augmented generation (RAG), knowledge graphs, world models, and general-purpose tool use. Knowledge-augmented frameworks like \cite{chen2024plan,cornelio2025hierarchical,meng2025telograf,zhong2024flexplanner, zhang2025atomic} inject structured representations (e.g., graphs, scene layouts) into the LLM context. RAG-style systems \cite{yoo2024exploratory,li2024benchmarking, zou2025gtr} retrieve relevant knowledge to support continual instruction planning. World model-based agents such as \cite{hao2023reasoning,guan2023leveraging,qiao2024agent,zhou2024behaviorgpt,zhou2024dino,gao2024flip,liu2025continual,wang2025adawm} learn or leverage environment models for model-based planning. Tool-oriented frameworks like HuggingGPT \cite{shen2023hugginggpt}, Tool-Planner \cite{liu2025toolplanner}, and RetroInText \cite{kang2025retrointext} use external APIs or modular toolchains to support planning execution. These systems often reflect agent-environment interaction and capitalize on external resources to scaffold or augment LLM capabilities.

\subsubsection{Post-training Planning}

\paragraph{Reward Design / Optimal Control.} Finally, planning as optimization entails designing suitable reward structures and solving for optimal behavior using RL or control-theoretic tools. Reflexion \cite{shinn2023reflexion}, Reflect-then-Plan \cite{jeong2025reflect}, and Rational Decision Agents \cite{ye2023rational} incorporate utility-based learning to guide planning behavior. Reward modeling appears in works such as \cite{chen2025scaling}, while others like \cite{luyten2025strategic} emphasize reward shaping. Optimal control is tackled explicitly in \cite{ma2024non,ni2025physics,matada2024generalizable, su2025toolorchestra}, and trajectory optimization via diffusion models is seen in \cite{xie2025latent,xiao2023safediffuser,shan2025contradiff}. Offline RL methods like \cite{kong2024latent,ruoss2024amortized,wang2023goplan} leverage pretrained dynamics or cost models. The control-theoretic orientation in these works complements symbolic or heuristic approaches by optimizing over continuous, structured, or learned reward spaces.

%%%%%%%%%

\subsection{Tool-Use Optimization}

\begin{table*}[!t]
\centering
\caption{
Representative \textbf{Tool-Use Optimization} systems categorized by \textit{Integration Stage}, \textit{Learning Type}, and \textit{Tool Strategy}. 
}
\renewcommand{\arraystretch}{1.15}
\setlength{\tabcolsep}{6pt}
\begin{tabular}{p{3cm}>{\centering\arraybackslash}p{3cm}>{\centering\arraybackslash}p{4cm}>{\centering\arraybackslash}p{5.5cm}}
\toprule
\textbf{Method} & \textbf{Stage} & \textbf{Learning} & \textbf{Tool Strategy} \\
\midrule
\rowcolor[HTML]{F5F5F5}
\multicolumn{4}{l}{\textbf{Modality I: In-Context Integration}} \\
ReAct~\cite{yao2023react} & Inference & Prompting & Interleaved reasoning–action \\
ART~\cite{paranjape2023artautomaticmultistepreasoning} & Inference & Few-shot & Retrieved multi-step demos \\
ChatCoT~\cite{chen-etal-2023-chatcot} & Inference & Prompting & CoT with tool calls \\
GEAR~\cite{lu-etal-2024-gear} & Inference & Delegation & Light model for tool selection \\
AVATAR~\cite{NEURIPS2024_2db8ce96} & Inference & Contrastive & In-context tool reasoning \\
\midrule
\rowcolor[HTML]{F5F5F5}
\multicolumn{4}{l}{\textbf{Modality II: Post-Training Integration}} \\
Toolformer~\cite{schick2023toolformer} & Post-train & Self-sup. + SFT & Self-generated API calls \\
ToolLLM~\cite{DBLP:conf/iclr/QinLYZYLLCTQZHT24} & Post-train & SFT & Large-scale API demos \\
ToolAlpaca~\cite{DBLP:journals/corr/abs-2306-05301} & Post-train & SFT & Simulated dialogues \\
ReSearch~\cite{chen2025learning} & Post-train & RL + Reflec. & Adaptive retrieval reasoning \\
ReTool~\cite{dong2025reinforcement} & Post-train & RL & Reinforced code execution \\
ToolRL~\cite{qian2025toolrl} & Post-train & RL & Multi-tool policy learning \\
\midrule
\rowcolor[HTML]{F5F5F5}
\multicolumn{4}{l}{\textbf{Modality III: Orchestration-based Integration}} \\
HuggingGPT~\cite{shen2023hugginggpt} & System & Planner–Exec. & Multi-tool coordination \\
TaskMatrix.AI~\cite{DBLP:journals/corr/abs-2303-16434} & System & Planner & Massive API ecosystem \\
ToolPlanner~\cite{liu2025toolplanner} & System & RL & Plan-before-act framework \\
OctoTools~\cite{lu2025octotoolsagenticframeworkextensible} & System & Rule-based & Hierarchical orchestration \\
ToolExpNet~\cite{DBLP:conf/acl/ZhangCZCDL25} & System & Embedding & Experience-based selection \\
ToolChain*~\cite{DBLP:conf/iclr/ZhuangC0MBRS024} & System & Search & A* decision over tools \\
\bottomrule
\end{tabular}
\label{tab:tool_use_methods}
\end{table*}

Tool use optimization is the capacity of an agent to augment its intrinsic capabilities by intelligently invoking external modules. This allows agents to overcome limitations such as outdated knowledge, inability to perform precise calculations, or lack of access to private information. The core challenge lies in the agent's ability to reason about \textbf{when} to use a tool, \textbf{which} tool to select from a library, and \textbf{how} to generate a valid call. In this section, we examine existing approaches to tool use optimization, which can be broadly classified into three styles: \textit{in-context tool-integration}, \textit{post-training tool-integration}, 
and \textit{orchestration-based tool-integration}.

% \ma{refine}
\subsubsection{In-Context Tool-integration}

The in-context demonstration paradigm is a training-free approach to empowering LLMs with new capabilities at inference time. This method leverages the remarkable in-context learning ability of modern LLMs, guiding a frozen, off-the-shelf model to perform complex tasks by providing carefully crafted instructions, examples, and contextual information directly in the prompt.

% CoT based:
\paragraph{Interleaving Reasoning and Tool Use.}
The foundation of in-context agentic reasoning lies in augmenting the Chain-of-Thought (CoT) process with the ability to take action.\citep{wei2022chain}. 
{ChatCoT}~\cite{chen-etal-2023-chatcot} formalizes this paradigm by structuring reasoning traces as alternating "thought-tool-observation" steps in natural language, allowing LLMs to reflect on intermediate outputs and dynamically plan the next tool query.
While CoT enables LLMs to break down problems into intermediate reasoning steps, it operates in a closed world, limited by the model's internal knowledge. The key innovation in agentic tool use is to interleave these reasoning steps with actions (tool calls), creating a dynamic loop that allows the agent to interact with external environments to gather information and execute tasks \citep{inaba-etal-2023-multitool,trivedi2022interleaving}. 
ReAct \cite{yao2023react} introduced the "Reasoning+Acting" synergy. This approach enables the model to use reasoning to create, track, and adjust its action plans, while the actions allow it to interface with and gather information from external environments like knowledge bases or the web. 
Similarly, ART \cite{paranjape2023artautomaticmultistepreasoning} provides a structured approach by maintaining a library of successful task demonstrations. For a new task, ART retrieves a relevant multi-step exemplar and uses it as a few-shot prompt, guiding the LLM to follow a proven reasoning and tool-use path.

\paragraph{Optimizing Context for Tool Interaction.}
While the foundational interleaved loop is powerful, its performance degrades when agents must handle large or complex toolsets. A significant branch of research addresses this by optimizing the in-context information provided to the agent. Recent studies demonstrate that well-written tool documentation enables LLMs to utilize new tools in a zero-shot manner \citep{hsieh2023tooldocumentationenableszeroshot,yuan2024easytool}. This finding aligns with the key insight that LLMs, much like humans, benefit from clear and concise instructions. Alternatively, GEAR \citep{lu-etal-2024-gear} introduces a computationally efficient, training-free algorithm that delegates the tool selection process to a small language model while reserving the more powerful LLM for the final reasoning step to reduce costs.
AVATAR \cite{NEURIPS2024_2db8ce96} enhances the robustness of this choice by prompting the agent to perform in-context "contrastive reasoning" before acting.

While these in-context methods are flexible, their performance is ultimately bounded by the inherent capabilities of the frozen LLM and the length of its context window. 
Consequently, subsequent research has focused on post-training methods.

\begin{figure}
    \centering
    \includegraphics[width=\linewidth]{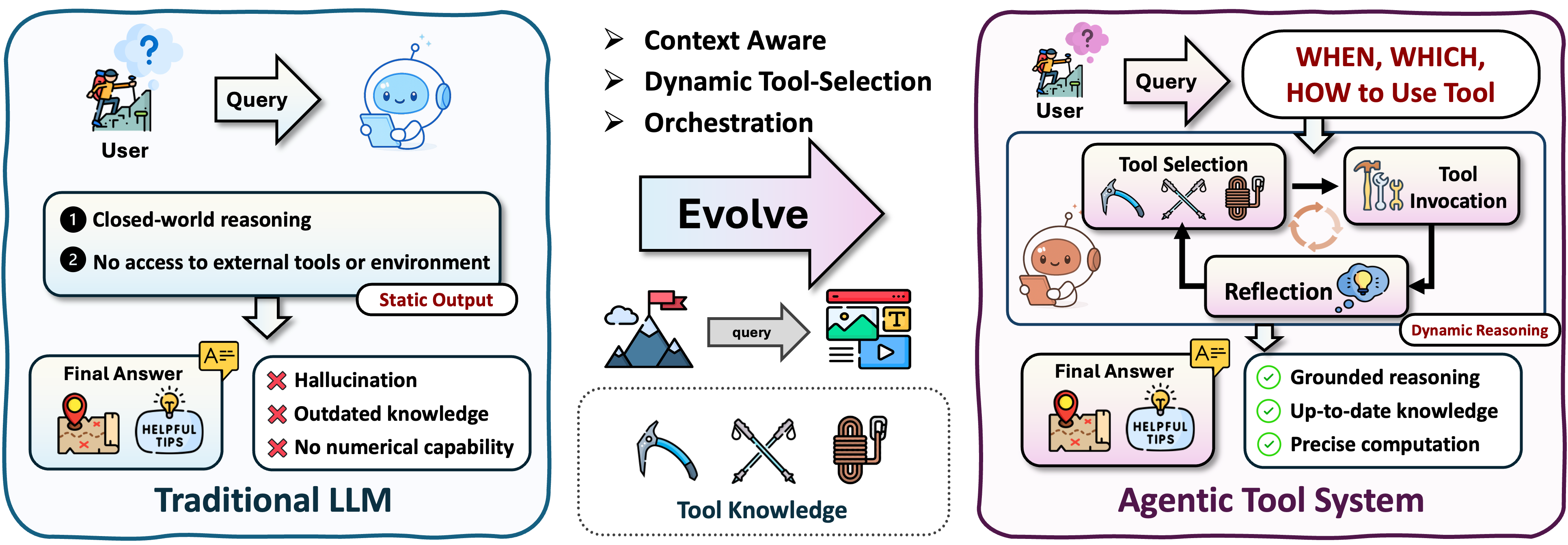}
    \caption{Comparison between \textbf{traditional LLM} and \textbf{agentic tool-use} systems. While traditional models operate in a closed world with fixed reasoning, agentic tool-use systems enable dynamic selection, orchestration, and integration of external tools, allowing agents to extend reasoning, improve precision, and dynamically adapt across domains.}
    \label{fig:tool_use}
\end{figure}

\subsubsection{Post-training Tool-integration} Tool integration \citep{yao2023react, qu2025tool,shi2025tool} with post-training techniques has emerged as a key strategy for addressing the inherent limitations of LLMs or LRMs, such as outdated knowledge, limited computational precision, and shallow multi-step reasoning. By \textit{learning how to interact with external tools}, reasoning models can dynamically access up-to-date information, execute precise symbolic or numerical computations, and decompose complex tasks into grounded, tool-assisted reasoning steps \citep{wang2024empowering,yang2024buffer,singh2025agentic, cheng2024seeclick, zou2025reasonflux}. With tools as intermediaries, models are enriched and augmented by external capabilities, enabling the generation of more accurate and generalizable agentic reasoning trajectories \citep{nam2024using,yuan2024easytool,wu2025agentic}.

\paragraph{Bootstrapping of Tool Use via SFT.} \space Early works on tool-integration \citep{yao2023react, schick2023toolformer, DBLP:conf/iclr/QinLYZYLLCTQZHT24, DBLP:journals/corr/abs-2306-05301, lu2023chameleon, song2023restgpt, prasad2023adapt, yin2023agent, shi2024learning} primarily apply supervised fine-tuning (SFT) over curated tool-use reasoning steps, where models were trained to imitate demonstrations of search queries, code executions, or API calls. The SFT stage provided an initial competency in invoking tools, interpreting tool outputs, and integrating the results into coherent reasoning chains \citep{song2023restgpt, shinn2023reflexion}. For example, 
Toolformer \cite{schick2023toolformer} introduces a self-supervised framework in which large language models generate, validate, and retain useful API calls within unlabeled text, followed by fine-tuning on the filtered data to enhance factual accuracy and practical utility. 
ToolLLM \cite{DBLP:conf/iclr/QinLYZYLLCTQZHT24} further scales SFT training to over 16,000 real-world APIs, applying supervised fine-tuning on massive curated demonstrations to endow models with robust planning and invocation abilities. ToolAlpaca \cite{DBLP:journals/corr/abs-2306-05301} extends the idea to compact LLMs by automatically constructing a diverse toolset and generating multi-turn tool-use dialogues via multi-agent simulation, followed by fine-tuning to enable generalized tool-use even for previously unseen tools. While effective at bootstrapping tool-awareness, applying SFT along suffers from overfitting to the specific patterns in the training data \cite{kirk2023understanding, li2024preserving, o2024attributing, zou2025transformer}, leading to brittle tool-selection strategies and limited adaptability in unseen downstream application scenarios \citep{zeng2025itool,qian2025toolrl,yu2025demystifying}.

\paragraph{Mastery of Tool Use via RL.} \space Recent studies \citep{zhou2025sweet, qian2025toolrl, wei2025swe, chen2025learning, zhang2025rlvmr, jin2025search, zou2025autotool, dong2025reinforcement} leverages reinforcement learning (RL) during model post-training to go beyond imitation and achieve mastery in tool-integrated reasoning. With the integration of RL, models refine their tool-use strategies through outcome-driven rewards, learning \textit{when}, \textit{how}, and \textit{which} tools to invoke via trial and error \cite{chen2025learning, feng2025retool, dong2025reinforcement, sun2025zerosearch}. For instance, SWE-RL \citep{wei2025swe} optimizes code-editing policies on large-scale software evolution data, improving not only software issue resolution but also general reasoning skills. ReSearch \cite{chen2025learning} embeds search operations into multi-hop reasoning chains, enabling adaptive retrieval during complex QA. ReTool integrates real-time code execution into reasoning rollouts, leading to optimal performance on advanced math reasoning benchmarks. ToolRL \cite{qian2025toolrl} generalizes this paradigm to diverse toolsets by introducing principled reward designs for stable and scalable multi-tool learning. Across these settings, RL has been shown to yield more robust, adaptive, and generalizable tool-use policies than SFT alone, often transferring effectively to out-of-domain tasks \cite{team2025kimi1.5, comanici2025gemini, team2025kimi2, zeng2025glm, zou2025tattoo}.

\subsubsection{Orchestration-based Tool-integration} In real-world applications, tool use within complex systems often extends beyond the single-model, single-tool setting, requiring orchestration among multiple tools to complete complex tasks. This orchestration typically involves planning, sequencing, and managing dependencies across tools, i.e., ensuring that intermediate outputs are passed and transformed appropriately. Several early works \citep{shen2023hugginggpt, DBLP:journals/corr/abs-2303-16434, DBLP:conf/nips/HaoLWH23} explore this direction by devising strategies for the coordinated use of multiple tools, enabling systems to solve multi-stage tasks that no single tool can handle in isolation. Specifically, HuggingGPT \citep{shen2023hugginggpt} employs a centralized agent that leverages a language interface to plan which tools to invoke and when, enabling the solution of complex tasks requiring multiple tools in sequence. TaskMatrix.AI \citep{DBLP:journals/corr/abs-2303-16434} connects foundation models with millions of APIs, using the models to generate task-solution outlines and automatically matching certain sub-tasks to off-the-shelf models and systems with specialized functionalities. ToolkenGPT \citep{lu2025octotoolsagenticframeworkextensible} augments frozen language models with massive tool sets by encoding each tool as a special token during next-token prediction.

\paragraph{Agentic Pipelines for Tool Orchestration.} There are many frameworks designed to enable LLMs to call and orchestrate tools effectively. Most of the current agentic paradigm follows a “plan before action” strategy, where the model first generates a structured plan for tool use and then executes it. ToolPlanner \citep{liu2025toolplanner} introduces a two-stage reinforcement learning framework with path planning and feedback, supported by MGToolBench, to bridge the gap between API-heavy training data and real-world user instructions.
Tool-MVR \citep{DBLP:journals/corr/abs-2506-04625} enhances reliability and reflection through meta-verification of tool calls and exploration-based reflection learning, achieving strong gains over GPT-4 and other baselines.
More recently, 
OctoTools \citep{lu2025octotoolsagenticframeworkextensible} provides a training-free, extensible framework with standardized tool cards, a hierarchical planner, and an executor, showing broad improvements across multi-domain reasoning tasks.
Chain-of-Tools \citep{DBLP:journals/corr/abs-2503-16779} leverages frozen LLMs’ semantic representations to dynamically compose unseen tools in chain-of-thought reasoning, enabling generalization to massive tool pools without fine-tuning.
PyVision \cite{DBLP:journals/corr/abs-2507-07998} introduces an interactive, multi-turn framework that enables MLLMs to dynamically generate, execute, and refine Python-based tools, moving beyond static toolsets in visual reasoning. ConAgents \cite{shi2024learning} makes an initial extension of tool use frameworks for interactive multi-agent settings.
We are also glad to see emerging applications of such agentic tool orchestration frameworks in the chemistry domain \citep{DBLP:journals/corr/abs-2505-02484}.

\paragraph{Tool Representations for Orchestration.} Beyond designing orchestration pipelines, another line of research focuses on optimizing the tools themselves to facilitate more accurate selection, composition, and coordination during orchestration. ToolExpNet \citep{DBLP:conf/acl/ZhangCZCDL25} models tools and their usage experiences as a network that encodes semantic similarity and dependency relations, allowing LLMs to distinguish between similar tools and account for interdependencies during selection. 
T2Agent \citep{DBLP:journals/corr/abs-2505-19768} addresses multimodal misinformation detection by representing tools with standardized templates and using Bayesian optimization to select a task-relevant subset. Coupled with Monte Carlo Tree Search over this reduced action space, T2Agent enables efficient multi-source verification.
ToolChain* \citep{DBLP:conf/iclr/ZhuangC0MBRS024} frames the entire tool action space as a decision tree and applies A* search with task-specific cost functions to guide navigation. This representation allows efficient pruning of high-cost branches and identification of optimal tool-use paths.
ToolRerank \citep{DBLP:conf/coling/ZhengLLLLW24} refines tool retrieval by introducing adaptive truncation for seen vs. unseen tools and hierarchy-aware reranking to balance concentration (for single-tool queries) and diversity (for multi-tool queries).

\subsection{Agentic Search}

Single-agent Agentic Retrieval-Augmented Generation (RAG) systems embed reasoning and control into a centralized agent that governs the entire retrieval-generation loop.
Unlike traditional RAG pipelines~\cite{lewis2020retrieval, huang2024survey, yang2024crag} that perform fixed, one-shot retrieval before generation, agentic RAG agents dynamically control \textit{when}, \textit{what}, and \textit{how} to retrieve based on real-time reasoning needs. This enables the model to adapt retrieval strategies mid-inference, refine its queries, and better integrate evidence from multiple sources. Based on how the agent selects, refines, and integrates retrieved content during reasoning, we categorize single-agent Agentic RAG systems into three distinct architectural styles: \textit{in-context}, \textit{post-training}, and \textit{structure-enhanced} agentic RAG.  

\begin{figure}
    \centering
    \includegraphics[width=\linewidth]{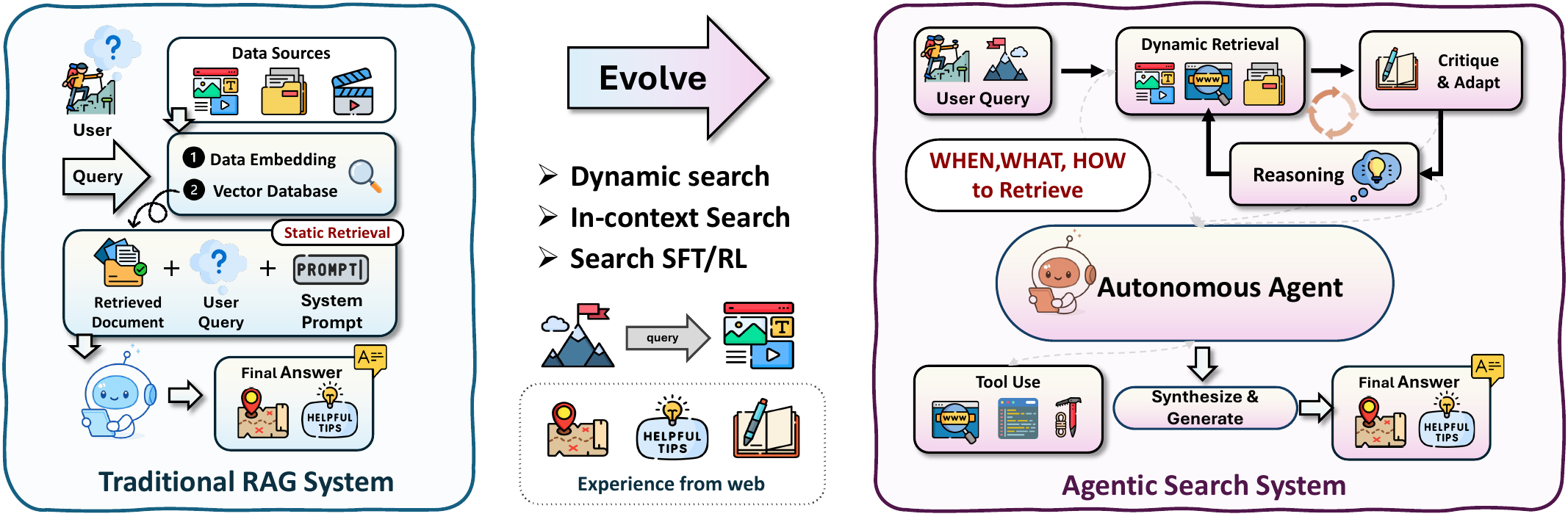}
    \caption{Comparison between \textbf{traditional RAG} systems and \textbf{agentic search} systems. Traditional RAG relies on static retrieval over a vector database, while agentic search introduces autonomous decision-making for when, what, and how to retrieve, enabling dynamic search, in-context retrieval, critique-and-adapt loops, and tool use.}
    \label{fig:agentic_rag}
\end{figure}

\begin{table*}[!t]
\centering
\caption{
Representative \textbf{Agentic Search} systems categorized by \textit{Reasoning Structure}, \textit{Format}, and \textit{Tool Use}. 
{NL} denotes natural language traces used during reasoning, {Ops} refers to symbolic or graph operations, and {KG} stands for knowledge graph. 
Tool use includes search APIs, browser actions, or KG-based retrieval.
}
\renewcommand{\arraystretch}{1.15}
\setlength{\tabcolsep}{6pt}
\begin{tabular}{p{3cm}>{\centering\arraybackslash}p{3cm}>{\centering\arraybackslash}p{4cm}>{\centering\arraybackslash}p{5.5cm}}
\toprule
\textbf{Method} & \textbf{Structure} & \textbf{Format} & \textbf{Tool} \\
\midrule
\rowcolor[HTML]{F5F5F5}
\multicolumn{4}{l}{\textbf{Modality I: In-Context Agentic Search}} \\
ReAct~\cite{yao2023react} & Interleaved & NL + Actions & Search API \\
Self-Ask~\cite{press2022measuring} & Decomposed & NL Queries & Search API \\
IRCoT~\cite{trivedi2022interleaving} & Sequential & NL + CoT & Search API \\
Self-RAG~\cite{asai2023self} & Reflective & NL Self-check & Conditional Search \\
DeepRAG~\cite{guan2025deeprag} & Iterative & NL Feedback & Search API \\
\midrule
\rowcolor[HTML]{F5F5F5}
\multicolumn{4}{l}{\textbf{Modality II: Post-Training Agentic Search}} \\
Toolformer~\cite{schick2023toolformer} & Sequential & Tool Tokens & APIs, Search \\
INTERS~\cite{zhu2024inters} & Sequential & Instructions & Search API \\
WebGPT~\cite{nakano2021webgpt} & Sequential & NL + Browser & Web Search \\
RAG-RL~\cite{huang2025rag} & Decision & NL Policy & Evidence API \\
Search\text{-}R1~\cite{jin2025search} & Iterative & NL + Tokens & Live Web \\
Deep-Researcher~\cite{zheng2025deepresearcher} & Multi-step & NL Trajectories & Browser Tools \\
ReSearch~\cite{chen2025learning} & Step-wise & NL Steps & Search + Verifier \\
ReARTeR~\cite{sun2025rearter} & Reflective & NL Policy & Tool Cluster \\
\midrule
\rowcolor[HTML]{F5F5F5}
\multicolumn{4}{l}{\textbf{Modality III: Structure-Enhanced Agentic Search}} \\
Agent\text{-}G~\cite{leeagent} & Modular & NL + Graph Ops & KG Query \\
MC-Search~\cite{ning2025mc} & Multi-step & NL & Multimodal Search \\
GeAR~\cite{shen2025gear} & Graph & Graph Ops & KG Expansion \\
ARG~\cite{zhang2025learning} & Reflective & NL + Symbols & KG Traversal \\
\bottomrule
\end{tabular}
\label{tab:agentic_search_methods}
\end{table*}

\subsubsection{In-Context Search}

\paragraph{Interleaving Reasoning and Search.}

In-context agentic RAG systems embed retrieval behavior directly into the inference process of language models through carefully designed prompting strategies. Rather than training the model to learn retrieval behavior, these methods guide it to alternate between reasoning and search within a single forward pass, typically via few-shot exemplars or special tokens. A representative example is ReAct~\cite{yao2023react}, which interleaves Chain-of-Thought reasoning with tool-use commands such as \texttt{<Search>} to dynamically invoke external APIs or knowledge sources. Extensions such as Self-Ask~\cite{press2022measuring} and IRCoT~\cite{trivedi2022interleaving} go beyond sequential reasoning by prompting the model to recursively decompose questions and retrieve sub-evidence accordingly. More recent methods~\cite{asai2023self, li2024benchmarking, guan2025deeprag, ning2025mc} introduce reflective retrieval, where the model explicitly assesses whether it needs additional information at each step, deciding to retrieve only when necessary. These approaches require no additional training, making them highly flexible and deployable, but often rely on prompt engineering and may struggle with stability across diverse domains.

\paragraph{Structure-Enhanced Search.}

Structure-enhanced agentic RAG systems enhance retrieval-augmented generation by enabling a single agent to reason over symbolic knowledge sources such as knowledge graphs through dynamic querying, tool invocation, and reflective self-monitoring. Unlike static KG retrievers or query executors, these agents decide when to access structured knowledge, how to formulate graph-based queries, and whether retrieved information suffices for continuing the reasoning trajectory. Agent-G~\cite{leeagent} introduces a modular agentic architecture that integrates unstructured document retrieval with structured graph reasoning, using feedback loops and specialized retriever modules to ensure accurate multi-hop responses. MC-Search~\cite{ning2025mc} introduces five canonical reasoning topologies to model multimodal search-enhanced reasoning process, and proposes a end-to-end agentic RAG and step-wise evaluation pipeline to evaluate model's planning and retrieval fidelity across heterogeneous sources. Similarly, GeAR~\cite{shen2025gear} incorporates graph expansion operations into an agentic controller to address challenges in complex multi-hop queries, enhancing coherence across structured and unstructured sources. Beyond retrieval orchestration, ARG~\cite{zhang2025learning} proposes a fully end-to-end agentic framework for reasoning over knowledge graphs via active self-reflection. The model autonomously determines when to retrieve, performs iterative critique based on symbolic inputs, and exhibits interpretable, step-wise reasoning behavior over graphs. Together, these systems represent a shift from passive graph access to active, feedback-driven symbolic reasoning, highlighting the potential of structured agentic RAG to achieve both factual reliability and interpretability.

\subsubsection{Post-Training Search}
Post-training agentic RAG methods endow language models with retrieval-aware capabilities by fine-tuning them to make informed decisions throughout multi-step reasoning. Unlike in-context prompting, these approaches train models, either via supervised fine-tuning (SFT) or reinforcement learning (RL), to determine when retrieval is necessary, how to formulate queries, and how to incorporate retrieved evidence.

\paragraph{SFT-Based Agentic Search.} These methods construct curated or synthetic datasets that interleave retrieval operations with natural language reasoning, and subsequently apply supervised fine-tuning to instill retrieval-aware capabilities into the model. Toolformer~\cite{schick2023toolformer} introduces a self-supervised approach to annotate tool-use behaviors within model-generated text, enabling LLMs to learn when and how to invoke tools such as web search or calculators. INTERS~\cite{zhu2024inters} extends this direction by performing instruction-based fine-tuning over a diverse, multi-task dataset compiled from over 40 sources, capturing a wide spectrum of retrieval-reasoning patterns. This class of methods benefits from scalable data generation pipelines~\cite{mao2024rag, zhang2024raft, li2025search}, which minimize the need for human annotation. Instructional reformulation techniques~\cite{lin2023ra,zhu2024inters, nguyen2024sfr} further enhance generalization by aligning tasks with human-preferred formats and reasoning.

\paragraph{RL-Based Agentic Search.} These methods optimize retrieval-aware behaviors through reward signals that reflect answer quality, factuality, or user preferences. WebGPT~\cite{nakano2021webgpt} introduces reward modeling to supervise search-augmented chains aligned with human judgment, while RAG-RL~\cite{huang2025rag} formulates retrieval as a sequential decision-making task over evidence access. More recent efforts such as Search-R1~\cite{jin2025search} and Deep-Researcher~\cite{zheng2025deepresearcher} go further by training agents to dynamically issue retrieval actions (e.g., generating \texttt{<Search>} tokens mid-reasoning) and operate in open-ended environments such as the live web. These agents exhibit emergent capabilities such as iterative decomposition, re-verification, and evidence planning. Finally, systems like ReSearch~\cite{chen2025learning} and ReARTeR~\cite{sun2025rearter} pursue not only accurate answers but also interpretable and faithful reasoning trajectories, highlighting the potential of reinforcement-learned retrievers to act as controllable and reflective agents.

\section{Self-evolving Agentic Reasoning}
\label{sec:selfevolve}

Self-evolving agentic reasoning refers to an agent’s capacity to \emph{improve its own reasoning process through experience}. 
At the core of this evolution lie two fundamental mechanisms: \textbf{feedback} and \textbf{memory}. 
\textbf{Feedback} provides evaluative signals for self-correction and refinement, allowing the agent to revise its reasoning strategies based on outcomes or environmental responses. 
\textbf{Memory}, in turn, acts as a persistent substrate for storing, organizing, and synthesizing past interactions, enabling knowledge accumulation and reuse across tasks. 
Together, these mechanisms transform reasoning from a static process into a dynamic, adaptive loop capable of continual improvement.

Building upon foundational capabilities such as \emph{planning}, \emph{search}, and \emph{tool use}, self-evolving agents integrate feedback and memory to refine their internal reasoning policies, adjust decision-making strategies, and generalize across diverse contexts, often without explicit external supervision. 
This continual adaptation marks a critical step toward lifelong reasoning and lays the groundwork for the collective intelligence explored in the next section.

\subsection{Agentic Feedback Mechanisms}

Agentic feedback mechanisms enable models to iteratively refine their reasoning and actions rather than relying on one-shot responses. By incorporating self-critique, verifier guidance, or validator-based resampling, these methods emulate human trial-and-error learning and form the foundation for autonomous self-improvement. Broadly, they operate through three distinct feedback regimes:
(1) reflective feedback, where models revise their reasoning through self-critique or verification;
(2) parametric adaptation, where feedback is consolidated into updated model parameters; and
(3) validator-driven feedback, where binary outcome signals guide resampling without introspection.

These regimes define a continuum between dynamic, inference-time adaptability, durable learning through parameter updates, and efficient correction through external signals. Together, they highlight how modern agents leverage feedback to balance flexibility, reliability, and efficiency.

\subsubsection{Reflective Feedback}

Reflective feedback methods improve model reliability by modifying the reasoning process during inference, without updating model parameters. These approaches expose intermediate reasoning outputs, such as chains of thought or partial solutions, and introduce additional assessment steps that directly influence how the model continues its generation.

Early self-critique and rationale-refinement methods \cite{shinn2023reflexion,madaan2023self} implement reflection through an explicit generate–critique–revise loop. A model first produces an answer together with its reasoning. The same model, or a separately prompted critic role, then analyzes this output to identify logical errors, unsupported assumptions, or missing steps. The critique is appended as context for a revised generation, and this process may be repeated multiple times or augmented with external evidence such as retrieval. More recent self-improvement frameworks \cite{selfimprovement2024a} extend reflective feedback beyond a single inference episode by accumulating critiques or failure cases across interactions. Instead of correcting only one response, these methods reuse past feedback to guide future generations through prompt refinement or curated supervision signals, while still operating without direct parameter updates at inference time. Search-based reasoning strategies \cite{wangself,yao2023tree,besta2024graph} improve reliability by generating and comparing multiple candidate reasoning paths. These methods explore the solution space through stochastic sampling or structured search, then select or aggregate outputs using voting schemes, heuristic scores, or learned evaluators. Improvement arises from comparison across alternatives rather than explicit revision of a single reasoning trajectory. Decomposition-based prompting methods \cite{zhou2022least,chenprogram} reformulate complex problems into ordered sequences of simpler subproblems. Intermediate results are reused in later steps, allowing partial inspection of reasoning progress and reducing error propagation, even when no explicit critique step is introduced.

Overall, reflective feedback alters inference-time reasoning trajectories by introducing additional reasoning or comparison steps. Feedback is used to guide generation within an episode, while the model’s parameters remain unchanged.

\begin{figure}
    \centering
    \includegraphics[width=\linewidth]{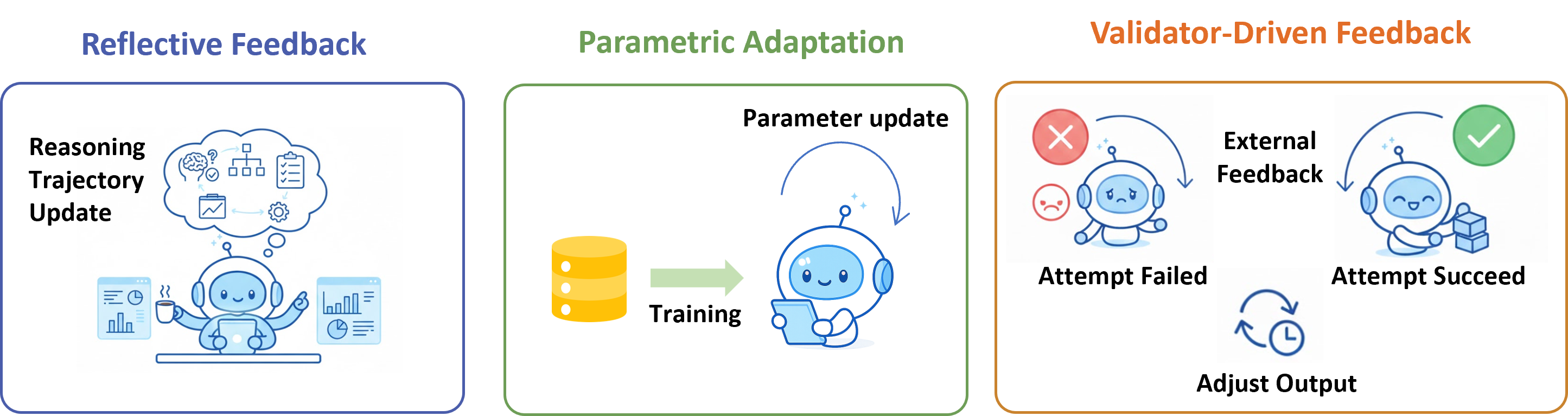}
    \caption{
    \textbf{Illustration of three forms of agentic feedback mechanisms.}
    \textit{Inference-time reflection} enables real-time self-critique and revision during reasoning;
    \textit{offline adaptation} consolidates feedback into model parameters for long-term improvement;
    and \textit{outcome-based feedback} relies on validator signals (success or failure) to refine behavior through retry.
    Together, they represent a continuum from adaptive reflection to stable learning and efficient validation.
    }
    \label{fig:agentic_feedback}
\end{figure}

\subsubsection{Parametric Adaptation}

Parametric adaptation incorporates feedback into a model’s parameters through additional training, producing persistent behavioral changes that generalize beyond individual inference episodes. Unlike reflective feedback, these methods transform feedback signals into supervised or preference-based training objectives that update the model’s weights.

Trajectory-level supervised fine-tuning approaches \cite{zeng2024agenttuning,aksitov2023rest} attach feedback to intermediate reasoning traces rather than only final answers. Models first generate multi-step trajectories, which are then reviewed by humans, auxiliary models, or automated verifiers. Incorrect steps are corrected or replaced, and the resulting feedback-enriched trajectories are used as supervised training data, encouraging the model to internalize improved reasoning patterns. Distillation-based methods \cite{hsieh2023distilling} further leverage improved reasoning traces by training student models on high-quality chains of thought or self-corrected solutions generated by stronger teachers. This process transfers structured reasoning behaviors into more stable or efficient models, removing the need for explicit reflection at inference time. Preference-alignment approaches \cite{christiano2017deep,rafailov2023dpo,bai2022constitutional} incorporate feedback in the form of comparative judgments that distinguish preferred from dispreferred outputs. Training objectives such as reward modeling or direct preference optimization adjust the model’s parameters so that preferred behaviors become more likely. Although feedback is often defined over final outputs, it implicitly shapes the internal reasoning strategies that produce them. Recent work shows that verification-augmented training data can further improve reasoning robustness across domains \cite{li2025reflectevo,zheng2025reasoningcv}. In these settings, trajectories are filtered or revised based on correctness or consistency signals before training, yielding datasets that emphasize reliable reasoning patterns.

In summary, parametric adaptation embeds feedback directly into the model’s parameters, yielding durable improvements across tasks. This durability comes at the cost of additional training and reduced flexibility compared to inference-time methods.

\subsubsection{Validator-Driven Feedback}

Validator-driven feedback improves model outputs using external success or failure signals, without modifying the model’s reasoning process or parameters. A validator, such as a unit test, constraint checker, simulator, or environment signal, evaluates candidate outputs and determines whether they satisfy predefined correctness criteria.

Retry-based systems \cite{dao2025rezero,potamitis2025retrials} implement this paradigm by repeatedly sampling candidate outputs until one passes validation. The model generates a complete solution, submits it to the validator, and discards it if validation fails. Subsequent attempts are generated independently, without conditioning on explicit information about previous failures. This strategy is particularly effective in domains with reliable and inexpensive validation, such as program synthesis and software engineering \cite{le2022coderl,ni2023lever,swebench}. Generated code can be executed against unit tests, providing an unambiguous correctness signal. The model iterates until a solution satisfies all tests, even in the absence of explicit reasoning correction. Similar mechanisms appear in embodied and interactive agents \cite{ahn2022icanisay,driess2023palm}, where action sequences are repeatedly executed until the environment signals task completion. Failed sequences are abandoned and new ones are attempted, based solely on external success signals. Some hybrid methods introduce lightweight guidance within the retry loop, for example by assigning higher reward to behaviors that eventually lead to successful outcomes \cite{bensal2025reflect}. However, the dominant mechanism remains selection through external validation rather than revision of reasoning steps or parameter updates.

Overall, validator-driven feedback offers an efficient and scalable way to improve output correctness when reliable validators are available. Its limitation is that feedback is non-diagnostic, correcting individual outputs without explaining failures or altering the model’s reasoning behavior.

\begin{table*}[!t]
\centering
\caption{Representative \textbf{Agentic Feedback Mechanisms} categorized by \textit{Feedback Stage}, \textit{Feedback Source}, and \textit{Update Target}.}
\renewcommand{\arraystretch}{1.15}
\setlength{\tabcolsep}{3pt}
\begin{tabular}{p{5.5cm}
  >{\centering\arraybackslash}p{3.8cm}
  >{\centering\arraybackslash}p{4.8cm}
  >{\centering\arraybackslash}p{3.3cm}}
\toprule
\textbf{Method / System} & \textbf{Feedback Stage} & \textbf{Feedback Source} & \textbf{Update Target} \\
\midrule

%%%%%%%%%%%%%%%%%%%%%%%%%%%%%%%%%%%%%%%%%%%%%%%%%%%%%%%%%%%%%%
% I. Reflective Feedback
%%%%%%%%%%%%%%%%%%%%%%%%%%%%%%%%%%%%%%%%%%%%%%%%%%%%%%%%%%%%%%
\rowcolor[HTML]{F5F5F5}
\multicolumn{4}{l}{\textbf{I. Reflective Feedback}} \\

Reflexion~\cite{shinn2023reflexion} & Inference & Self-generated critique & Trajectory \\
Self-Refine~\cite{madaan2023self} & Inference & Self-evaluation & Trajectory \\
Constitutional AI~\cite{bai2022constitutional} & Inference & Normative rules & Trajectory \\
RLAIF~\cite{lee2023rlaif} & Inference & AI verifier & Trajectory \\
SelfCheckGPT~\cite{manakul2023selfcheckgpt} & Inference & Cross-sample divergence & Trajectory \\
Zero-Shot Verification-CoT~\cite{chowdhury2025zeroshot} & Inference & External verifier & Trajectory \\
ASCoT~\cite{zhang2025ascot} & Inference & Vulnerability detection & Trajectory \\
MM-Verify~\cite{sun2025mmverify} & Inference & Multimodal verifier & Trajectory \\
ReAct~\cite{yao2023react} & Inference & Action outcomes & Trajectory \\
PAL~\cite{gao2023pal} & Inference & Code execution & Trajectory \\
WebGPT~\cite{nakano2021webgpt} & Inference & Web evidence & Trajectory \\
MemGPT~\cite{Packer2023MemGPTTL} & Inference & Retrieved memory & Trajectory \\
Voyager~\cite{wang2023voyager} & Inference & Environment + memory & Trajectory \\

\midrule

%%%%%%%%%%%%%%%%%%%%%%%%%%%%%%%%%%%%%%%%%%%%%%%%%%%%%%%%%%%%%%
% II. Parametric Adaptation
%%%%%%%%%%%%%%%%%%%%%%%%%%%%%%%%%%%%%%%%%%%%%%%%%%%%%%%%%%%%%%
\rowcolor[HTML]{F5F5F5}
\multicolumn{4}{l}{\textbf{II. Parametric Adaptation}} \\

AgentTuning~\cite{zeng2024agenttuning} & Training & High-quality trajectories & Model parameters \\
ReST~\cite{aksitov2023rest} & Training & Critique--revision pairs & Model parameters \\
ReFT~\cite{dou2024re} & Training & Reflection-augmented data & Model parameters \\
Distill-CoT~\cite{hsieh2023distilling} & Training & Expert CoT & Model parameters \\
ReflectEvo~\cite{li2025reflectevo} & Training & Reflection traces & Model parameters \\
Reasoning-CV~\cite{zheng2025reasoningcv} & Training & Verification signals & Model parameters \\

\midrule

%%%%%%%%%%%%%%%%%%%%%%%%%%%%%%%%%%%%%%%%%%%%%%%%%%%%%%%%%%%%%%
% III. Validator-Driven Feedback
%%%%%%%%%%%%%%%%%%%%%%%%%%%%%%%%%%%%%%%%%%%%%%%%%%%%%%%%%%%%%%
\rowcolor[HTML]{F5F5F5}
\multicolumn{4}{l}{\textbf{III. Validator-Driven Feedback}} \\

ReZero~\cite{dao2025rezero} & Inference & Binary validator & Output only \\
Retrials~\cite{potamitis2025retrials} & Inference & Acceptance signal & Output only \\
CodeRL~\cite{le2022coderl} & Inference & Unit tests & Output only \\
LEVER~\cite{ni2023lever} & Inference & Execution results & Output only \\
SWE-bench~\cite{swebench} & Inference & Test suite & Output only \\
SayCan~\cite{ahn2022icanisay} & Inference & Environment state & Output only \\
PaLM-E~\cite{driess2023palm} & Inference & Environment feedback & Output only \\
Reflect--Retry--Reward~\cite{bensal2025reflect} & Inference & Validator + reflection signal & Output only \\

\bottomrule
\end{tabular}
\label{tab:agentic_feedback_methods}
\end{table*}

\subsection{Agentic Memory}

Recent advances in memory-augmented LLM agents have shifted the focus from static memory storage to more dynamic, interactive mechanisms that directly support agentic reasoning. Rather than merely extending the context window or storing historical inputs, memory is increasingly treated as an integral component of the reasoning loop, used for reflecting on past experiences, guiding future actions, and dynamically adapting to complex, long-horizon tasks. Formally, an agent maintains a memory module where each memory entry may represent a raw observation, summarized trajectory, subgoal, tool invocation trace, or other structured element depending on the system design.

\begin{figure}
    \centering
    \includegraphics[width=\linewidth]{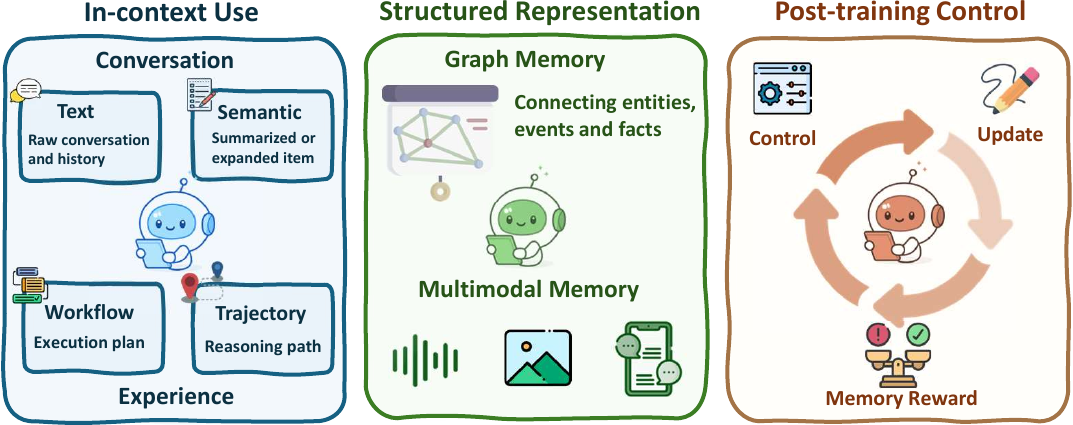}
    \caption{Overview of \textbf{Agentic Memory} in LLM agents, showing three parallel dimensions:
in-context use (text and experience), structured representation (graph and multimodal memory), and post-training control (reward-guided memory management).}
    \label{fig:agentic_memory}
\end{figure}

The agent’s reasoning process then operates not only on its immediate context but also on this persistent memory, enabling reflection, generalization, and long-term goal tracking. In this section, we organize prior work along four emerging trends in the use of memory to support and enable agentic reasoning. Figure \ref{fig:agentic_memory} summarizes how agentic memory progresses from contextual recall to adaptive control.
In-context memory captures textual and semantic information from prior interactions; structured memory integrates these into graph and multimodal representations; post-training control enables agents to evolve, update, and retrieve memory through learned reward-based mechanisms.

\begin{table*}[!t]
\centering
\caption{Representative \textbf{Agentic Memory} systems categorized by \textit{Setting}, \textit{Format}, and \textit{Memory Type}.}
\renewcommand{\arraystretch}{1.15}
\setlength{\tabcolsep}{6pt}
\begin{tabular}{p{5cm}>{\centering\arraybackslash}p{3.0cm}
  >{\centering\arraybackslash}p{3.0cm}
  >{\centering\arraybackslash}p{3.0cm}}
\toprule
\textbf{Method / System} & \textbf{Setting} & \textbf{Format} & \textbf{Memory Type} \\
\midrule

\rowcolor[HTML]{F5F5F5}
\multicolumn{4}{l}{\textbf{I. Agentic Use of Flat Memory (In-Context)}} \\

LangMem~\cite{LangChain2023} & In-Context & Text & Factual \\
LlamaIndex~\cite{Liu_LlamaIndex_2022} & In-Context & Text & Factual \\
MemGPT~\cite{Packer2023MemGPTTL} & In-Context & Text & Factual \\
MemoryBank~\cite{Zhong2023MemoryBankEL} & In-Context & Semantic & Factual \\
Amem~\cite{xu2025mem} & In-Context & Semantic & Factual \\
Workflow Memory~\cite{Wang2024AgentWM} & In-Context & Workflow & Experience \\
MemOS~\cite{li2025memos} & In-Context & Semantic & Factual \\
LightMem~\cite{fang2025lightmem} & In-Context & Semantic & Factual \\
Nemori~\cite{nan2025nemori} & In-Context & Semantic & Factual \\
ACE~\cite{zhang2025agentic} & In-Context & Workflow & Experience \\
Reasoning Bank~\cite{ouyang2025reasoningbank} & In-Context & Workflow & Experience \\
Dynamic Cheatsheet~\cite{suzgun2025dynamic} & In-Context & Trajectory & Experience \\
Sleep-time Compute~\cite{Lin2025SleeptimeCB} & In-Context & Trajectory & Experience \\
Evo-Memory~\cite{wei2025evo} & In-Context & Semantic & Experience \\
\midrule
\rowcolor[HTML]{F5F5F5}
\multicolumn{4}{l}{\textbf{II. Structured Memory Representations}} \\

GraphRAG~\cite{Edge2024FromLT} & In-Context & Graph & Factual \\
MEM0~\cite{chhikara2025mem0} & In-Context & Graph & Factual \\
Zep~\cite{rasmussen2025zep} & In-Context & Graph & Factual \\
Optimus-1~\cite{li2024optimus1} & In-Context & Multimodal & Experience \\
RAP~\cite{rap} & In-Context & Multimodal & Experience \\
M3-Agent~\cite{long2025seeing} & In-Context & Multimodal & Factual \\
Mem-Gallery~\cite{bei2026mem} & In-Context & Multimodal & Factual \\
Agent-ScanKit~\cite{cheng2025agent} & In-Context & Multimodal & Experience \\
% AutoFlow~\cite{Li2024AutoFlowAW} & In-Context & Workflow & Experience \\
% AFLOW~\cite{zhang2024aflow} & In-Context & Workflow & Experience \\
% FlowMind~\cite{Zeng2023FlowMindAW} & In-Context & Workflow & Experience \\

\midrule
\rowcolor[HTML]{F5F5F5}
\multicolumn{4}{l}{\textbf{III. Post-training Memory Control}} \\
Mem1~\cite{zhou2025mem1} & Post-training & Semantic & Factual \\
Memory-as-Action~\cite{zhang2025memory} & Post-training & Semantic & Factual \\
MemAgent~\cite{yu2025memagent} & Post-training & Semantic & Factual \\
Mem-$\alpha$~\cite{wang2025mem1} & Post-training & Semantic & Factual \\
Memory-R1~\cite{yan2025memoryr1} & Post-training & Semantic & Factual \\
Agent Early Experience~\cite{zhang2025agent2} & Post-training & Implicit & Experience \\
Agentic Memory~\cite{yu2026agentic} & Post-training & Semantic & Experience \\
MemRL~\cite{zhang2026memrl} & Post-training & Semantic & Experience \\
\bottomrule
\end{tabular}
\label{tab:agentic_memory_methods}
\end{table*}

\subsubsection{Agentic Use of Flat Memory}

\paragraph{Factual Memory.} Traditional memory systems for LLM agents typically treat memory as a passive buffer, mainly used to store dialogue histories or recent observations to address the limited context window of transformer models. Examples include dense retrieval methods~\cite{lewis2020retrieval,asaiself,Zhong2023MemoryBankEL}, pre-defined modules in LangChain and LLamaIndex~\cite{Liu_LlamaIndex_2022}, and cache-inspired designs like MemGPT~\cite{Packer2023MemGPTTL}. These approaches usually retrieve semantically similar past content to augment prompts, without influencing the agent’s internal reasoning. Enhancements such as RET-LLM with differentiable memory~\cite{Modarressi2023RETLLMTA}, SCM with controller-based mechanisms~\cite{Liang2023SCMEL}, as well as LOCOMO and LongMemEval benchmarks for long-term retention~\cite{maharana2024evaluating,wu2024longmemeval} further improve recall but remain largely static. These systems often rely on fixed heuristics and unstructured token lists~\cite{Zhong2023MemoryBankEL}, limiting adaptability for tasks involving goal decomposition~\cite{yang-etal-2025-selfgoal, zhang2025atomic}, long-term planning~\cite{zheng2024planagent}, or iterative self-improvement~\cite{Patel2024LargeLM}. In contrast, emerging agentic memory treats memory as part of the reasoning loop, supporting reflection~\cite{bo2024reflective}, and decision-making~\cite{yangyang2023finmem}. Amem~\cite{xu2025mem} enables LLM agents to autonomously generate contextual memory descriptions, build dynamic links between related experiences, and evolve memory content in response to new information. Similarly, Zep \cite{rasmussen2025zep}, Mirix \cite{wang2025mirix}, MemOS~\cite{li2025memos}, LightMem \cite{fang2025lightmem}, and Nemori \cite{nan2025nemori} leverage LLMs to automatically produce context-aware memory representations. Beyond LLM-driven approaches, recent work has explored reinforcement learning to explicitly train agents to acquire and organize factual memory, such as Mem-$\alpha$ \cite{wang2025mem1} and Memory-R1 \cite{yan2025memoryr1}, which we discuss in detail in later sections.

\paragraph{Experience Memory.} Workflow Memory~\cite{Wang2024AgentWM} tracks procedural traces to enable plan recovery and consistent reasoning. Sleep‑time Compute enables LLM agents to \textbf{pre-compute} and store anticipated reasoning steps before user interaction, effectively “thinking offline” using memory as a preparatory resource~\cite{Lin2025SleeptimeCB}. Dynamic Cheatsheet (DC)~\cite{suzgun2025dynamic} equips black-box models with external memory to store reusable strategies, reducing redundant reasoning. Other efforts explore complementary paradigms of agentic memory. In parallel, workflow memory has emerged as another structured approach, particularly suited for procedural and tool-augmented tasks. It explicitly tracks procedural traces during execution, supporting plan recovery, long-term consistency, and interpretable chaining of actions. Atomic reasoning~\cite{zhang2025atomic} proposes a structured trace over 
a finite set of reusable atomic skills in a streamlined generation space to reduce spurious reasoning patterns. Context evolution (ACE)~\cite{zhang2025agentic} treats contexts as evolving playbooks rather than building a static structured store, whereas Reasoning Bank~\cite{ouyang2025reasoningbank} focuses on reusing failed reasoning traces to enhance future task performance. Evo-Memory~\cite{wei2025evo} synthesizes these ideas by benchmarking self-evolving memory under streaming task settings, highlighting experience reuse as a central capability for stateful, long-horizon agentic reasoning. In addition to factual memory, Mirix~\cite{wang2025mirix} further introduces a procedural memory component to capture reusable action patterns, while Agentic Memory~\cite{yu2026agentic} and MemRL~\cite{zhang2026memrl} adopt reinforcement learning to optimize the acquisition and management of experiential memory.

This marks a shift from static buffers toward structured, reasoning-centric memory architectures. In these agentic memory systems, memory serves as a dynamically growing context: agents not only record past actions but actively \textbf{reflect, edit, and refine} their strategy over time.

\subsubsection{Structured Use of Memory}

Beyond flat memory usage and control, the structure of memory plays a critical role in enabling complex reasoning. Recent work increasingly explores structured representations, such as semantic graphs, workflows, and hierarchical trees, often extended to multimodal settings, to better capture dependencies, and contextual relationships.

Graph-based representations provide a flexible substrate for organizing relational knowledge in agents \cite{Ouyang2024RepoGraphEA}. GraphRAG~\cite{Edge2024FromLT} serves as a foundational technique that augments retrieval with graph-structured reasoning, enabling more contextually coherent and multi-hop information integration. Building on this foundation, agent systems such as MEM0~\cite{chhikara2025mem0} and Zep~\cite{rasmussen2025zep} organize memory explicitly as dynamic knowledge graphs, allowing agents to store, retrieve, and reason over entities, attributes, and their relations with improved efficiency and semantic grounding. Beyond graphs, structured memory has also been explored through alternative organizational forms. MemTree~\cite{rezazadeh2024isolated} leverages a dynamic tree-structured representation to hierarchically organize and integrate information, while workflow-oriented systems such as AutoFlow~\cite{Li2024AutoFlowAW}, AFLOW~\cite{zhang2024aflow}, and FlowMind~\cite{Zeng2023FlowMindAW} represent reasoning workflows explicitly in memory, capturing sequences of subgoals, tool invocations, and decision points.

New benchmarks have pushed reasoning memory into multimodal domains, where agents are required to ground, retrieve, and reuse information across heterogeneous modalities. M3-Agent \cite{long2025seeing} evaluates visual–audio–text reasoning through “see, listen, and reason,” while Agent-Scankit \cite{cheng2025agent} proposes multimodal agents with integrated memory modules for adaptive retrieval and grounding. Optimus-1~\cite{li2024optimus1} proposes a hybrid multimodal memory architecture that represents world knowledge as a hierarchical directed knowledge graph and abstracts past interactions into a multimodal experience pool. RAP~\cite{rap} retrieves relevant experiences based on contextual similarity, enabling adaptive reuse of multimodal memory.

These structured memory formats align task semantics, temporal dependencies, and multimodal signals, enabling agents to reason compositionally and maintain coherent behavior over extended interactions. As task complexity increases, the abstraction and organization of memory become increasingly critical for building robust and generalist agents.

\subsubsection{Post-training Memory Control}
Conversely, memory systems can also be controlled by the agent's reasoning process itself. Rather than relying on fixed heuristics for reading and writing memory, recent work has explored agent-controllable memory operations, where the agent explicitly decides what to store, when to retrieve, and how to interact with memory. This reframes memory as a policy target, no longer a passive buffer, but a resource that is actively shaped by reasoning.

MemAgent~\cite{yu2025memagent} formulates memory overwrite as a reinforcement learning problem: the agent is rewarded for preserving information that proves useful and for discarding irrelevant content. By using a newly proposed DAPO algorithm, the model learns to maintain a constant-sized memory across conversations while maximizing future utility. Mem1~\cite{zhou2025mem1} presents an end-to-end reinforcement learning framework where agents maintain a compact, shared internal state across turns, jointly supporting reasoning and memory consolidation. Memory-R1 \cite{yan2025memoryr1} further advances this line by introducing a dual-agent design: a Memory Manager that dynamically decides when to add, update, or delete entries in the memory store, and an Answer Agent that distills the most relevant retrieved memories to guide response generation. Recent work such as Mem-$\alpha$ \cite{wang2025mem1} also explores RL-based control of multi-component memory construction in agents, providing a unified perspective on adaptive memory construction and reasoning control. Memory-as-Action~\cite{zhang2025memory} integrates memory editing including insertions, deletions, and modifications directly into the reasoning policy, proposing a Dynamic Context Policy Optimization algorithm to handle non-prefix trajectory changes caused by memory operations. Agent Learning via Early Experience~\cite{zhang2025agent2} further relaxes reward dependence by enabling agents to learn from their own interaction traces through self-prediction and reflection, bridging imitation and reinforcement learning. Moreover, Agentic Memory~\cite{yu2026agentic} and MemRL~\cite{zhang2026memrl} adopt reinforcement learning to optimize the acquisition and management of experiential memory.

Together, these systems mark a shift toward \textbf{learning-based memory control}, where memory usage is optimized through reinforcement or imitation learning. By integrating memory management into the reasoning policy, agents become more adaptive, scalable, and capable of long-horizon decision-making in dynamic environments.

\subsection{Evolving Foundational Agentic Capabilities}

\begin{figure}
    \centering
    \includegraphics[width=0.8\linewidth]{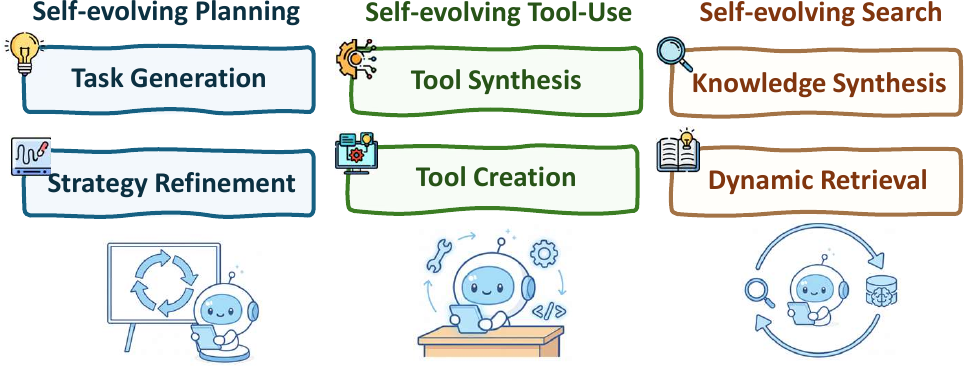}
    \caption{An overview of evolving foundational agentic capabilities along three key dimensions: \emph{planning} (task generation and strategy refinement), \emph{tool-use} (tool creation and synthesis), and \emph{search} (dynamic retrieval and knowledge synthesis). These dimensions reflect how agentic systems autonomously enhance their reasoning and problem-solving capacity over time.}
    \label{fig:evolve}
\end{figure}

\subsubsection{Self-evolving Planning}

Recent advances view planning not as a fixed reasoning routine but as an evolving capability. Instead of relying on static datasets or human-designed curricula, agents can autonomously generate tasks, learn from their own feedback, and adapt strategies through iterative interaction with the environment. This enables continuous improvement without external supervision.

A representative direction is self-generated task construction. 
For example, SCA enables agents to alternate between generating problems and solving them, reusing successful trajectories for fine-tuning \cite{zhou2025selfchallenginglanguagemodelagents}. 
Self-rewarding frameworks further allow agents to assess their own outputs, producing high-quality training signals without human labels \cite{simonds2025selfrewardingselfimproving,simonds2025self}. 
Other works directly leverage execution feedback for online adaptation, such as SELF, SCoRe, PAG, TextGrad, and AutoRule, which transform natural-language critiques or traces into training rewards, enabling continual policy refinement \cite{lu2023self,kumar2024training,yuksekgonul2024textgrad,wang2025autorule}.

Beyond internal feedback, agents can also evolve through environment shaping. 
AgentGen constructs adaptive environments to induce curriculum learning \cite{hu2025agentgen}, while Reflexion and AdaPlanner use self-reflective or adaptive strategies to refine plans at runtime \cite{shinn2023reflexion,sun2023adaplanner}. 
Self-Refine iteratively critiques and improves outputs \cite{madaan2023self}, and SICA allows self-modification of code and reasoning tools \cite{robeyns2025self}. 
From an RL perspective, RAGEN and DYSTIL model planning as a Markov Decision Process and optimize strategies with dense feedback \cite{wang2025ragen,wang2025dystil}.

Together, these methods establish a self-improving planning loop, where agents generate their own tasks, shape their environments, and refine strategies, laying the groundwork for autonomous, open-ended planning evolution.

\subsubsection{Self-evolving Tool-use}

\paragraph{Creating and Synthesizing Tools.}
The culmination of in-context reasoning is the emergent capability of agents to autonomously create new tools. This is achieved not through training, but by prompting a frozen LLM to act as a programmer when it encounters a problem that its existing toolset cannot solve.
The LATM framework \cite{cai2024large} uses a powerful model as a one-time "tool maker" and a cheaper, lightweight model as a frequent "tool user," thus amortizing the cost of creation. 
To enable specialization beyond the limits of general-purpose APIs, frameworks like CRAFT \cite{yuan2024craft} and CREATOR \cite{qian-etal-2023-creator} generate custom tools tailored for specific domains. Taking this a step further, ToolMaker \cite{wölflein2025llmagentsmakingagent} can convert entire public code repositories into usable tools, allowing agents to leverage complex, human-written codebases on the fly.

\subsubsection{Self-evolving Search}

Search plays a central role in agentic reasoning, enabling models to retrieve, select, and synthesize relevant knowledge across large and evolving memory spaces. In early systems, search was typically static—built on fixed retrieval heuristics or similarity-based dense retrievers~\cite{lewis2020retrieval,asai2023self,Zhong2023MemoryBankEL,Packer2023MemGPTTL}. These methods augmented prompts with retrieved information but lacked adaptive control over how memory evolves or how search strategies are improved over time.

Recent research increasingly links search and memory in a \textbf{co-evolutionary loop}: agents continuously update their \emph{memory base} during task execution, while dynamically adjusting how search is performed over this evolving knowledge. Agentic memory systems such as MemGPT~\cite{Packer2023MemGPTTL}, MemoryBank~\cite{Zhong2023MemoryBankEL}, and Workflow Memory~\cite{Wang2024AgentWM} already highlight how retrieved information can be synthesized and re-inserted into memory, gradually improving retrieval quality. Dynamic Cheatsheet (DC)~\cite{suzgun2025dynamic} further demonstrates how reusable strategies can be accumulated and leveraged across queries, effectively transforming static search into a \emph{living retrieval substrate} that evolves with agent experience.

\paragraph{Evolving Memory Bases.}  
Unlike static index-based retrieval, self-evolving agents actively refine their memory base through reflection and post-execution updates. Reflexion~\cite{shinn2023reflexion} allows agents to critique their own reasoning traces and store distilled insights, improving future search relevance. Reasoning Bank~\cite{ouyang2025reasoningbank} and context evolution methods~\cite{zhang2025agentic} explicitly restructure memory representations to align retrieval results with evolving problem-solving strategies, effectively making the retrieval target itself adaptive over time.

\paragraph{Dynamic Search and Synthesis.}  
Beyond memory updates, search strategies themselves can evolve through dynamic prioritization and synthesis. Structured memory representations—such as workflows~\cite{Li2024AutoFlowAW,zhang2024aflow,Zeng2023FlowMindAW} and knowledge graphs~\cite{Ouyang2024RepoGraphEA,Edge2024FromLT,chhikara2025mem0,rasmussen2025zep}—provide semantic scaffolding that enables multi-hop and compositional search, supporting richer reasoning over longer horizons. Systems like MemOS~\cite{li2025memos} and Memory-as-Action~\cite{zhang2025memory} take this further by integrating search decisions directly into the reasoning policy, allowing retrieval targets, strategies, and sources to co-adapt as agents accumulate experience.

Overall, self-evolving search transforms retrieval from a static utility into a continuously adapting component of the reasoning loop. By evolving memory bases, dynamically adjusting search strategies, and synthesizing retrieval results into structured knowledge, agents can maintain more relevant, structured, and actionable information over extended time horizons.

\section{Collective Multi-agent Reasoning}
\label{sec:collective}

Building upon the single-agent foundation, where reasoning supports planning, search, and tool use within a unified perception–action loop, \textbf{multi-agent reasoning} extends these principles to collaborative settings. 
In a multi-agent system (MAS), multiple reasoning agents interact to jointly solve complex tasks. 
Rather than identical problem solvers, agents assume \emph{complementary roles}, such as \textit{Manager} for task decomposition, \textit{Worker} for execution, and \textit{Verifier} for evaluation, enabling specialization and division of cognitive labor. 
This role differentiation marks the first step toward collective intelligence, where reasoning is distributed and coordinated across multiple agents.

Beyond role assignment, the essence of multi-agent reasoning lies in how these agents \emph{collaborate, communicate, and co-evolve}. 
Collaboration schemas define how reasoning traces are exchanged, conflicts are resolved, and shared memory is maintained to achieve alignment. 
Through such interaction, reasoning transitions from an individual process into a distributed, iterative loop, in which agents refine each other’s outputs and collectively converge toward better solutions.

Compared with single-agent systems, multi-agent reasoning introduces new challenges that require rethinking reasoning at the system level:
\begin{itemize}
    \item \textbf{Role differentiation:} how to design static or adaptive roles that align with task structure and expertise distribution;
    \item \textbf{Collaboration and communication:} how agents exchange intermediate reasoning, negotiate consensus, and divide labor efficiently;
    \item \textbf{Collective memory and evolution:} how shared or distributed state supports long-term coordination and continual adaptation.
\end{itemize}

These challenges motivate the following structure of our analysis.
Section~\textbf{5.1} examines the \emph{role taxonomy} of multi-agent systems, from generic organizational roles to domain-specific specializations. 
Section~\textbf{5.2} focuses on \emph{collaboration and division of labor}, including in-context and post-training coordination strategies. 
Finally, Section~\textbf{5.3} explores how \emph{memory} enables multi-agent systems to evolve over time and maintain collective consistency.
Together, these perspectives provide a unified view of how reasoning scales from individual agents to adaptive, collaborative intelligence.

\subsection{Role Taxonomy of Multi-Agent Systems (MAS)}
In this subsection, we first summarize the generic roles that often appear in a multi-agent system (MAS). Then, we introduce the specific functions of different roles when an MAS is applied in different domains, such as software engineering, finance, legal activities, education, healthcare, biomedicine, and music applications. 

\begin{figure}[h]
    \centering
    \includegraphics[width=0.8\linewidth]{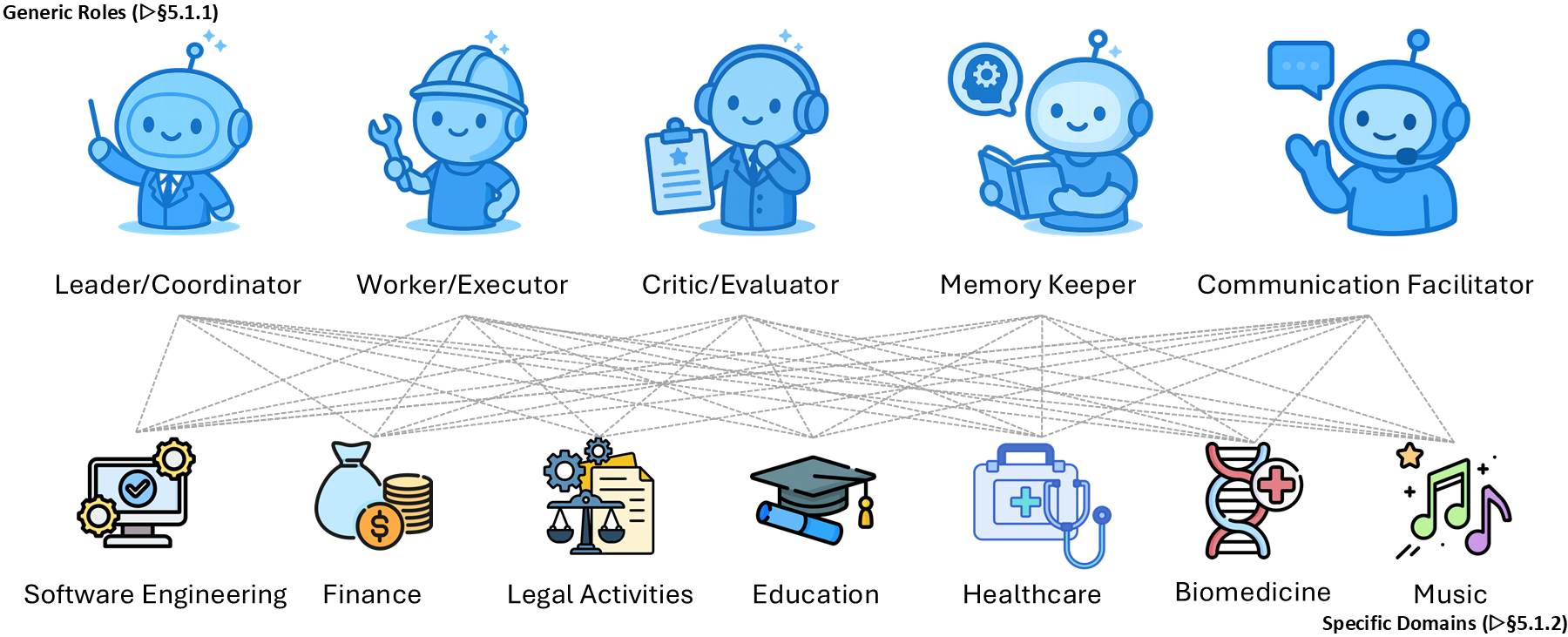}
    \caption{An overview of generic roles of agent and their specific domain adaptations in Section 5.1.}
    \label{fig:agentic_role}
\end{figure}

\subsubsection{Generic Roles}
\begin{itemize}[leftmargin=1.2em]
\item \textbf{Leader/Coordinator}: 
The leader, or coordinator, is responsible for maintaining high-level coherence within the system. 
This role involves setting global objectives, decomposing tasks into manageable subgoals, and assigning them to appropriate agents.
In addition, the leader arbitrates conflicts that emerge between agents with overlapping or contradictory outputs. 
In practice, this role often manifests itself as a meta-controller that monitors the progress of other agents and ensures that execution adheres to an overarching plan.
% sets goals, decomposes tasks, assigns roles, arbitrates conflicts.
\item \textbf{Worker/Executor}: 
Executors, often called workers, are the operational backbone of MAS. They engage in concrete actions such as invoking external tools, writing or executing code, retrieving documents, or interfacing with the environment. 
Although they typically act under the directives of a leader, well-designed systems allow for adaptive autonomy, 
where executors can refine or optimize their assigned tasks when new local information becomes available.
% performs tool calls, coding, retrieval, environment interaction.
\item \textbf{Critic/Evaluator}: 
The critic/evaluator role centers on quality assurance. This role includes verifying correctness, testing hypotheses, red-teaming responses, and surfacing potential risks. In LLM-based systems, this often corresponds to \emph{LLM-as-a-judge} setups, where dedicated evaluators assess the factuality, safety, or stylistic alignment of output. Critic roles help introduce checks and balances into otherwise generative workflows, thereby mitigating error propagation.
% verifies, tests, judges, or red-teams outputs; can be an \emph{LLM-as-a-judge}.
\item \textbf{Memory Keeper}: 
Effective MAS requires persistent memory to accumulate context, prevent repetitive failures, and enable learning across episodes. 
The memory keeper curates and maintains long-term knowledge structures such as episodic logs, semantic embeddings, retrieval indices, or knowledge graphs. 
By abstracting memory management into a dedicated role, the system can better balance short-term reactivity with long-term continuity and adaptation.
% curates episodic/semantic memory, retrieval indices, knowledge graphs.
\item \textbf{Communication Facilitator}:
Communication overhead can easily undermine MAS efficiency. 
This role governs protocols for inter-agent exchange, including defining message schemas, managing communication bandwidth, enforcing gating mechanisms, and orchestrating consensus-building. 
By reducing ambiguity and ensuring structured information flow, the communication facilitator prevents bottlenecks and coordination failures in large-scale or heterogeneous agent populations.
\end{itemize}

\subsubsection{Domain-Specific Roles}
Beyond generic agent roles, domain-specific tasks often require specialized functions. 
These roles reflect professional practices in particular industries and map naturally onto MAS architectures.

\textbf{Software Engineering}: 
In software engineering, MAS generally maps onto roles that mirror the software development lifecycle: \underline{\emph{architects}}, \underline{\emph{developers}}, \underline{\emph{code reviewers/testers}}, \underline{\emph{CI orchestrators}}, and \underline{\emph{release managers}}~\cite{hong2024metagpt,ChatDev}. 
The rationale is to distribute the responsibilities in a way that balances creativity, verification, automation, and governance, just as in industrial software practice.
\begin{itemize}
    \item Architects define system-level design principles and establish structural blueprints. 
    \item Developers translate these abstractions into concrete implementations. 
    \item Code reviewers and testers safeguard reliability, checking correctness, maintainability, and functional coverage. 
    \item CI orchestrators automate builds, testing, and artifact pipelines, reducing integration frictions. 
    \item Finally, release managers oversee deployment, aligning new versions with milestones and safety protocols.
\end{itemize}

Previous work has demonstrated similar mappings, such as MetaGPT~\cite{hong2024metagpt}, which decomposes development into Product Manager, Architect, and Engineer agents. ChatDev~\cite{ChatDev} further emphasizes communicative collaboration among specialized agents to support requirement analysis, coding, and testing. More recently, self-evolving collaboration networks have expanded this paradigm by enabling MAS to dynamically reorganize and optimize their roles throughout the software lifecycle~\cite{hu2024self}. A variant of MAS is also applied to the High-Performance Computing (HPC) domain~\cite{pauloski2025empowering}
By structuring MAS around these stages, the architecture gains the same robustness and scalability as professional engineering workflows.

\noindent \textbf{Finance}: 
The financial domain can be roughly decomposed into four archetypal roles: \underline{\emph{analysts}}, \underline{\emph{risk managers}}, \underline{\emph{traders/execution agents}}, and \underline{\emph{compliance officers}}~\cite{Chen2012, Hull2018}. This division reflects the established institutional design of financial organizations, where the responsibilities are segmented to balance profit generation with systemic stability.
\begin{itemize}
    \item Analysts operate at different levels (e.g., fundamental, sentiment, or technical), each extracting distinct signals from raw market or textual data.
    \item Risk Managers then monitor portfolio exposure, apply stress tests, and enforce safeguards to prevent cascading vulnerabilities. 
    \item Traders take responsibility for market interaction, while Execution agents ensure that orders are placed with speed and efficiency under liquidity constraints.
    \item Finally, Compliance roles ensure that activities remain aligned with regulatory requirements, enabling traceable decision-making and proper oversight.
\end{itemize}
Together, this layered ecology mirrors real-world financial institutions, where specialization and checks-and-balances are indispensable.
Recent advances in MAS for finance mirror this layered ecology. R\&D-Agent-Quant~\cite{R&D-Agent-Quant} demonstrates how agents can specialize in factor discovery and joint optimization for quantitative strategies. FinRobot~\cite{yang2024finrobot} provides an open source multi-agent platform tailored to financial applications, reflecting the practical need for modularity and scalability. PEER~\cite{wang2024peer} introduces expertization and tuning methods to adapt MAS to domain-specific responsibilities, while FinCon~\cite{finconf} highlights the role of conceptual verbal reinforcement to enhance decision-making and compliance. Together, these works underscore how MAS can replicate the specialization, checks, and balances of real-world financial institutions.

\textbf{Legal Activities}:
Multi-agent systems are also designed to model the collaborative and adversarial processes inherent in legal practice, with roles assigned to manage consultation, reasoning, and argumentation.
\begin{itemize}
    \item For legal consultation, frameworks often simulate a law firm's structure with a \underline{\emph{receptionist agent}} for client intake, specialized \underline{\emph{lawyer agents}} for providing advice, a \underline{\emph{secretary agent}} for documentation, and a \underline{\emph{boss agent}} for quality control. In a consultation model, the \emph{receptionist agent} first clarifies a user's query before routing it to the appropriate \emph{lawyer agent}. After the multi-turn consultation, the \emph{secretary agent} summarizes the interaction, and the \emph{boss agent} provides an evaluation, ensuring a comprehensive and high-quality service~\cite{sun2024lawluo}.
    \item For statutory reasoning, tasks are decomposed between \underline{\emph{knowledge acquisition agents}} that interpret legal texts and \underline{\emph{knowledge application agents}} that apply formalized rules to case facts. To be specific, in reasoning systems, the \emph{knowledge acquisition agent} first builds a reusable ontology from legal statutes; then, the \emph{knowledge application agent} uses this formal structure to analyze the specifics of a new case, ensuring consistent and transparent logic~\cite{sadowski2025verifiable}.
    \item To simulate courtroom dynamics, roles such as \underline{\emph{judge}}, \underline{\emph{plaintiff}}, \underline{\emph{defendant}}, and adversarial \underline{\emph{lawyer agents}} are created~\cite{chen2024agentcourt}. In courtroom simulations, adversarial \emph{lawyer agents} engage in debate before a \emph{judge agent}, reflecting on their performance after each trial to iteratively improve their argumentation strategies by updating their internal knowledge bases~\cite{chen2024agentcourt}.
\end{itemize}

\textbf{Education}:
In education, MAS is being developed to provide personalized and adaptive learning experiences by distributing pedagogical functions among specialized agents.
\begin{itemize}
    \item For personalized tutoring, a central \underline{\emph{tutor agent}} might engage a student using Socratic dialogue, while a \underline{\emph{memory dispatcher}} agent tracks the student's progress and misconceptions to adapt the difficulty and focus of the lesson in real-time~\cite{chudziak2025ai}.
    \item For curriculum design, a pipeline of agents collaborates: a \underline{\emph{research agent}} gathers relevant information, a \underline{\emph{planning agent}} structures it into a coherent course, and other agents generate specific learning activities or assessments. Also, it can be modeled by an adversarial process, where \underline{\emph{evaluator agent}} critiques a lesson plan created by \underline{\emph{generator agent}}, and \underline{\emph{optimizer agent}} refines it based on feedback~\cite{zhang2025eduplanner}. 
\end{itemize}

These systems demonstrate a shift towards creating intelligent, adaptive platforms that can support educators and provide students with more effective, engaging, and individualized learning journeys.

\textbf{Healthcare}:
In the healthcare domain, multi-agent systems are structured to mirror clinical and research workflows, distributing complex tasks among specialized AI agents.

\begin{itemize}
    \item For clinical diagnostics and consultation, these roles often include a \underline{\emph{triage agent}} (or \emph{moderator}) for initial case assessment, various \underline{\emph{specialist agents}} (e.g., pathologists, neurologists), a \underline{\emph{doctor agent}} for patient interaction, and a \underline{\emph{measurement agent}} to provide test results~\cite{kim2024mdagents, ghezloo2025pathfinder}. To be more specific, in the diagnostic setting, a \emph{triage agent} first assesses the complexity of a case and routes it to the appropriate \emph{specialist agents} for analysis. These specialists may then engage in multiround discussions, with a \emph{lead physician} agent synthesizing their opinions to reach a consensus. In addition to that, a \emph{doctor agent} conducts a multi-turn dialogue with a \emph{patient agent}, requesting specific data from a \emph{measurement agent} to gather information dynamically.
    \item For autonomous research, roles are modeled after the scientific process, featuring a \underline{\emph{meta agent}} for strategic planning, an \underline{\emph{executor}} for running analyses, an \underline{\emph{evaluator}} for assessing outcomes, and a \underline{\emph{reflector}} for synthesizing knowledge~\cite{zhu2025healthflow}. This division of labor allows for a systematic and comprehensive approach to multifaceted health challenges. Especially, the \emph{meta agent} plans an experiment, the \emph{executor} carries it out, the \emph{evaluator} provides immediate feedback, and the \emph{reflector} distills successful strategies into a persistent knowledge base, creating a self-improving cycle that enhances future planning.
    \item For public health events, ShortageSim~\cite{cui2025shortagesim} models FDA regulators, manufacturers, and healthcare buyers interacting under information asymmetry, enables counterfactual policy testing and evaluates how announcements and disruptions shape investment, stockpiling, and resolution timing against historical trajectories.
\end{itemize}

Other frameworks such as MMedAgent and MedAgent-Pro focus on orchestrating specialized medical tools, using a central agent to plan actions and aggregate results from various tool-based agents to handle multimodal data~\cite{li2024mmedagent, wang2025medagent}. 

\textbf{Biomedicine}:
In biomedicine, particularly in drug and material discovery, MAS is designed to automate and accelerate the scientific process by assigning roles that reflect the iterative cycle of design, testing, and refinement.

For de novo molecule design, key roles include the \underline{\emph{actor}} (or \emph{reasoner}) for generating novel structures, the \underline{\emph{evaluator}} for assessing chemical properties, and the \underline{\emph{self-reflector}} for refining future hypotheses based on results.
To be specific, the \emph{actor} agent proposes new candidates, which are then passed to the \emph{evaluator} agent. The evaluator uses computational chemistry tools to calculate properties like binding affinity and synthetic accessibility, providing quantitative feedback~\cite{averly2025liddia}. This feedback is then analyzed by the \emph{self-reflector} agent to update the system's strategy for the next generation cycle, creating a feedback-driven process of optimization~\cite{OSDA}.

Similarly, LIDDIA acts as a "digital chemist" with a \emph{Reasoner}, \emph{Executor}, \emph{Evaluator}, and \emph{Memory} component to navigate the drug discovery process and balance the exploration of new chemical spaces with the exploitation of promising candidates~\cite{averly2025liddia}. To streamline the creation of machine learning workflows, DrugAgent uses an \emph{LLM Planner} and an \emph{LLM Instructor} to automate programming for tasks like ADMET prediction~\cite{liu2024drugagent}. In genomics, GenoMAS orchestrates six specialized agents through a guided-planning framework to analyze complex gene expression data, integrating the reliability of structured workflows with the adaptability of autonomous agents~\cite{liu2025genomas}.

\textbf{Music}: In the creative domain of music composition, MAS is being explored to decompose the intricate process of creating music into collaborative, specialized roles. A system like ComposerX might feature a \underline{\emph{conductor agent}} that interprets a high-level user prompt and oversees the project, a \underline{\emph{melody agent}} that generates primary musical themes, a \underline{\emph{harmony agent}} that creates supporting chord progressions, and a \underline{\emph{rhythm agent}} that lays down the percussive and temporal foundation. These agents would interact iteratively, with the conductor agent synthesizing their outputs and providing feedback to ensure the different musical layers are coherent and aligned with the initial creative vision. This mirrors the collaborative process of a human orchestra or band, distributing creative responsibilities to achieve a complex and harmonious final product~\cite{ComposerX}.

\subsection{Collaboration and Division of Labor}

\begin{figure}
    \centering
    \resizebox{\linewidth}{!}{
    \includegraphics[width=\linewidth]{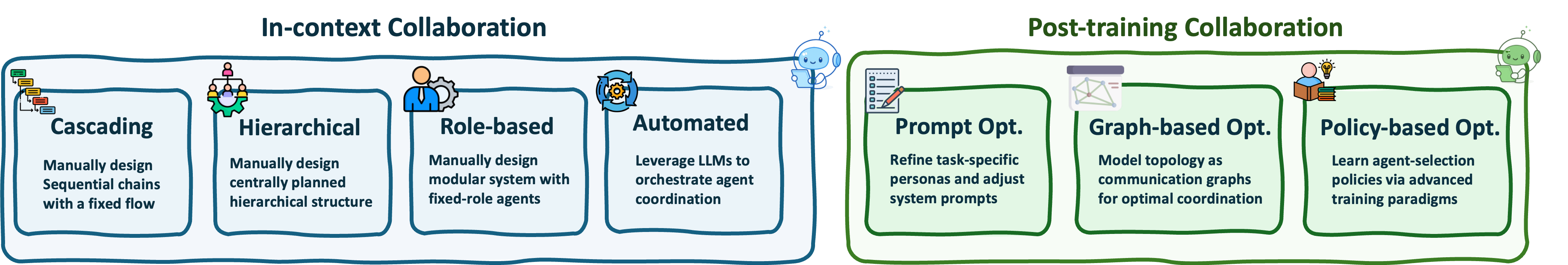}
    }
    \caption{An overview of \textbf{Agentic Collaboration} in the multi-agent system, containing two parallel dimensions: in-context collaboration (training-free task-specific coordination design) and post-training collaboration (optimization-based automated workflow generation). }
    \label{fig:/multi_agent_collab}
\end{figure}

Collaboration and division of labor constitute a central organizing principle in modern multi-agent systems. Instead of treating agents as homogeneous components, recent work emphasizes how responsibilities are decomposed and coordinated across specialized agents to improve efficiency and robustness. From this perspective, existing approaches can be broadly organized along two dimensions. \textbf{In-context collaboration} focuses on coordination strategies that are specified or induced at inference time without additional training. \textbf{Post-training collaboration} instead optimizes agent roles, interaction structures, or routing policies through learning or search. In addition, \textbf{agentic routing} can be viewed as a special case of this division of labor, where routing decisions explicitly offload cognition and computation to different agents based on task demands.

\subsubsection{In-context Collaboration}
In the design of multi-agent systems, several studies have observed that leveraging task-specific in-context information is often sufficient to build highly effective systems without the need for explicit training. Among these works, one line of research relies on manually crafted pipelines, where researchers design the agent interactions and workflows tailored to the target task. In contrast, another line explores LLM-driven automatic pipeline generation, allowing the model itself to construct and adapt the system’s structure dynamically based on the task context.

\paragraph{Manually Crafted Pipelines.} 

These approaches rely on predefined hierarchies or fixed collaboration workflows, where agent roles, execution order, and communication rules are determined before execution. Hierarchical systems such as AgentOrchestra \cite{agentorchestra}, MetaGPT \cite{hong2024metagpt}, and SurgRAW \cite{low2025surgraw} feature a central planner or conductor directing subordinate agents through structured subgoals. Cascading pipelines like Collab-RAG \cite{xu2025collab}, MA-RAG \cite{marag}, Chain of Agents \cite{zhang2024chain}, and AutoAgents \cite{chen2023autoagents} process information sequentially, passing intermediate outputs downstream with limited revision. Modular role-decomposed frameworks such as RAG-KG-IL \cite{ragkgil}, SMoA \cite{smoa}, and MDocAgent \cite{han2025mdocagentmultimodalmultiagentframework} define fixed functional roles (e.g., retriever, reasoner, or vision agent) but allow minimal dynamic coordination. While these manually designed pipelines offer interpretability, modularity, and low execution complexity, their rigidity restricts adaptability to ambiguous or evolving reasoning tasks, motivating more flexible, reasoning-driven coordination mechanisms.

\paragraph{LLM-Driven Pipelines.}

This category leverages LLMs as orchestrators that decompose high-level goals into subgoals, route them to role-specialized agents or tools, and iteratively refine workflows based on intermediate feedback until completion. AutoML-Agent \cite{trirat2024automl} proposes a full-pipeline, orchestrator-led agent team that plans, assigns, and coordinates web/API/code tools through role-specialized micro-agents (e.g., coder/tester/runner), enabling end-to-end software development workflows. Magentic-One \cite{fourney2024magentic} introduces a generalist multi-agent system where a central Orchestrator plans, tracks progress, and performs ledger-based routing over specialized agents (WebSurfer, FileSurfer, Coder, ComputerTerminal), achieving competitive results on GAIA, AssistantBench, and WebArena. MAS-GPT \cite{ye2025mas} trains an LLM to emit executable MAS code conditioned on a user query, so a single forward pass generates a query-specific multi-agent workflow. MetaAgent \cite{zhang2025metaagent} presents a finite-state-machine (FSM) abstraction to declare states, transitions, and tools, from which a LLM designer automatically constructs the MAS pipeline. AOP \cite{li2024agent} formalizes orchestrator responsibilities and introduces three design principles, i.e., solvability, completeness, non-redundancy, and then operationalizes them with fast decomposition/assignment plus a reward-model evaluator.

\emph{Agent Routing.} Closely related to LLM-driven orchestration, a line of work explicitly models \textbf{agent routing} as a decision layer that selects appropriate specialists for each query or subtask. For example, AgentRouter \cite{zhang2025agentrouter} proposes a knowledge-graph-guided router that leverages structured task semantics to dispatch questions to relevant agents, enabling effective collaborative question answering without modifying individual agents. Similarly, Talk to Right Specialists \cite{wu2025talk} frames routing and planning as a unified inference-time process, where a controller dynamically assigns subtasks to domain-specialized agents based on intermediate reasoning states. These approaches highlight that agentic routing itself can be viewed as an inference-time realization of division of labor, where cognition is selectively offloaded to specialized agents.

\paragraph{Theory-of-Mind-Augmented Collaboration.}

Another interesting line of research is Theory of Mind (ToM), which refers to the ability of an agent to infer and reason about the beliefs, intentions, and mental states of other agents. \citet{li2023theory} first showed that equipping LLM agents with explicit belief-state representations in a cooperative text game improves both collaboration performance and the accuracy over ToM-free LLM baselines. Building on this, Hypothetical Minds~\cite{cross2024hypothetical} scaffolds ToM as a modular hypothesis-generation and refinement loop for other agents’ strategies, while MindForge~\cite{licua2024mindforge} extends ToM-aware reasoning to embodied collaborative learning. In parallel, \citet{wu2025large} provides a mechanistic account of how LLMs encode ToM, identifying sparse parameter patterns whose perturbation selectively disrupts social reasoning. Pushing toward, ToM-agent~\cite{yang2025large} augments LLM generative agents with counterfactual reflection over counterparts’ beliefs and BeliefNet~\cite{sagara2025beliefnest} offers a ToM-centric joint-action simulator where embodied agents act based on nested belief states.

\subsubsection{Post-training Collaboration}

In multi-agent systems, the design of agent prompts (or personas) and the interaction topology plays a critical role in determining the system’s ability to solve complex tasks. Recently, optimizing these components during the post-training phase has emerged as an important research direction. Based on the optimization objective, existing studies can be broadly categorized into two lines of work: prompt optimization and topology optimization.

\textbf{Multi-agent Prompt Optimization.} Prompt optimization in multi-agent systems focuses on how agent roles, workflows, and feedback are encoded in prompts to yield reliable coordination and stronger task performance. For example, AutoAgents \cite{chen2023autoagents} extends prompt optimization from single-agent contexts to multi-agent teams, refining role specialization and execution plans through structured dialogue among meta-agents. SPP~\cite{unleashing2024} introduces a cognitive synergist that dynamically selects multiple personas during multi-agent collaboration for knowledge-intensive and reasoning-intensive tasks, enabling complementary expertise to emerge. DSPy Assertions \cite{singhvi2023dspy} introduces LM Assertions that can be either hard (Assert) or soft (Suggest). When violated, these assertions trigger backtracking and prompt revision using erroneous outputs and error traces. During compilation, the mechanism bootstraps examples and counterexamples to reinforce few-shot prompts, which improves both recall and accuracy. MASS \cite{zhou2025multi} demonstrates that prompts are often the dominant factor in MAS performance, and further applies automatic prompt optimization \cite{pryzant2023automatic} by incorporating local and global topology information to refine each agent’s prompt in a fine-grained manner.

As for topology optimization, two categories of research have emerged, each pursuing relatively independent optimization pathways. The first category of work treats the multi-agent topology as a communication graph, leveraging graph-based methods to identify an optimal structure that achieves strong performance under constrained communication costs (i.e., limited graph size). The second category adopts a policy-based perspective, where variable training paradigms are employed to learn an agent-selection policy with specially designed rewards or supervision signals. Through iterative, policy-based selection of subsequent agents, these approaches aim to progressively construct topologies that yield optimal overall performance. We discuss these two categories of approaches in greater detail in the following paragraphs.

\paragraph{Graph-based Topology Generation.} 
A large body of work models multi-agent systems (MAS) as graphs where agents are nodes, and inter-agent communication forms edges.  Then MAS design becomes a problem of learning the communication/coordination topology. These works could be roughly divided into three groups as follows. 

\emph{Graph generation.}
These methods aim to construct communication topologies from scratch by adaptively generating task-conditioned graphs. GommFormer \cite{hu2024learning} uses an encoder-decoder framework to learn the communication graph via continuous relaxation of the graph representation, optimizing topology end-to-end under bandwidth constraints. G-designer \cite{zhang2024g} starts from a task-anchored network with a virtual task node, then uses a variational graph auto-encoder to decode a query-adaptive communication graph. MCGD \cite{zenggraph} builds a sparse coordination graph with continuous node and discrete edge attributes, and performs categorical diffusion on edges and anisotropic diffusion on actions to capture structure diversity.

\emph{Graph pruning.}
These works start from dense collaboration graphs and aim to prune them into compact, task-appropriate pipelines while preserving utility and lowering token and compute costs.  For example, AgentPrune \cite{zhang2024cut} first formulates the MAS problem as a spatial-temporal graph sparsification problem, and then applies one-shot magnitude pruning to learn a sparse and effective pipeline. AGP \cite{li2025adaptive} learns a dual-pruning policy, i.e., soft-pruning on edges and hard-pruning on nodes, to acquire a per-query topology. G-Safeguard~\cite{wang2025g} introduces pruning as a security mechanism. It treats communication edges as the search space, employs a graph neural network to identify risky nodes, and applies deterministic rules to prune their outward edges based on a model-driven threshold, thereby defending the system against adversarial attacks. 

\emph{Topology search.}
This line of research explores the graph space by searching over agentic operators and communication edges to identify effective pipelines. Specifically, AFlow \cite{zhang2024aflow} automates multi-agent workflow design with Monte-Carlo Tree Search over a fixed library of operators. MASS pre-defines some influential graph motifs, such as debating and tool-using, and then implements topology search inside this pruned motif subset. Then MASS \cite{zhou2025multi} performs a prompt search on that topology to maximize performance. MaAS \cite{zhang2025multi} replaces single-graph search with a probabilistic “agentic supernet” over layered operator choices and uses a controller to sample a query-conditioned subgraph.  DynaSwarm \cite{leong2025dynaswarm} broadens the design space from a single optimized communication graph to a portfolio of candidate structures. It employs Actor–Critic (A2C) optimization to refine this portfolio and introduces a lightweight graph selector that chooses the most suitable topology for each instance. GPTSwarm \cite{zhuge2024gptswarm} formulates the search space as inter-agent connections within a computational graph. It relaxes the discrete topology into continuous edge probabilities and leverages reinforcement learning to optimize the resulting connection schemes, thereby enabling flexible and adaptive graph structures.

\paragraph{Policy-based Topology Generation.}

A growing line of research strengthens multi-agent pipeline generation by learning the policy of selecting subsequent agents with advanced training paradigms such as supervised fine-tuning (SFT), and reinforcement learning (RL). These approaches embed auxiliary signals into the optimization process, enabling agents to acquire stronger reasoning skills and more reliable coordination. Routing can be viewed as a special case of collaboration, in which a router conditions on task state and system context to learn a policy for selecting agents that maximize efficiency and performance \cite{yue2025masrouter,liu2025rcr,qian2025xrouter,wang2025optimal}. Broadly, these methods can be grouped into three categories based on the signal type they inject into learning.

\emph{Relative-advantage policy learning.}
Several approaches rely on critic-free objectives to form advantages, thereby avoiding centralized value models and providing effective guidance to optimize policy. For example,  MAGRPO \cite{liu2025llm} proposes a Dec-POMDP formulation for LLM collaboration and replaces centralized critics with a group-relative advantage signal, enabling decentralized training/execution at dialog-turn  granularity. MHGPO \cite{chen2025heterogeneous} extends GRPO-style signals to heterogeneous groups: it jointly optimizes different agent roles via a shared group-relative objective, and introduces practical sampling/optimization tweaks. COPY \cite{ma2024coevolving} utilizes two-agent co-training framework with shared rewards and KL regularization (to a frozen ref and cross-agent policies), improving stability and transfer between pioneer/observer roles on reasoning tasks.

\emph{LLM-generated prior guidance.}
Other methods leverage LLMs to generate rewards or priors for learning. Specifically, LGC-MARL \cite{jia2025enhancing} uses an LLM to propose a Reward Function Generator (RFG) that turns natural-language objectives into structured reward terms. LAMARL \cite{zhu2025lamarl} lets an LLM synthesize a prior policy and a task-specific reward function, then fine-tunes agents with RL. MAPoRL \cite{park2025maporl} defines rewards as weighted sums of LLM verifier scores on current and future turns, then updates policies with multi-agent PPO. COPPER \cite{bo2024reflective} learns a shared reflector with a counterfactual-PPO pipeline in which a learned reward model scores each agent’s reflection by its marginal contribution to task improvement. SIRIUS \cite{zhao2025sirius} builds an experience library by retaining trajectories that lead to successful outcomes and augmenting failures, while a Judgment–Critic–Actor triad supplies LLM-generated correctness signals that filter and supervise subsequent fine-tuning across reasoning tasks. Multiagent Finetuning \cite{subramaniam2025multiagent} bootstraps reasoning by running multi-agent debates among generator LLMs and using LLM critics plus majority voting to produce self-generated supervisory signals, then fine-tunes role-specialized agents on critic-selected trajectories to improve both accuracy and diversity.

\emph{Human preference signals.}
This line of research replaces or augments environment rewards with human-derived feedback to align behavior with human intent, in both online and offline regimes. For instance, M3HF \cite{wang2025m3hf} organizes human input into multi-phase feedback (e.g., scalar ratings, pairwise comparisons, and natural-language rationales) processed by LLMs into reward shaping signals. O-MAPL \cite{bui2025mapl} introduces an end-to-end preference-based learning framework and directly learns Q-values from offline preference data, bypassing the two-stage reward-model-then-RL pipeline.

\subsection{Multi-Agent Evolution}

While self-evolving agents enable individual models to continuously improve through interaction and feedback, many real-world applications require collective intelligence supported by cooperation among multiple agents. Therefore, recent studies extend self-evolution from single-agent settings including planning, tool-use, and search evolution~\cite{shinn2023reflexion, sun2023adaplanner, madaan2023self, wang2025ragen, wang2023voyager} to multi-agent co-evolution, where adaptation emerges across distributed agents~\cite{zhang2024aflow, 10.1145/2970276.2970311, wan2025rema, feng2025group, ke2025mas}. Beyond evolving model parameters, memory, prompts, and tools~\cite{chhikara2025mem0, zhao2024expel, khattab2024dspy, qiu2025alita}, multi-agent evolution further targets shared memory, communication mechanisms, and collaboration protocols~\cite{zhou2025multi, ke2025mas, Wang2024AgentWM}. 

As a result, multi-agent memory must jointly evolve along architecture, topology, content, and management dimensions, supported by hierarchy-structured, role-aware architectures \cite{zhang2025gmemory}, governed and distributed storage topologies \cite{xu2025sedm, rezazadeh2025collaborative}, modular and task-structured memory contents \cite{wang2025mirix, han2025legomem}, and active management mechanisms for compression, verification, and continual updating \cite{kaiya2023lyfe, tang2025agent} to ensure coherent and scalable collaboration.

The goal thus shifts from optimizing a single agent’s capability to improving the collective performance of multiple agents on complex, long-horizon tasks~\cite{zhang2024aflow, 10.1145/2970276.2970311, wan2025rema,novikov2025alphaevolve}.

\subsubsection{From Single-Agent Evolution to Multi-Agent Evolution}

While the shift from single-agent evolution to multi-agent co-evolution broadens the spatial dimension of adaptation from an individual model to a collective, the temporal dimension of evolution remains equally crucial. Beyond determining who evolves (a single agent or a population), recent studies also investigate when and how fast agents should adapt during interaction. This perspective leads to a complementary axis of analysis that distinguishes short-horizon, within-episode updates from long-term, cross-episode improvements, commonly referred to as intra-test-time evolution and inter-test-time evolution. We summarize these temporal modes of self-evolving behavior.

Intra-test-time evolution refers to the ability of agents to adapt and improve during task execution, enabling them to correct failures and refine strategies on the fly when facing unseen states or unexpected feedback. Unlike static inference pipelines, this paradigm embeds self-reflection, dynamic planning, memory rewriting, or even localized fine-tuning into the execution loop. Representative works leverage natural-language self-critique~\cite{shinn2023reflexion, madaan2023self} and runtime adaptive planning~\cite{sun2023adaplanner, hua2024trustagent} to generate corrective signals without external supervision. Reflexion~\cite{shinn2023reflexion} allows agents to store distilled reflective feedback for immediate behavior improvement, while AdaPlanner~\cite{sun2023adaplanner} dynamically revises and replans mid-trajectory based on environmental mismatch detection. Beyond contextual adaptation, methods such as test-time supervised updating~\cite{zweiger2025self} and test-time reinforcement learning (TTRL)~\cite{zuo2025ttrl, simonds2025ladder} directly modify model behavior when encountering difficult cases, often through problem-variant generation and targeted optimization. These approaches demonstrate that performance at inference time can improve within a single episode, forming short-horizon adaptation loops where the model learns while solving, rather than merely executing a fixed policy.

Inter-test-time evolution extends the self-improving process to across-task learning, where adaptations made in one task can be consolidated and transferred to future tasks. This enables the accumulation of persistent, generalizable capabilities over a lifelong interaction stream. A prominent paradigm involves offline self-distillation, where the agent generates responses and then refines them via self-evaluation before using them for supervised fine-tuning-such as in SELF~\cite{lu2023self}, STaR~\cite{zelikman2022star}, and Quiet-STaR~\cite{ferrag2025reasoning}. These methods turn incorrect initial reasoning into high-quality labeled data for future performance gains. Additionally, online reinforcement learning frameworks such as RAGEN~\cite{wang2025ragen} and DYSTIL~\cite{wang2025dystil}  continuously update policies based on dense interaction feedback, allowing agents to gradually internalize complex decision-making strategies over long horizons. Inter-test-time evolution can also incorporate curriculum mechanisms that automatically adjust task difficulty and environment complexity~\cite{qi2024webrl, lee2024animals}, as well as experience structuring via memory evolution to preserve accumulated reasoning heuristics~\cite{liang2025sage, salama2025meminsight, Wang2024AgentWM}. This temporal mode focuses on stable long-term improvement, transforming short-lived corrections from individual tasks into continual competence growth across diverse task distributions.

To support these new capabilities, mechanisms evolve from individual reward-based or reflective adaptation~\cite{lu2023self, kumar2024training, yuksekgonul2024textgrad} to multi-agent reinforcement learning and game-theoretic co-optimization~\cite{wan2025rema, feng2025group}, enabling collaborative structures to self-organize under evolving task requirements. Moreover, memory-driven multi-agent evolution (e.g., shared workflow memory or knowledge graphs) helps maintain accumulative group intelligence across episodes~\cite{Wang2024AgentWM, li2025memos}.
Overall, multi-agent evolution transforms isolated self-improvement loops into adaptive intelligent ecosystems capable of self-correction, self-organization, and social learning. This transition marks a critical step toward artificial collective intelligence, where cooperative dynamics drive continuous progress beyond the capabilities of any individual agent~\cite{zhou2025multi, zhang2024aflow, 10.1145/2970276.2970311, latentmas}.

\begin{figure}
    \centering
    \resizebox{0.6\linewidth}{!}{
    \includegraphics[width=\linewidth]{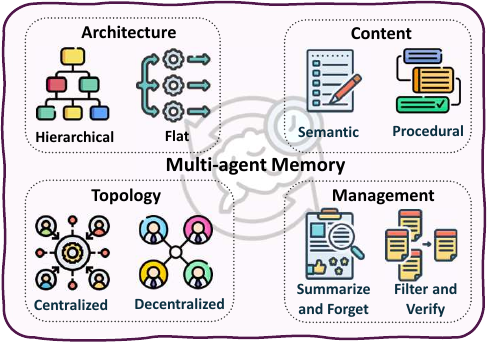}
    }
    \caption{\textbf{Four dimensions of multi-agent memory design.} 
    The framework includes (1) \textbf{Architecture}, how memory is structured; 
    (2) \textbf{Topology}, where it is stored and shared; 
    (3) \textbf{Content}, what type of knowledge is stored; and 
    (4) \textbf{Management}, how it is maintained and updated.}
    \label{fig:multi_agent_memory}
\end{figure}

\subsubsection{Multi-agent Memory Management for Evolution}

Multi-agent LLM systems pose unique challenges for memory design compared with single-agent settings. 
Beyond maintaining an individual agent's local context, they must capture inter-agent interactions, track roles and dependencies over time, and preserve both shared and private knowledge coherently. 
Memory must also remain scalable as collaboration grows and interactions accumulate. 
To provide a clearer understanding of this landscape, \textbf{we categorize existing approaches along four key dimensions}: 
(1) \textit{architecture}, how memory is organized within and across agents; 
(2) \textit{topology}, whether it is centralized, distributed, or hybrid; 
(3) \textit{content}, the type and structure of stored knowledge; and 
(4) \textit{management}, how memory is written, retrieved, and updated over time. Illustrations are shown in Figure \ref{fig:multi_agent_memory}.

% ============================================================
% DIMENSION 1: ARCHITECTURE
% ============================================================

\paragraph{Architecture Dimension: Hierarchical and Heterogeneous Designs.}

Recent work highlighted that prevailing multi-agent memory mechanisms were overly simplistic and lacked per-agent customization~\cite{zhang2025gmemory}. To address this, G-Memory constructs a three-tier graph hierarchy (insight, query, interaction graphs) that separates high-level generalizable insights from fine-grained execution traces. This hierarchical approach enables bi-directional memory traversal for retrieving both abstract lessons and concrete precedents across episodes. However, instead of global aggregation, Intrinsic Memory Agents adopts an opposing strategy by maintaining dedicated role-aligned memory templates for each agent~\cite{yuen2025intrinsic}. This heterogeneous approach preserves specialized perspectives on collaborative planning benchmarks by reducing irrelevant information per agent. Recent work further explores hybrid strategies, with some systems employing adaptive hierarchical knowledge graphs in decentralized architectures that allow agents to reason over past interactions and share only relevant information rather than raw experiences~\cite{yang2025damcs}. These contrasting approaches reveal a fundamental trade-off: hierarchical designs optimize for global coherence and cross-episode learning, while heterogeneous designs optimize for role fidelity and computational efficiency.

% ============================================================
% DIMENSION 2: TOPOLOGY & GOVERNANCE
% ============================================================

\paragraph{Storage Topology and Memory Governance.}

Systems employ different topologies to balance scalability, privacy, and coherence, each reflecting different assumptions about trust and coordination. SEDM (Self-Evolving Distributed Memory)~\cite{xu2025sedm} tackles memory management by turning memory into an active, self-optimizing component through verifiable write admission (via reproducible replay) and utility-based consolidation. This centralized approach with verification gates ensures that only factual or useful information enters the repository and performs cross-domain knowledge diffusion to enable transfer across heterogeneous tasks. In contrast, when privacy and organizational boundaries matter, Collaborative Memory ~\cite{rezazadeh2025collaborative} distinguishes private versus shared memory fragments using bipartite graph policies. Every entry carries immutable provenance (source agent, accessed resources, timestamp), enabling compliance auditing and safe cross-agent knowledge transfer in federated systems. At the other end of the spectrum, some systems like Memory Sharing~\cite{gao2024memory} adopt uncontrolled pooling where all agents freely exchange experiences in a shared memory pool. Research shows that memory sharing among LLM agents leads to a more diverse collective memory pool, which improved performance on open-ended tasks by creating emergent collective intelligence. These three topologies represent increasing levels of formality and control, reflecting different priorities for managing the trade-off between knowledge diversity and verification rigor.

% ============================================================
% DIMENSION 3: GRANULARITY
% ============================================================

\paragraph{Memory Content: Semantic, Task, and Cognitive-Phase Decomposition.}

Different content decomposition strategies suit different task characteristics, and the choice of content structure fundamentally shapes how agents interact with memory. 
MIRIX~\cite{wang2025mirix} pioneered \textit{semantic decomposition} by defining six specialized memory types (Core, Episodic, Semantic, Procedural, Resource, Knowledge Vault) managed by distinct agents, achieving a 35\% accuracy gain on multimodal QA tasks while reducing storage through flexible routing. 
Building on this modular principle, LEGOMem~\cite{han2025legomem} instead employs \textit{task-based decomposition}, breaking execution traces into reusable memory units flexibly assigned to either central planners or specialist task agents. 
This design shows that orchestrator memory improves task decomposition and delegation, while agent memory enhances subtask execution, effectively narrowing performance gaps between small and large LLM teams. 
Recently, MAPLE introduced Cognitive-phase Decomposition~\cite{bai2025maple}, using specialized agents (Solver, Checker, Reflector, Archiver) to enable systematic error detection and plan repair cycles. 
The Reflector diagnoses errors after each episode, and the Archiver stores refined plans to avoid repeated mistakes, supporting feedback-driven learning. 
These three content decomposition strategies reveal that memory design should align with task structure: semantic content for heterogeneous information, task-based for workflow automation, and cognitive-phase for error-sensitive reasoning.

% ============================================================
% DIMENSION 4: MANAGEMENT
% ============================================================

\paragraph{Memory Management Strategies.}

Effective long-term memory requires active management balancing relevance, efficiency, and coherence through different approaches that trade off simplicity against sophistication. Lyfe Agents~\cite{kaiya2023lyfe} pioneered the forgetting-based approach using Summarize-and-Forget mechanisms to regularly compress memory, retaining only critical context. This strategy is suitable when storage is severely constrained, though it risks losing nuanced details for edge cases. To improve upon simple forgetting, AGENT-KB~\cite{tang2025agent} introduced more sophisticated management by organizing procedural traces into structured (entity, action, observation) triples and learning pattern abstractions reusable across tasks. Agents collaborate to retrieve, update, and reason over memory segments, enabling generalization without explicit retraining while central coordination ensures long-term consistency for scalable embodied planning. The choice among these strategies depends on system priorities: forgetting prioritizes storage efficiency, verification prioritizes reliability, and learning-based approaches prioritize adaptability. Production systems typically combine strategies, e.g., verification for critical memories and forgetting for low-utility peripheral information, to balance multiple objectives.

% ============================================================
% FUTURE DIRECTIONS
% ============================================================

\paragraph{Discussions.}

Despite substantial progress, multi-agent memory systems remain largely unexplored with respect to post-training and model adaptation. Current approaches focus primarily on memory organization and retrieval for pre-trained models, with little investigation into how multiple agents can jointly optimize their memories through post-training procedures such as reinforcement learning or supervised fine-tuning. This represents a notable gap: while post-training techniques have been actively explored for single-agent memory systems, extending them to enable multi-agent teams to co-evolve their memory structures and management policies remains an open problem.

\subsubsection{Training Multi-agent to Evolve}

Recent advancements have shifted multi-agent systems from fixed, hand-designed coordination toward training paradigms that enable agents to evolve over time \cite{ma2024coevolving, multi-agent-evolve, zhao2025sirius}. Training multi-agent systems to evolve represents a critical step toward realizing adaptive, long-horizon intelligence beyond static coordination. In this emerging paradigm, agents improve collectively through interaction, feedback, and shared memory, rather than isolated or independently optimized behaviors. By embedding reasoning into the learning loop, via reinforcement learning \cite{marft}, self-play \cite{strongermas}, curriculum evolution \cite{park2025maporl}, and verifier-driven feedback \cite{marlhf}, multi-agent systems can internalize coordination strategies, address inter-agent credit assignment, and progressively refine divisions of labor. This evolution transforms multi-agent reasoning from a static ensemble of cooperating LLMs into a self-improving organization that adapts its structure, communication patterns, and policies in response to task complexity and environmental change \cite{zhao2023chatenvironmentinteractivemultimodal}.

\paragraph{Co-evolution via Interaction and Intrinsic Feedback.}
A growing body of work has operationalized multi-agent evolution through explicit training objectives that couple interaction, feedback, and role specialization. For instance, Multi-Agent Evolve \cite{multi-agent-evolve} instantiates a closed-loop co-evolution framework containing three interacting roles (\emph{Proposer}, \emph{Solver}, and \emph{Judge}), all of which are derived from a shared LLM backbone and jointly optimized via reinforcement learning. This forms a self-improving curriculum that enables collective skill growth without external supervision. In a related spirit, CoMAS \cite{CoMAS} emphasizes intrinsic interaction rewards, extracting learning signals directly from multi-agent discussion dynamics through an LLM-based judge, thereby enabling decentralized co-evolution driven purely by collaborative interaction.

\paragraph{Multi-Agent Reinforcement Fine-Tuning for Collective Adaptation.}
Additional works have focused on principled reinforcement fine-tuning frameworks tailored to LLM-absed multi-agent systems. For example, MARFT \cite{marft} formalizes multi-agent reinforcement fine-tuning by highlighting key mismatches between classical MARL assumptions and LLM-based agent organizations, such as role heterogeneity, dynamic coordination, and long-horizon dialogue, and provides a systematic framework for stabilizing collective post-training. Stronger-MAS \cite{strongermas} further adapts on-policy reinforcement learning to multi-role, multi-turn settings by introducing agent- and turn-wise grouping strategies that extend GRPO-style optimization, enabling more effective coordination learning across complex agent workflows. Similarly, MAPoRL \cite{park2025maporl} proposes multi-agent post-co-training, where multiple LLMs are jointly optimized using a collaboration-aware verifier that rewards not only final outcomes but also the quality of intermediate discussions, encouraging the emergence of transferable communication strategies.

\begin{figure}[!t]
    \centering
    \includegraphics[width=\linewidth]{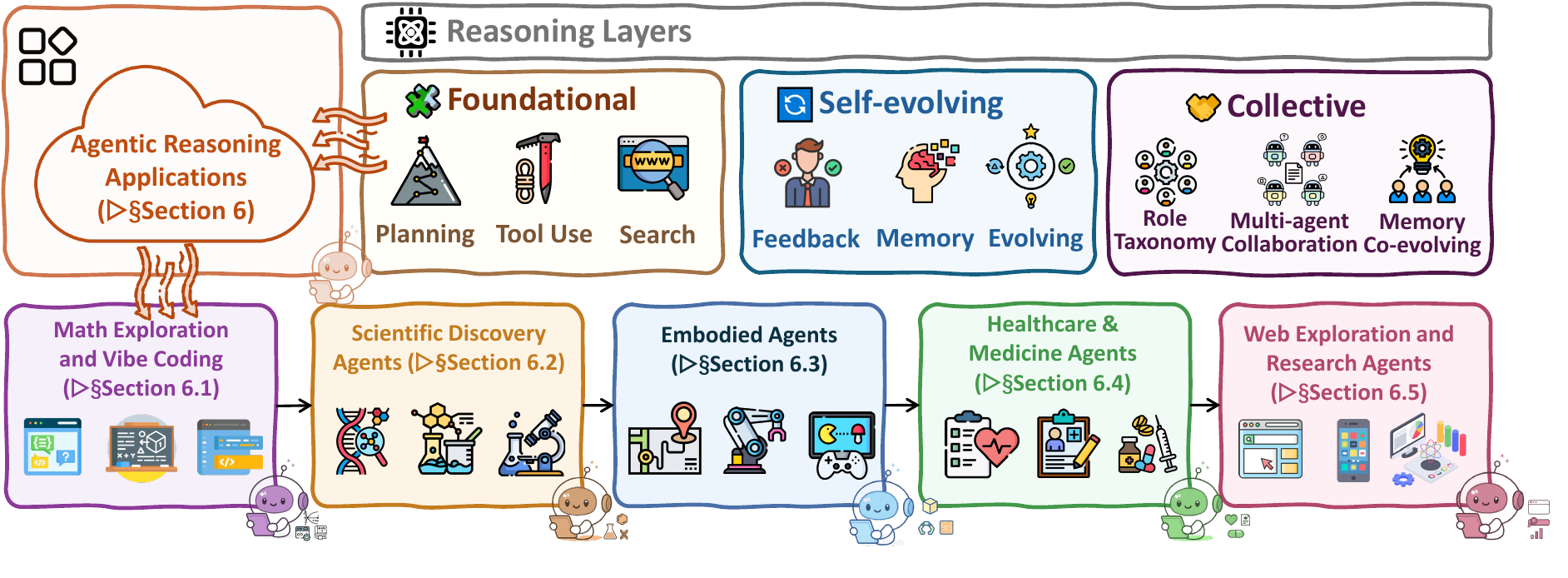}
    \caption{An overview of the applications of agentic reasoning.}
    \label{fig:application}
\end{figure}

\paragraph{Role Specialization and Joint Credit Assignment.}
Other approaches have explored structured role specialization and joint credit assignment. MALT \cite{MALT} trains sequential pipelines of heterogeneous agents using trajectory expansion and outcome-based reinforcement signals, allowing each agent to improve its specialized function while optimizing end-to-end collaborative performance. MARS \cite{mars} extends this idea to long-horizon research settings by jointly training complementary System~1 (fast, intuitive) and System~2 (deliberate, tool-using) agents via multi-agent reinforcement learning, enabling adaptive division of labor under complex tool interactions.

\paragraph{Preference- and Alignment-Driven Multi-Agent Evolution.}
Finally, another line of work has studied evolution under preference- and alignment-driven objectives. Preference-based multi-agent reinforcement learning \cite{marlhf} studies how collective policies and equilibria can be learned from preference-only feedback, addressing data coverage and stability challenges inherent in multi-agent settings. From a safety perspective, Alignment Waltz \cite{alignment-waltz} frames alignment as a cooperative co-evolution process between a generation agent and a feedback agent, where evolving guidance enables the system to iteratively refine unsafe or unhelpful behaviors. Collectively, these methods demonstrate how embedding reinforcement learning, co-evolution, and verifier-driven feedback into multi-agent training enables LLM-based systems to evolve from static collaborations into adaptive, self-improving organizations.
\addtocontents{toc}{\protect\setcounter{tocdepth}{2}}

\section{Applications}
\label{sec:applications}

Building on the established three-layer taxonomy (i.e. foundational, self-evolving and collective reasoning) mentioned in previous sections, we now examine how these capabilities manifest across real-world applications.
This section surveys representative reasoning-empowered agentic systems across several key domains, as illustrated in Figure~\ref{fig:application}, including math exploration and vibe coding (Section~\ref{sec:app-math}), scientific discovery (Section~\ref{sec:app-science}), robotics (Section~\ref{sec:app-embodied}),  healthcare (Section~\ref{sec:app-healthcare}), and autonomous web exploration and research (Section~\ref{sec:app-research}). Specifically, each domain exhibits distinctive forms of reasoning, influenced by its data modalities and environmental constraints.
Accordingly, our discussion in each subsection is organized around three layers:
(1) \textbf{core abilities} such as planning, tool use and search that span scientific hypothesis generation, embodied control, medical reasoning, automated experimentation and symbolic problem solving, for example;
(2) \textbf{self-evolving abilities} that integrate feedback, reflection and memory modules which refine domain-specific competence through iterative experiment loops, lifelong skill learning and clinical adaptation; and
(3) \textbf{collective multi-agent reasoning} that enables collaboration and specialization from cooperative scientific assistants to coordinated robotic teams, diagnostic ensembles or multi-aspect experts. This section highlights how agentic reasoning frameworks adapt to domain-specific knowledge structures and tasks, illustrating the transition from traditional LLM reasoning to goal-directed, domain-aware and active agentic intelligence.

\subsection{Math Exploration \& Vibe Coding Agents}
\label{sec:app-math}
Mathematics and code have traditionally served as two of the most widely used domains for evaluating reasoning in artificial intelligence, as both require structured symbolic manipulation and precise multi-step deduction. Traditional benchmark-driven evaluation in these domains is showing clear limitations. Widely used math datasets such as GSM8K \cite{cobbe2021training}, MATH \cite{hendrycks2021measuring}, and AIME \cite{MAA_AIME} are increasingly saturated, which makes it difficult to distinguish among modern high-performing models. The problems in these datasets often rely on a small set of recurring techniques and do not require the sustained and exploratory reasoning needed to assess more advanced mathematical capabilities. Even recent evaluations such as FrontierMath \cite{glazer2024frontiermath} continue to emphasize final-answer accuracy, which offers only a partial view of an agent’s reasoning process and its ability to adjust strategies during problem solving.

Under the agentic reasoning paradigm, however, both areas are undergoing a substantial shift from static problem solving to dynamic processes that emphasize exploration, adaptation, and collaboration. In mathematics, recent systems \citep{novikov2025alphaevolve,luong2025towards,trinh2024solving,romera2024mathematical} demonstrate that agents can engage in competition-level reasoning, building on the success of LLMs in coding tasks. Work in foundational mathematics \citep{swirszcz2025advancing,georgiev2025mathematical} further shows that agents can search for new problems, propose conjectures, construct auxiliary lemmas, and explore deeper structures in mathematical concepts. These developments position mathematics not merely as an evaluation benchmark but as a domain of active \emph{mathematical exploration}.

Large Language Models have also reshaped coding through the emerging workflow known as agentic coding and \emph{vibe coding} \citep{karpathy2025vibecoding,willison2025vibecoding}. In this paradigm, the model acts as an interactive collaborator that engages in multi-turn natural-language dialogue. Users iteratively design and refine programs while the agent maintains context, adapts to evolving requirements, and continuously self-corrects. Modern tools such as Copilot\footnote{\url{https://github.com/features/copilot}} and Cursor\footnote{\url{https://cursor.com}} have further popularized this collaborative workflow, making interactive programming a common practice in real-world software development.

In this section, we organize our discussion according to the three-layer framework introduced earlier. The foundational layer (Section~\ref{sec:app-math-planning}) concerns the core reasoning and execution skills: mathematical agents perform symbolic manipulations and step-by-step derivations across arithmetic, algebra, geometry, and calculus, while code agents carry out syntax-aware generation, implement functions, and verify correctness through interpreter or compiler feedback. The self-evolving layer (Section~\ref{sec:app-math-evolving}) introduces mechanisms for reflection and adaptation. Mathematical agents learn from intermediate reasoning traces to correct missteps or explore alternative solution paths, and code agents iteratively debug, refine, and optimize implementations based on runtime feedback or test results. The collective layer (Section~\ref{sec:app-math-collaboration}) focuses on collaboration, where agents exchange intermediate results, share reusable modules, and jointly develop complex proofs or codebases. Taken together, these layers reveal how mathematics and coding are becoming domains in which agentic reasoning enables increasingly creative and adaptive problem solving.

\subsubsection{Foundational agentic reasoning}
\label{sec:app-math-planning}

\paragraph{Planning.}

Explicit planning is widely recognized as a core mechanism for enhancing the structured 
reasoning capabilities of LLMs. In the domain of mathematical discovery, several systems exhibit structures that can be interpreted as forms of planning. In representation theory and knot theory, the system of \citet{davies2021advancing} guides human mathematicians by proposing intermediate objects and promising avenues of exploration, which function as high-level suggestions for organizing problem-solving workflows. In geometric reasoning, \citet{trinh2024solving} solves Olympiad-level geometry problems by decomposing them into sequential stages of construction, lemma generation, and verification, yielding a structured multi-step process that resembles a planned reasoning trajectory. Program-search approaches \citep{romera2024mathematical} iteratively refine candidate programs and mathematical structures, a procedure that naturally forms a coarse-to-fine exploration path. Large-scale exploration frameworks \citep{georgiev2025mathematical,swirszcz2025advancing} also operate through cycles of proposing, testing, and modifying conjectures or geometric objects, which collectively create a procedural structure aligned with planning. Efforts toward more robust mathematical reasoning \citep{luong2025towards} similarly rely on stepwise reasoning patterns, further reinforcing the presence of implicit planning dynamics. of implicit planning dynamics across mathematical agents.

In code agents, planning has likewise emerged as an essential component for organizing 
multi-step reasoning and enabling more structured decision-making. Early systems such as 
CodeChain \citep{le2023codechain} and CodeAct \citep{wang2024executable} introduce explicit planning 
or action spaces to support modular code construction, while KareCoder \citep{huang2024knowledge} 
integrate external knowledge sources or domain-specific information 
into the planning process. Subsequent works explore more structured planning organizations, including 
multi-stage control flows \citep{bairi2024codeplan,multi-stage}, tree-shaped planning structures 
\citep{li2024codetree,ni2024tree}, and adaptive refinement mechanisms \citep{aggarwal2025dars}. 
Planning has also been linked to improved exploration breadth: GIF-MCTS \citep{dainese2024generating} 
incorporates Monte Carlo Tree Search to explore multiple code-generation trajectories.
Recent extensions demonstrate applicability in specialized domains such as hardware design, 
where VerilogCoder \citep{ho2025verilogcoder} employs graph-structured planning and waveform-based 
verification. To address environments where state serialization is difficult, Guided Search 
\citep{zainullina2025guidedsearchstrategiesnonserializable} introduces lookahead and trajectory 
selection strategies for evaluating candidate actions without full environment access.

\paragraph{Tool-Use.}
Integrating external computational tools with LLMs has become a central mechanism for extending the reasoning and generation capabilities of single-agent systems. A defining characteristic of many mathematical reasoning systems is their integration with external computational tools. Formal theorem-proving agents such as \citet{thakur2024context} operate directly within the Lean proof assistant, selecting tactics and interacting with the underlying prover through in-context guidance. Position papers on formal mathematical reasoning \citep{yang2024formal} emphasize that progress in mathematical AI will depend on systems that can call theorem provers, satisfiability solvers, and computer algebra systems as part of a broader reasoning loop. Program-search frameworks for discovery \citep{romera2024mathematical} rely on executing generated programs and employing symbolic routines for verification. Generative modelling approaches \citep{ellenberg2025generative} make use of computational number-theoretic tools to check and filter generated candidates. Geometry-focused systems \citep{trinh2024solving,hubert2024ai} integrate automated geometric solvers and checkers to validate constructions and derived relations. Across these systems, external computational resources play a central role in enabling correct and scalable mathematical reasoning.

In code agents, external tools have similarly become crucial for extending the capabilities of 
LLM-based agents beyond pure text generation. Early work such as Toolformer \citep{schick2023toolformer} 
and ToolCoder \citep{zhang2023toolcoder} explored how models can learn to invoke APIs or search tools 
to obtain missing information during generation. Subsequent systems integrate increasingly rich 
toolchains: ToolGen \citep{wang2024toolgen} leverages automatic completion tools to resolve undefined 
dependencies, while CodeAgent \citep{zhang2024codeagent} incorporates multiple programming utilities 
 including search, documentation reading, symbol navigation, and code execution to support more 
realistic software workflows. Several methods focus on improving tool-feedback loops, such as ROCODE 
\citep{jiang2024rocode}, which combines real-time error detection with adaptive backtracking, and 
CodeTool \citep{lu2025codetool}, which introduces process-level supervision to improve the reliability 
of tool invocation. Collectively, these systems show that tool integration provides essential external 
signals, via search results, documentation, static analysis, or execution feedback, that extend the 
reasoning and generation capabilities of single-agent LLMs.

\paragraph{Search and Retrieval.}

Search and retrieval has emerged as a complementary mechanism that enriches model contexts through external information sources.
Search is a recurring mechanism in mathematical discovery. Program-search based systems \citep{romera2024mathematical} treat mathematical discovery as navigating a program space in which candidate programs encode conjectures or structural hypotheses, with iterative filtering based on symbolic or numerical checks. Generative modelling approaches \citep{ellenberg2025generative} explore families of mathematical objects by sampling from flexible distributions that capture structural regularities. Geometric systems such as \citet{trinh2024solving} and \citet{swirszcz2025advancing} search over constructions, configurations, and high-dimensional polytopes, guided by learned heuristics or structural constraints. Large-scale discovery frameworks \citep{georgiev2025mathematical} operate through repeated propose--test--refine cycles across conjectures, supporting wide exploration over mathematical landscapes. All these systems rely on systematic search procedures that structure the exploration of mathematical ideas.

In code generation, repository-level retrieval systems 
such as RepoHyper \citep{phan2024repohyper} locate reusable code segments from large-scale code bases 
to provide more informative contexts for generation. CodeNav \citep{gupta2024codenav} dynamically 
indexes real repositories during generation, retrieving relevant functions and adjusting based on 
execution feedback. AUTOPATCH \citep{acharya2025optimizing} applies retrieval to performance 
optimization, combining historical code examples with control flow graph analysis for 
context-aware improvements. Structure-aware retrieval has also been explored: knowledge-graph-based 
repository representations \citep{athale2025knowledge} improve retrieval quality by capturing 
symbolic and relational structure, while cAST \citep{zhang2025cast} introduces AST-based chunking to 
enhance syntactic coherence and retrieval granularity. These retrieval methods demonstrate how 
external knowledge sources can augment single-agent LLMs by providing high-quality, structured 
contexts that guide both understanding and generation.

\subsubsection{Self-evolving agentic reasoning}
\label{sec:app-math-evolving}

\paragraph{Agentic Feedback and Reflection.}
Across mathematical and code reasoning tasks, feedback operates as an external signal that 
highlights discrepancies, confirms correct inferences, and directs the agent toward more reliable 
subsequent computations. Feedback mechanisms appear prominently across mathematical discovery systems. In program-search 
based discovery \citep{romera2024mathematical}, executing candidate programs and evaluating their 
outputs against constraints yields counterexamples or confirmations, enabling iterative refinement of 
conjectures. In geometry, automated checkers validate constructions and derived relationships 
\citep{trinh2024solving,hubert2024ai}, providing correctness signals that guide subsequent revisions. 
Interactive evaluation frameworks \citep{collins2024evaluating} show that human clarifications and 
follow-up prompts expose reasoning errors and improve model responses. Position work on formal 
reasoning \citep{yang2024formal} highlights verification, proof checking, and model checking as 
essential sources of structured feedback. In several systems involving multiple candidate hypotheses 
\citep{romera2024mathematical,ellenberg2025generative,georgiev2025mathematical}, 
the use of verification signals to retain promising candidates functions analogously 
to a fitness-based evaluation step, since these signals determine which hypotheses 
survive and which are discarded, thereby shaping the direction of subsequent exploration 
without introducing an explicit learning signal.

For code agents, feedback and reflection are central to improving reliability over multi-step 
reasoning. Fault-aware editing methods such as Self-Edit \citep{zhang2023self} incorporate 
execution-based signals to refine erroneous code, while Self-Repair 
\citep{olausson2024selfrepairsilverbulletcode} integrates code and feedback models to diagnose test 
failures and propose targeted corrections. More structured systems like LeDeX \citep{jiang2024ledex} 
combine stepwise annotation, execution-driven verification, and automated repair into a closed-loop 
pipeline in which feedback continually informs the next revision. Reflection also functions as a 
form of memory: iterative self-improvement frameworks such as Self-Refine \citep{madaan2023self}, 
Self-Iteration \citep{chang2023self}, and Self-Debug \citep{chen2023teaching} reuse earlier drafts, 
analyses, and explanations to guide subsequent revisions, while artifact-level mechanisms such as 
CodeChain \citep{le2023codechain} and LeDeX \citep{jiang2024ledex} retain reusable components, 
corrected snippets, and execution traces as persistent representations. Together, these approaches 
demonstrate how feedback—whether symbolic, execution-based, or self-generated—interacts with 
iterative memory to support structured refinement and long-horizon improvement in code-oriented 
agentic systems.

\paragraph{Memory.}
Memory provides agents with a mechanism for retaining and leveraging information from earlier 
reasoning steps, allowing them to maintain consistency, improve intermediate states, and improve 
their performance over extended problem-solving horizons. While few systems introduce an explicit memory module, many mathematical agents rely on forms of persistent state that can be viewed as implicit memory. Interactive evaluation frameworks \citep{collins2024evaluating} maintain conversational and problem-state context across multiple turns, allowing models to build upon earlier partial derivations. Formal-theorem-proving agents \citep{thakur2024context} operate over evolving proof states in Lean, which accumulate tactics, subgoals, and intermediate lemmas, functioning as structured persistent information. Program-search and discovery systems \citep{romera2024mathematical,georgiev2025mathematical} retain conjecture histories, counterexamples, and successful constructions as part of their iterative refinement processes. Their role in preserving and reusing information across reasoning steps aligns with the broader notion of memory in agentic systems.

In code agents, memory increasingly takes the form of explicit structures that maintain coherence
over long-horizon generation. Several systems construct shared or structured workspaces: 
Self-Collaboration \citep{dong2024self} introduces a blackboard memory for storing task descriptions,
intermediate drafts, and revision records, enabling agents to coordinate through a common 
representation. Architectural approaches such as L2MAC \citep{holt2023l2mac} and Cogito 
\citep{li2025cogito} extend this idea by organizing context into dedicated registers, hierarchical 
memory units, or long-term knowledge stores, overcoming context-window limits and supporting 
multi-file or large-function reasoning. Across these designs, the underlying insight is consistent: effective code agents require 
persistent, structured, and often domain-aware memory that preserves intermediate reasoning and 
enables self-improvement across extended development trajectories.

\subsubsection{Collective multi-agent reasoning}
\label{sec:app-math-collaboration}
To address the growing complexity of tasks in mathematical discovery and code generation, recent 
systems increasingly rely on multi-agent or modular designs that decompose problems into 
cooperating specialized components. Mathematical discovery frameworks often organize reasoning into explicitly defined 
multi-agent or multi-component workflows that collaborate to explore and validate mathematical ideas. The polytope-generation system \citep{swirszcz2025advancing} uses multiple specialized components that generate, evaluate, and refine geometric objects, forming a genuine collaborative workflow. Large-scale exploration frameworks \citep{georgiev2025mathematical} often divide discovery into modules for proposing conjectures, identifying counterexamples, and refining statements, which, although implemented within a unified system, mirror multi-agent role specialization. Early work on AI-assisted mathematical research \citep{davies2021advancing} and Olympiad-level systems \citep{hubert2024ai} also involve human--AI collaboration, where human mathematicians interact with AI systems in a complementary manner. These developments indicate that mathematical discovery is an inherently collaborative process, and multi-agent architectures provide a natural vehicle for expressing such collaboration in agentic systems.

Multi-agent systems for code generation have progressed from simple role-based pipelines to adaptive, 
collaborative frameworks capable of handling long-horizon software development. Early approaches such 
as Self-Collaboration \citep{dong2024self} and AgentCoder \citep{huang2023agentcoder} decompose tasks 
into sequential roles, while hierarchical designs like PairCoder \citep{zhang2024pair} and FlowGen 
\citep{lin2024soen} introduce an architecture in which high-level agents handle planning and lower-level 
agents carry out concrete implementation. Flexible systems such as SoA \citep{ishibashi2024self} further 
adjust the number and specialization of agents in response to task complexity. Other frameworks, 
including MapCoder \citep{islam2024mapcoder}, AutoSafeCoder 
\citep{nunez2024autosafecoder}, and QualityFlow 
\citep{hu2025qualityflow}, rely on repeated cycles in which multiple agents 
generate, test, analyze, and repair code. Recent work explores self-evolving system structures, as in 
SEW \citep{liu2025sew}, which reorganizes collaboration pathways based on runtime feedback, and EvoMAC 
\citep{hu2024self}, which adjusts agent strategies through an iterative text-based update mechanism. 
Collaborative optimization methods such as Lingma SWE-GPT \citep{ma2024lingmaswegptopendevelopmentprocesscentric}, 
CodeCoR \citep{pan2025codecor}, SyncMind \citep{guo2025syncmind}, and CANDOR 
\citep{xu2025hallucinationconsensusmultiagentllms} explicitly improve cross-agent coordination. 
Together, these systems show a clear shift toward multi-agent code generators that rely not only on 
role decomposition, but also on reflection, distributed evaluation, adaptive restructuring, and 
team-level optimization, transforming code generation into an increasingly coordinated and resilient 
problem-solving process.

\subsection{Scientific Discovery Agents}
\label{sec:app-science}
Scientific-discovery agents aim to accelerate the entire life cycle for scientific research, from hypothesis generation through experimental execution, by coupling LLMs with domain-specific simulators, laboratory automation and up-to-date literature. These systems ground decision in verifiable processes while handling heterogeneous data, safety constraints and long-horizon goals.  

In this subsection, we begin with the foundational layer (Section~\ref{sec:app-science-planning}), which encompasses planning under scientific context, tool-augmented interaction with scientific resources, search and retrieval mechanisms including RAG-based systems and execution-time integration with laboratory hardware.
Building upon these capabilities, the self-evolving layer (Section~\ref{sec:app-science-memory}) introduces agentic memory, feedback and reflection, which enable scientific agents to refine hypotheses, adapt protocols and learn from experimental outcomes.
Finally, the collective layer (Section~\ref{sec:app-science-multi}) explores multi-agent collaboration, where agents coordinate roles, share intermediate knowledge and jointly reason toward complex scientific goals.

\subsubsection{Foundational agentic reasoning}
\label{sec:app-science-foundational}

\paragraph{Planning.}
\label{sec:app-science-planning}

Scientific agents utilize reasoning-enhanced planning ability to decompose a research goal into steps, decides which tool or simulator to call next, then revises the plan as evidence arrives. In short, the \textit{chain of thought} emerges from LLM reasoning that compiles instructions into rigorous executable plans~\cite{wei2022chain}.
% There are certain ways that LLM reasoning interacts with agentic planning ability. 
For example, ProtAgents~\cite{ghafarollahi2024protagents} materializes a planner agent that utilizes LLM reasoning capability to formulate a concrete plan for protein analysis and keep modifying it with feedback from another critic agent, and Eunomia~\cite{ansari2023agentbasedlearningmaterialsdatasets} uses ReAct-style~\cite{yao2023react} workflow to make in-context reasoning: after retrieving a top-$k$ evidence set, the backbone LLM quote a warranting sentence, and that citation drives the next action choice. Other examples include MatExpert~\cite{ding2024matexpertdecomposingmaterialsdiscovery}, which deploys a chain-of-thought LLM to author a stepwise transition pathway and then emits a structured crystal candidate from a feedback loop.

Planning can also act as reasoning constraint. For instance, Curie~\cite{kon2025curierigorousautomatedscientific} utilizes a rigor engine to align, setup and do reproducibility check within planning steps proposed by the Architect LLM. Thus, the Architect's free-form reasoning cannot advance unless these rigor gates are satisfied, which transforms planning into both a guide and a regulator of the reasoning process. In addition, a general purpose biomedical agent, Biomni~\cite{huang2025biomni} constrains its reasoning within a dynamically constructed biomedical action space of comprehensive tools,  software packages and databases, requiring each hypothesis to be operationalized as executable code.

% ------------------------------------------------------------------ %
\paragraph{Tool-Use.}
\label{sec:app-science-tool}

Tool use is an important part of the reasoning loop for scientific agents nowadays. Specifically, rather than following rigid rules,   these agents can decide which tool and when to call, how to fill parameters and verify or revise based on evidence. For example, SciAgent~\cite{ma2024sciagenttoolaugmentedlanguagemodels} formalizes \textit{tool-augmented reasoning} as a four-step procedure: planning, retrieval, tool-based action and execution. Agents are trained to decide when to call a tool, which one, and how to integrate it into solving scientific tasks. Through domain-specific tools, ChemCrow~\cite{bran2023chemcrowaugmentinglargelanguagemodels} chains various expert chemistry tools so intermediate calculations become premises in the next reasoning step, which enables end-to-end planning and autonomous syntheses. CACTUS~\cite{mcnaughton2024cactus} similarly grounds explanations in cheminformatics outputs, reducing reliance on free-form reasoning by language models alone. 

Other notable examples include ChemToolAgent~\cite{yu2024chemtoolagent} and CheMatAgent~\cite{wu2025chematagentenhancingllmschemistry}. In particular, ChemToolAgent~\cite{yu2024chemtoolagent} employs a ReAct-like~\cite{yao2023react} architecture with multiple specialized chemistry tools, allowing the LLM to choose and parameterize tool calls while CheMatAgent~\cite{wu2025chematagentenhancingllmschemistry} pushes further by learning tool use: it integrates over 100 chemistry/materials tools, curates a tool-specific benchmark, and uses Monte Carlo Tree Search with step-level fine-tuning to learn both which tool to pick and how to fill arguments.

For biomedical agents, TxAgent~\cite{gao2025txagentaiagenttherapeutic} scales therapeutic reasoning across 211 vetted tools and it carries out multi-step reasoning that reconciles drug labels, interactions, and patient context—turning clinical justification into an executable trace. On the other hand, AgentMD~\cite{jin2024agentmdempoweringlanguageagents} builds a two-stage tool memory: it first mines thousands of clinical calculators from literature (i.e. making tools), then selects and applies the right ones at inference (i.e. using tools), pinning predictions to concrete computations. Other recent systems~\cite{lála2023paperqaretrievalaugmentedgenerativeagent, skarlinski2024languageagentsachievesuperhuman,chiang2024llamp,zhang2024honeycombflexiblellmbasedagent,qu2025crisprgptagenticautomationgeneediting,gao2025pharmagentsbuildingvirtualpharma} reinforce similar design: co-design tool-use with reasoning so each claim is computable and auditable.

Another notable category of tool-use is the ability of agentic execution, which includes but not limited to run codes and simulate environments. Execution layers bridge high-level plans to physical infrastructure, which enables scientific agents to autonomously operate laboratory hardware, orchestrate simulation pipelines, and manage large-scale data workflows. Recent works such as Organa~\cite{darvish2025organaroboticassistantautomated} ties LLM reasoning to task-and-motion planning plus scheduling and perception, executing multi-step experiments with autonomous robots;  AtomAgents~\cite{ghafarollahi2024atomagentsalloydesigndiscovery} exemplifies the simulation side of execution: a physics-aware system that plans and runs atomistic workflows, coordinating tools for code execution, analysis, and hypothesis checking; and Chemist-x~\cite{chen2025chemistxlargelanguagemodelempowered} shows wet-lab execution beyond a digital-only scenario, where agents generate control scripts and drive an automated platform to validate conditions without human intervention. 

Several other platforms couple execution with optimization or team-based autonomy. For instance, SGA~\cite{ma2024llmsimulationbileveloptimizers} formalizes the workflow of \textit{LLM-as-proposer and simulator-as-optimizer}  while MatExpert~\cite{ding2024matexpertdecomposingmaterialsdiscovery} operates a \textit{retrieval, transition and generation} workflow for material discovery tasks, and CellAgent~\cite{xiao2024cellagentllmdrivenmultiagentframework} coordinates planner, executor and evaluator roles to run full single-cell analysis pipelines.

\paragraph{Search and retrieval.}

\label{sec:app-science-search}

Beyond simple context stuffing, recent scientific agentic systems elevate retrieval into a deliberate reasoning step: agents decide when and what to fetch, and how to use the evidence before committing to a hypothesis. With retrieval ability, BioDiscoveryAgent~\cite{roohani2025biodiscoveryagentaiagentdesigning} pulls literature and interim assay results inside a closed loop so the model’s next gene-perturbation choices are conditioned on what was read and measured; while DrugAgent~\cite{inoue2025drugagentexplainabledrugrepurposing} coordinates knowledge graph queries, targeted literature search through web API and machine learning predictors. Its planner selects retrieval actions and then reconciles heterogeneous evidence into an explainable rationale. To facilitate scientific research, ARIA~\cite{ramirezmedina2025acceleratingscientificresearchmultillm} operationalizes a \textit{search, filter then synthesis} workflow as role-bound steps that carry citations forward, turning literature into actionable procedures. Similarly, AI Scientist-v2~\cite{yamada2025ai} employs an agentic tree-search framework in which the agent actively queries scientific literature database during hypothesis formulation and manuscript drafting, ensuring that analyses and writing are grounded in existing evidence. For research idea generation, another recent work~\cite{qi2023largelanguagemodelszero} constrains the process with curated background packets, using retrieval as an experimental control.

Building on these developments, retrieval-augmented generation (RAG) frameworks position external sources not merely as supporting references but as active components of the reasoning process. Specifically, RAG-enhanced scientific agents make external sources as primary inputs to LLM context and reasoning material, mostly with explicit planning, passage extraction, citation and contradiction checks. For example, PaperQA~\cite{lála2023paperqaretrievalaugmentedgenerativeagent} and PaperQA2~\cite{skarlinski2024languageagentsachievesuperhuman} treat retrieval as the main loop. By deciding which documents to read, attributing every claim, and detecting conflicts to steer synthesis, these works can yield expert-level literature reviews that are inherently verifiable. In material science, LLaMP~\cite{chiang2024llamp} extends RAG beyond text. Specifically, it utilizes  hierarchical ReAct~\cite{yao2023react} agents call material-specific APIs to fetch band gaps or elastic tensors, edit structures and then reason with computed properties.

\subsubsection{Self-evolving agentic reasoning}
\label{sec:app-science-evolving}

Scientific discovery agents can go beyond static reasoning and acquire the ability to self-evolve, which is to learn from experience, refine their internal representations and improve decision quality over successive interactions. This self-evolving layer equips agents with mechanisms to monitor and revise their own reasoning, retain and reuse intermediate hypotheses and adjust future plans based on external feedback or environmental signals. In the following paragraphs, we discuss how memory modules enable the accumulation  of scientific knowledge and how feedback and reflection mechanisms support continual adaptation and reasoning consistency throughout long-horizon scientific workflows.

\paragraph{Memory.}
\label{sec:app-science-memory}

ChemAgent~\cite{tang2025chemagentselfupdatinglibrarylarge} implements a self-updating library. It decomposes chemistry problems into sub-tasks and writes reusable \textit{skills} (ex: procedures, patterns, solutions) that later prompts can retrieve and adapt, stabilizing long multi-step reasoning without re-deriving everything from scratch. On the other hand, MatAgent~\cite{takahara2025acceleratedinorganicmaterialsdesign} emphasizes interpretable generation for inorganic materials, where \textit{short-term memory} recalls recent compositions and feedback, \textit{long-term memory} preserves successful designs together with their reasoning traces, and both are reused across iterations to guide proposal refinement and enable transparent audit.

\paragraph{Agentic Feedback and Reflection.}
\label{sec:app-science-feedback-reflect}

Firstly, Scientific Generative Agent~\cite{ma2024llmsimulationbileveloptimizers} ties discrete LLM proposals to inner-loop simulations that optimize continuous parameters, advancing only when evidence improves. The reflection ability is driven by measurable loss reductions. Next, ChemReasoner~\cite{sprueill2024chemreasoner} performs heuristic search over the LLM’s idea space but scores and steers candidates with quantum-chemical feedback, turning electronic-structure signals into a principled critique of linguistic hypotheses. Complementing these physics-based signals, Curie~\cite{kon2025curierigorousautomatedscientific} embeds rigor check directly into control flow via intra-agent checks, inter-agent gates and an experiment-knowledge module. In parallel, LLMatDesign~\cite{jia2024llmatdesignautonomousmaterialsdiscovery} builds explicit self-reflection into materials workflows, prompting the agent to surface and repair inconsistencies before they propagate to tool calls. Moreover, NovelSeek~\cite{novelseekteam2025novelseekagentscientist} utilizes reflection as a closed loop, updating code and plans with human-interactive feedback after each round. Finally, a recent study~\cite{kumbhar2025hypothesisgenerationmaterialsdiscovery} regularizes the process up front with explicit goals \& constraints and afterwards with standardized scoring to provides an objective standard that makes reflection repeatable.

\subsubsection{Collective multi-agent reasoning}
\label{sec:app-science-multi}

Multi-agent frameworks for scientific discovery distribute labor across specialized LLM-driven roles, where advanced LLM reasoning not only orchestrates coordination between scientific agents but also adjudicates conflicting evidence to maintain coherence in the process. 

To illustrate, we introduce some important multi-agent frameworks as follows. Firstly, ProtAgents~\cite{ghafarollahi2024protagents} exemplifies this pattern in protein design. The framework involve agents for literature retrieval, structure analysis, physics simulation, and results analysis. Specifically, the backbone LLM directs reasoning over multi-modal outputs, choosing when to iterate or convergence-check based on feedback signals. PiFlow~\cite{pu2025piflow}, on the other hand, instantiates reasoning as principle-aware uncertainty reduction with a multi-agent loop in which a Planner agent relays strategy to a Hypothesis agent and a validation loop, explicitly tying multi-agent communication to hypothesis–evidence alignment.  AtomAgents~\cite{ghafarollahi2024atomagentsalloydesigndiscovery} also brings similar role specialization to alloy discovery. In particular, the agent uses LLM-guided reasoning to control over when to trigger simulations and how to evaluate multi-modal results, letting reasoning allocate computational resources and prune alloy candidates.

With a similar \textit{planner, executor and evaluator} framework, CellAgent~\cite{xiao2024cellagentllmdrivenmultiagentframework} instantiates researches on single-cell analysis, where the planner LLM reasoning selects tools or hyper-parameters and the evaluator LLM triggers self-iterative re-runs when quality checks fail. Some other notable works include ARIA~\cite{ramirezmedina2025acceleratingscientificresearchmultillm} that introduces a four-agent framework (scout, filter, synthesizer and procedure-drafter), Curie~\cite{kon2025curierigorousautomatedscientific} that embeds rigor into multi-agent planning, Team of AI-made Scientists (TAIS)~\cite{liu2024toward} for gene-expression discovery and the Virtual Lab~\cite{swanson2024virtual} for nano-body design with role agents. 

\subsection{Embodied Agents}
\label{sec:app-embodied}

Embodied agents extend reasoning beyond text, anchoring language in robotic perception, manipulation and navigation.  By embedding LLMs within robotic and simulated bodies, these embodied agents tackle real-world generalization, continual adaptation and multi-modal grounding.  

In this subsection, we begin with the foundational layer (Section~\ref{sec:app-embodied-planning}), which covers long-horizon embodied planning, tool-assisted perception, manipulation and execution. Building upon these capabilities, the self-evolving layer (Section~\ref{sec:app-embodied-evolve}) introduces agentic memory, feedback and self-reflection capabilities  enabling robots to refine control policies, adapt to novel environments and improve performance through continual interaction. Finally, the collective reasoning layer explores multi-robot collaboration (Section~\ref{sec:app-embodied-multi}), where agents coordinate perception, share learned representations and jointly reason about tasks to achieve complex embodied goals.

\subsubsection{Foundational agentic reasoning}

\label{sec:app-embodied-planning}

\paragraph{Planning.}

Early work such as SayCan~\cite{ahn2022icanisay} established the template by mapping linguistic descriptions to skill affordance estimates and  SayPlan~\cite{DBLP:conf/corl/RanaHGA0S23} refined this grounding by leveraging 3D scene graphs to align goal references with object-centric representations and spatial models. Beyond symbolic representations, EmbodiedGPT~\cite{mu2023embodiedgptvisionlanguagepretrainingembodied} use curated video CoT annotations of sub-goals to train models taht map multi-model input to structured sequences for embodied planning, while context-aware planning system~\cite{kim2024contextawareplanningenvironmentawarememory} adds semantic spatial map and object location information to the planning pipeline, enabling dynamical planning during execution.  In addition, DEPS~\cite{DBLP:journals/corr/abs-2302-01560} introduces an interactive planning loop (i.e. describe, explain, plan and select) for open-world multi-task agents.

Embodied agents also rely on multi-modal reasoning traces that explicitly align perception with action. For example, Embodied CoT~\cite{zawalski2024robotic} trains vision-language-action models to generate reasoning steps incorporating visual features before executing an action. Fast\,ECoT~\cite{duan2025fast} accelerates this by caching and re-using reasoning segments across time-steps, reducing inference latency while preserving task success. More recently, Cosmos-Reason1~\cite{nvidia2025cosmosreason1physicalcommonsense} establishes an ontology of space, time and dynamics that lets CoT sequences encode structured physical priors. CoT-VLA~\cite{zhao2025cot} builds a visual chain-of-thought by predicting future image frames as intermediate sub-goals prior to action generation. Finally, Emma-X~\cite{sun2024emmaxembodiedmultimodalaction} integrates grounded chain-of-thought with look-ahead spatial reasoning, improving long-horizon embodied task performance.

Another line of works strengthen embodied planning through reinforcement learning, considering planning not only as static decomposition but as a self-evolving process that adapts to environment feedback. Robot-R1~\cite{kim2025robotr1reinforcementlearningenhanced} trains large VLMs to predict keypoint transitions under visual context, turning RL into a mechanism for learning physically grounded forward models. ManipLVM-R1~\cite{song2025maniplvm} exploits verifiable physical reward signals (e.g., trajectory match and affordance correctness) to reduce reliance on dense expert annotation. Embodied-R~\cite{zhao2025embodied} presents a collaborative framework where VLMs handle perception and smaller LMs handle reasoning, and the whole is trained via RL for embodied spatial reasoning. VIKI-R~\cite{kang2025viki} further extends this direction into heterogeneous multi-agent cooperation with a two-stage design, employing a two‐stage pipeline of chain-of-thought fine-tuning followed by hierarchical RL across agents coordinating activation and planning.

\paragraph{Tool-use.}

\label{sec:app-embodied-tool}

Embodied agents can also be strengthened to interact with external tools to enhance perception and compensate for incomplete observations. GSCE~\cite{wang2025gscepromptframeworkenhanced}, for example, provides a prompt-framework that binds skill APIs and constraints for safe LLM-driven drone control. MineDojo~\cite{fan2022minedojo} links agents to internet-scale corpora and thus enabling richer affordance grounding. Physical AI Agents~\cite{bousetouane2025physicalaiagentsintegrating} further introduces a modular architecture and a retrieval augmented generation design pattern for embedding real-world physical interaction into LLM-driven agents. Beyond offline tool use, some systems treat the environment itself as an API. For example, Matcha agent~\cite{zhao2023chatenvironmentinteractivemultimodal} uses an LLM to issue queries about objects and scenes and thereby acquire perceptual information needed for task completion.

On the other hand, execution module is one of the most important tool type. It translates high-level language instructions into continuous motor commands, enabling embodied agents to act reliably in physical environments.  Early systems such as SayCan~\cite{ahn2022icanisay} uses language to invoke robot pick-and-place skills; while LEO~\cite{huang2024embodiedgeneralistagent3d} broaden execution to more general manipulation settings and Hi Robot~\cite{shi2025hi} uses a VLM reasoner to process complex prompts and a low-level action policy executes the chosen step. More recent efforts broaden the execution space: Gemini Robotics~\cite{geminiroboticsteam2025geminiroboticsbringingai} introduces a large-scale vision-language-action model for real-world robot control and Octopus~\cite{yang2024octopusembodiedvisionlanguageprogrammer} generates executable code in simulated environments that bridges planning and manipulation.

Beyond single-agent control, hybrid pipelines couple reactive reflexes with language-guided policies to support complex domains. For example, CaPo~\cite{liu2024capo} incorporates an execution phase where agents carry out decomposed sub-tasks and adapt their meta-plan based on progress; COHERENT~\cite{liu2024coherent} embeds a robot executor module within its PEFA (i.e. proposal, execution, feedback and adjustment) loop, which ensures each assigned sub-task is acted and refined appropriately; and MP5~\cite{qin2024mp5multimodalopenendedembodied} integrates multi-modal perception to generate executable plans in open-ended Minecraft. At the perception–action interface, LLM-Planner~\cite{song2023llmplannerfewshotgroundedplanning} generates sub-goals and maps them into action sequences via a low-level controller and EmbodiedGPT~\cite{mu2023embodiedgptvisionlanguagepretrainingembodied} illustrate how LLM-generated plans can be translated into control policy for embodied control in physical environments.

\paragraph{Search and retrieval.}

Embodied agents can also use search and retrieval ability to ground language in spatial structure and past experience. Early navigation systems such as L3MVN~\cite{yu2023l3mvn} use LLMs to query a semantic map and select promising frontiers as long-term goals during visual target navigation, while SayNav~\cite{rajvanshi2024saynav} and SayPlan~\cite{DBLP:conf/corl/RanaHGA0S23} build 3D scene graphs and then search task-relevant subgraphs so language instructions can be translated into grounded waypoints and sub-tasks in large environments. Long-horizon navigation works like ReMEmbR~\cite{anwar2025remembr} maintain a structured spatio-temporal memory that can be queried to answer “where” and “when” questions about past robot experience. Additionally, RAG-style systems make retrieval a first-class part of the planning loop: Embodied-RAG~\cite{quanting2024embodiedrag} and EmbodiedRAG~\cite{meghan2024embodiedrag} treat an agent’s experience and 3D scene graphs as non-parametric memories from which task-relevant episodes or subgraphs are retrieved for navigation and task planning; Retrieval-Augmented Embodied Agents~\cite{zhu2024retrieval} retrieve policies from a shared memory bank and condition action generation on them; and MLLM-as-Retriever~\cite{junpeng2024mllm} trains a multi-modal LLM retriever to rank past trajectories so each decision step can condition on the most useful prior experience rather than only the current observation.

\subsubsection{Self-evolving agentic reasoning}
\label{sec:app-embodied-evolve}

Embodied agents reliably achieve long-horizon autonomy when they can self-evolve over time: monitor their own internal states, store and update task-relevant knowledge and adjust behaviors when plans deviate. In the following paragraphs, we examine how memory modules, feedback signals and agentic reflection enable embodied agents to turn planning from a one-shot process into a continually improving cycle of behavior.

\paragraph{Memory.}
Effective memory mechanisms enable agents to reuse past experiences and maintain coherent task execution over extended interactions. Many systems cache recent observations in episodic buffers while summarizing long-term semantics in structured graphs, as in household planning~\cite{glocker2025llm} and long-horizon agents with hybrid multi-modal memory~\cite{li2024optimus1}. Skills and routines can be shared across tasks via indexed memory stores. For example, HELPER-X~\cite{sarch2023openendedinstructableembodiedagents} indexes discovered skills and action scripts, which aid future dialogue and can be shared across domains.
Spatial navigation methods such as BrainNav~\cite{ling2025endowingembodiedagentsspatial} maintain biologically inspired dual-map memories linked by a hippocampal hub to reduce hallucinations and drift. Broader contexts also benefit: CAPEAM~\cite{kim2024contextawareplanningenvironmentawarememory} incorporates environment-aware memory modules that track object states and spatial changes. Finally, lifelong episodic systems such as Ella~\cite{zhang2025ellaembodiedsocialagents} maintains long-term multi-modal memory system to support social-robot interaction.

\paragraph{Agentic Feedback and Reflection.}
Dialogue-based critique, calibrated uncertainty and environment-aware reward shaping refine policies beyond binary success signals. For example, Matcha agent~\cite{zhao2023chatenvironmentinteractivemultimodal} treats objects and scenes as interactive information sources before acting and FAMER~\cite{wang2025communicationefficientdesirealignmentembodied} uses lightweight preference feedback to adapt embodied agents to user intentions in real time. Uncertainty-aware planners such as KnowNo~\cite{ren2023robotsaskhelpuncertainty}, which proactively solicit guidance when confidence falls below guarantees, and Octopus~\cite{yang2024octopusembodiedvisionlanguageprogrammer}, which exploits environmental feedback to improve generated executable programs over time. At the multi-agent level, MindForge~\cite{licua2024mindforge} introduces theory-of-mind style perspective feedback so heterogeneous robots adapt to each other’s reasoning strategies; while ReAd~\cite{zhang2024efficientllmgroundingembodied} introduces a advantage-based feedback loop that enables an LLM planner to self-refine its collaboration strategies across embodied multi-agent tasks.

Robust reflection mechanisms help agents anticipate failures by monitoring their own reasoning and actions and then adjusting plans. Optimus-1~\cite{li2024optimus1} couples a \emph{Knowledge-guided Planner} with an \emph{Experience-Driven Reflector} to revise decisions using stored experience, while another recent study~\cite{Moncada_Ramirez_2025} defines structured agentic workflows (including self-Reflection, multi-Agent reflection and LLM Ensemble) that enable robots to reflect on and refine LLM-generated object-centered plans, thus reducing reasoning errors. Systems such as EMAC+~\cite{ao2025emac} interleave perception, planning and verification steps to perform online plan refinement and earlier works such as Voyager~\cite{wang2023voyager} also embeds an iterative prompting loop that uses environment feedback and execution errors to refine its skill library over time.

\subsubsection{Collective multi-agent reasoning}
\label{sec:app-embodied-multi}

Multi-agent collaboration enables embodied systems to divide labor and coordinate complex tasks more efficiently, with language often serving as the primary medium for negotiation and role allocation. For instance, SMART-LLM~\cite{kannan2024smartllmsmartmultiagentrobot} decomposes high-level instructions and allocates sub-tasks across multiple robots, while CaPo~\cite{liu2024capo} optimizes cooperative plans to avoid redundant exploration. For heterogeneity and coordination mechanisms, COHERENT~\cite{liu2024coherent} deploys a propose-execution-feedback-adjust loop across diverse robot types to enable seamless joint operation. In addition, Theory of Mind (ToM), which refers to an embodied agent’s ability to infer and reason about others' beliefs and mental states, is also highly related to embodied multi-agent systems~\cite{li2023theory, wu2025large, cross2024hypothetical}. For example, MindForge~\cite{licua2024mindforge} equips agents with explicit theory-of-mind representations and natural inter-agent communication to coordinate collaboratively.

For multi-modal frameworks, EMAC+~\cite{ao2025emac} integrate vision and language modules and continuously refine plans via visual feedback, COMBO~\cite{zhang2025combocompositionalworldmodels} integrates vision and language modules and continuously refine plans via visual feedback, and VIKI-R~\cite{kang2025viki} demonstrates reinforcement learning as a scalable coordination mechanism among embodied agents. At larger scales, studies such as RoCo~\cite{mandi2024roco} show how role negotiation and flexible protocols support adaptable teamwork in dynamic environments.

\subsection{Healthcare \& Medicine Agents}
\label{sec:app-healthcare}

Healthcare and medical agents seek to support the full clinical decision pipeline, from initial symptom triage to treatment planning and integrating LLMs with structured patient records, medical ontologies and expert guidelines. Unlike general assistants, these systems must operate under strict safety constraints, multi-modal evidence and legal justification.

In this subsection, we begin with the foundational layer (Section~\ref{sec:app-healthcare-foundational}), which includes medical and diagnostic reasoning and tool-augmented access to various biomedical knowledge bases and APIs. Building on these primitives, the self-evolving layer (Section~\ref{sec:app-healthcare-evolve}) examines memory, feedback and reflective modules that allow these agents to accumulate patient-specific context, adapt to longitudinal trajectories and revise clinical plans over time. Finally, the collective layer (Section~\ref{sec:app-healthcare-multi}) highlights multi-agent collaboration, which includes doctor–agent co-planning, human–AI shared autonomy and specialist model ensembles.

\subsubsection{Foundational agentic reasoning}
\label{sec:app-healthcare-foundational}

\paragraph{Planning.}

Planning is a core capability for healthcare agents, which enables them to structure long-horizon clinical pathways into diagnostic and treatment phases, refine workflows dynamically as patient conditions evolve and coordinate across teams and tools toward cohesive care delivery. We discuss several various recent advancements as follows. For instance, a recent agentic clinical system~\cite{ferber2024autonomousartificialintelligenceagents} orchestrates specialized tools and guideline citations to support oncology decision-making, EHRAgent \cite{wenqi2024ehragent} decomposes multi-table EHR inference into code-execution steps with feedback learning and PathFinder \cite{ghezloo2025pathfinder} presents a multi-agent, multi-modal histopathology workflow for diagnostic reasoning.  

Other frameworks model planning as an explicit orchestration layer across levels of abstraction. For example, MedAgent‑Pro \cite{wang2025medagent} proposes a hierarchical workflow which first generates disease-level diagnostic plans from guideline criteria and then dispatches tool-agent modules for execution. MedOrch \cite{he2025medorchmedicaldiagnosistoolaugmented} treats tool invocation itself as a planning primitive across modalities, orchestrating reasoning agents for multi-step diagnostic execution. On the other hand, ClinicalAgent \cite{yue2024clinicalagentclinicaltrialmultiagent} coordinates multi-agent workflows for clinical planning, leveraging LLM reasoning to allocate tools and synthesize evidence. In addition, planning in healthcare agents is increasingly adaptive, responding to new information and evolving contexts. For example, DoctorAgent‑RL \cite{feng2025doctoragentrlmultiagentcollaborativereinforcement} models clinical consultation as a dynamic decision-making process under uncertainty, optimizing questioning strategies and diagnostic paths via reinforcement learning; while DynamiCare \cite{shang2025dynamicaredynamicmultiagentframework} adjusts specialist-agent teams across multi-round interactions as new patient information emerges.

\paragraph{Tool-use.}

Tool integration significantly expands a healthcare agent’s action space, enabling precise calculations, medical image interpretation and access to specialized databases. Recent studies are summarized as follows. Several systems explicitly foreground extensibility. MedOrch~\cite{he2025medorchmedicaldiagnosistoolaugmented} introduces a modular architecture that allows new diagnostic APIs to be incorporated without retraining, while TxAgent~\cite{gao2025txagentaiagenttherapeutic} integrates over two hundreds pharmacological tools to support therapeutic decision-making across drug–disease–treatment relationships. AgentMD~\cite{jin2024agentmdempoweringlanguageagents} similarly curates and leverages over two thousands executable clinical calculators to learn risk-prediction pipelines.

Other approaches focus on structured function calling for safe execution. For example, LLM-based agents can reliably invoke bedside calculators when provided with explicit function signatures, ensuring arithmetic correctness in dosing and risk scoring~\cite{Goodell_2025}. MeNTi~\cite{zhu2025mentibridgingmedicalcalculator} goes further by enabling nested tool calls across multi-step medical calculators. Complementing these text-based integrations, MMedAgent~\cite{li2024mmedagent} demonstrates that agents can learn to select among multi-modal tools. 

In addition, execution is crucial for translating high-level clinical plans into concrete actions such as code operations, database queries or robotic procedures. VoxelPrompt~\cite{hoopes2024voxelpromptvisionlanguageagentgrounded} embed 3-D volumetric priors so that language instructions drive spatial segmentation and analysis of medical image volumes. On the other hand, embodied ultrasound-robot controllers~\cite{xu2024enhancingsurgicalrobotsembodied} translate LLM-generated plans into closed-loop robotic scanning via a “think-observe-execute” loop.
Adaptive reasoning-and-acting systems~\cite{dutta2024adaptivereasoningactingmedical} further refine both the reasoning and actions over time in simulated clinical environments. In medical imaging, systems like MedRAX~\cite{fallahpour2025medraxmedicalreasoningagent} materializes multi-step reasoning by integrating specialized chest-X-ray tools and LLM reasoning into an end-to-end diagnostic agent. PathFinder~\cite{ghezloo2025pathfinder} similarly executes multi-agent, multi-modal diagnostic workflows in histopathology.

Another class of healthcare agents deploys code-level workflows. For example, Conversational Health Agents~\cite{abbasian2024conversationalhealthagentspersonalized} compile dialogue actions into function calls and code execution for downstream processing, while EHRAgent~\cite{wenqi2024ehragent} materializes EHR operations via executable code. MedAgentGym~\cite{xu2025medagentgymscalableagentictraining} trains agents to produce code that is directly executed and graded, enforcing reliability of reasoning traces. DoctorAgent-RL~\cite{feng2025doctoragentrlmultiagentcollaborativereinforcement} validates multi-turn dialogue acts by executing reinforcement-learned strategies in simulated consultations, while AIPatient~\cite{yu2025simulated} materializes realistic patient scenarios for execution-based evaluation and another recent study~\cite{almansoori2025selfevolvingmultiagentsimulationsrealistic} demonstrates how self-evolving multi-agent simulations allow execution behaviors themselves to improve over time.

\paragraph{Search and retrieval.}
Search-based agents enhance clinical decision-making by linking LLM reasoning with external biomedical knowledge sources. For instance, MeNTi~\cite{zhu2025mentibridgingmedicalcalculator} supplements therapeutic reasoning by bridging LLM calls into multi-step medical calculators while EHRAgent~\cite{wenqi2024ehragent} dynamically executes code operations over multi-table EHR data to support complex tabular inference. Conversational Health Agents~\cite{abbasian2024conversationalhealthagentspersonalized} enrich personalized dialogue by integrating developer-defined external sources and orchestrating action flows. Another line of work explicitly embeds retrieval-augmented generation (RAG) into healthcare agents. For example, CLADD~\cite{lee2025ragenhancedcollaborativellmagents} retrieves molecular graphs and prior assay results before proposing compound hypotheses and MedReason~\cite{wu2025medreasonelicitingfactualmedical} issues targeted knowledge-graph sub-queries to anchor each reasoning step for clinical QA.

\subsubsection{Self-evolving agentic reasoning}
\label{sec:app-healthcare-evolve}

Self-evolving capabilities enable healthcare agents to maintain longitudinal clinical coherence. Representative use cases include accumulating relevant medical context across encounters, updating beliefs as new evidence arrives and revising decisions when inconsistencies surface. In the following paragraphs, we examine how memory, feedback and reflective mechanisms collectively turn clinical reasoning from a one-shot prediction into a continually improving care process.

\paragraph{Memory.}
Persistent memory is essential for tracking medical or patient history and maintaining context across interactions. For instance, epidemic-modeling agents~\cite{williams2023epidemicmodelinggenerativeagents} maintain temporal contact histories to trace infection chains over time; while MedAgentSim~\cite{almansoori2025selfevolvingmultiagentsimulationsrealistic} stores experience histories and refine diagnostic strategies over time. In structured data settings, EHRAgent~\cite{wenqi2024ehragent} records intermediate computations over tabular EHRs so subsequent steps can reference prior results. EvoPatient~\cite{du2025llmssimulatestandardizedpatients} interleaves memory with coevolution maintains evolving clinical state across dialogue phases while AIPatient~\cite{yu2025simulated} persists longitudinal EHR-derived variables to drive consistent responses. Multi-agent systems such as MedOrch~\cite{he2025medorchmedicaldiagnosistoolaugmented} contain clinical knowledge graph agent which can be considered as external memory that can be queried to retrieve known relationships or diagnostic patterns.

\paragraph{Agentic Feedback and Reflection.}
Agentic feedback and self-reflection are complementary mechanisms that improve reliability and adaptability of healthcare agents. Feedback converts execution outcomes into learning signals: DynamiCare~\cite{shang2025dynamicaredynamicmultiagentframework} updates multi-agent treatment strategies when newly observed patient state contradicts prior plans; DoctorAgent-RL~\cite{feng2025doctoragentrlmultiagentcollaborativereinforcement} optimizes questioning policies from consultation rewards; and MedAgentGym~\cite{xu2025medagentgymscalableagentictraining} enforces correctness by executing and grading generated code. Tool-use pipelines also propagate execution feedback. For example, the success/failure of table queries in EHRAgent~\cite{wenqi2024ehragent} or calculator calls in MeNTi~\cite{zhu2025mentibridgingmedicalcalculator} and clinical-calculation agents~\cite{Goodell_2025} to refine subsequent actions.

\subsubsection{Collective multi-agent reasoning}
\label{sec:app-healthcare-multi}

Multi-agent collaboration is central to healthcare AI, since clinical decision-making often depends on consensus among specialists, negotiation of competing hypotheses and coordination across roles such as physicians, patients and trial designers. In the following, we discuss several strands of research centered around multi-agent capabilities.

For collaborative decision-making, notable frameworks include MDAgents~\cite{kim2024mdagents}, which automatically assigns tailored collaboration structures to teams of LLMs depending on medical task complexity, and DoctorAgent-RL~\cite{feng2025doctoragentrlmultiagentcollaborativereinforcement}, which uses a multi-agent reinforcement-learning framework to optimize multi-turn doctor-patient consultation dialogues. In addition, Agent-derived Multi-Specialist Consultation (AMSC)~\cite{wang2024directdiagnosisllmbasedmultispecialist} explores staged multi-specialist dialogues for differential diagnosis that mimics the medical scene of a patient consulting with multiple specialists. Other notable works include ClinicalAgent~\cite{yue2024clinicalagentclinicaltrialmultiagent}, which organizes clinical trial workflows via role-based agent collaboration / LLM reasoning and PathFinder~\cite{ghezloo2025pathfinder}, which integrates a diverse set of agents that can gather evidence and provide comprehensive diagnoses with natural language explanations. 

On the other hand, there are studies focusing on simulation-driven collaboration. These works highlight how multi-agent setups enrich training and evaluation. MedAgentSim~\cite{almansoori2025selfevolvingmultiagentsimulationsrealistic} co-evolves doctor and patient agents to simulate real-world multi-turn clinical interactions, and EvoPatient~\cite{du2025llmssimulatestandardizedpatients} uses co-evolution of patient and doctor agents to generate diagnostic dialogue data and therefore gathers experience to improve the quality of both questions and answers to enable accurate human doctor training. In addition, DynamiCare~\cite{shang2025dynamicaredynamicmultiagentframework} initiates a team of specialist agents that iteratively queries the patient system to integrate new information and adapts the composition and strategy. Finally, medical agents can also collaborative to aid medical reasoning process. For example, MedAgents~\cite{xiangru2023medagents} demonstrates zero-shot cooperation among domain-specialist agents in medical reasoning tasks, CLADD~\cite{lee2025ragenhancedcollaborativellmagents} uses retrieval-augmented generation to support drug-discovery workflows across agents and GMAI-VL-R1~\cite{su2025gmaivlr1harnessingreinforcementlearning} combines multi-modal reasoning and reinforcement learning in a multi-agent framework to support large-scale medical decision-making.

\subsection{Autonomous Web Exploration \& Research Agents}
\label{sec:app-research}

Web agents, GUI agents and autonomous research agents constitute three interlinked but distinct trajectories of agentic reasoning systems. Firstly, web agents specialize in navigating online resources, issuing web API calls or browser actions to retrieve dynamic evidence and steer research direction. GUI agents go further by manipulating software interfaces and multi-modal dashboards directly (i.e. clicking, typing, navigating) to execute experiments, data workflows and interface-based tasks. Autonomous research agents sit at the top of this hierarchy, pairing LLM reasoners with scientific workflows, tool-chains and meta-loops to drive hypothesis generation, data synthesis and paper writing. The core connection is a progression of autonomy: first web agents retrieve evidence from online resources, then GUI agents operationalize actions inside software interfaces, and finally autonomous research agents orchestrate full scientific workflows end-to-end.

In this subsection we begin with the foundational layer (Section \ref{sec:app-auto-foundational}), which captures the core capabilities that any autonomous agent must support: perceiving its environment, reasoning about goals, planning actions and grounding those into tool-augmented workflows. Building on these primitives, the self-evolving layer (Section \ref{sec:app-auto-evolving}) examines how agents incorporate feedback, memory and reflection to iteratively refine their behaviors and improve methods over time. Finally, the collective layer (Section \ref{sec:app-auto-multi-agent}) highlights how agents move beyond individual competence into coordination, specialization and emergent collaboration. While web agents, GUI agents and autonomous agents share common themes of goal-directed autonomy, tool-use and iterative improvement, they differ in where they act on, how they manipulate their environment and what goal they aim to achieve.

\subsubsection{Foundational agentic reasoning}

\label{sec:app-auto-foundational}

\paragraph{Planning.}

Planning is essential for web agents because they must decompose long-horizon tasks into manageable steps, adapt to dynamic pages and coordinate tool/invocation strategies. Early work such as WebGPT~\cite{nakano2021webgpt} fine-tuned GPT-3~\cite{brown2020language} to answer open-ended questions via a text-based web-browser interface. Then, various web-based methods deepened the planning paradigm: for example, SEEACT~\cite{zheng2024gpt} explored large multi-modal models as generalists that integrating visual and HTML grounding for web-based tasks, and AutoWebGLM~\cite{lai2024autowebglm} introduced HTML simplification and various learning techniques for open-domain web task decomposition and navigation. These works paved the way for recent systems such as Agent Q~\cite{putta2024agent} that integrate guided MCTS, self-critique and off-policy preference optimization on web-task benchmarks, and set the stage for even more advanced long-horizon web planners such as WebExplorer~\cite{liu2025webexplorer} and WebSailor~\cite{li2025websailor}.

% rl for web agents
In addition, reinforcement learning has become a core tool for improving the decision-making and planning behavior of web-based LLM agents. WebRL~\cite{qi2024webrl} introduces a self-evolving online curriculum that generates new tasks from unsuccessful attempts and trains an outcome-supervised reward model to guide policy optimization. WebAgent-R1~\cite{wei2025webagent} performs end-to-end multi-turn RL, learning web interaction policies directly from online rollouts with binary success rewards. DeepResearcher~\cite{zheng2025deepresearcher} scales RL to real-world web environments, using a multi-agent browsing architecture and exhibiting emergent behaviors such as plan formulation, cross-source corroboration, and self-reflection. Hybrid pipelines like AutoWebGLM~\cite{lai2024autowebglm} combine supervised training with RL fine-tuning to strengthen task decomposition and structured navigation, while Navigating WebAI~\cite{thil2024navigating} combines supervised learning and RL techniques to improve web navigation performance. Methods such as Pangu DeepDiver~\cite{shi2025pangu}, EvolveSearch~\cite{zhang2025evolvesearch}  and WebEvolver~\cite{fang2025webevolver} use RL-based self-improvement, for example by adaptively scaling search depth or jointly training an agent and a world-model-like simulator to improve long-horizon web decision-making. Hierarchical approaches like ArCHer~\cite{zhou2024archer} optimize high-level and low-level policies with a multi-turn hierarchical RL framework, while PAE~\cite{zhou2025proposer} combines a task proposer, an acting agent and an evaluator to support autonomous skill discovery via RL in internet environments.

% gui agents

% normal planning for gui agents
Planning is a core capability for GUI agents, enabling them to coordinate long, multi-step interactions across applications and operating environments. OS-Copilot~\cite{wu2024copilot} approaches this by treating the desktop as a unified control space in which a generalist agent continually refines its multi-step workflows. Agent S~\cite{agashe2024agent} builds an experience-augmented planning stack that decomposes tasks into sub-goals while retrieving past trajectories and external knowledge to guide action sequencing. InfiGUIAgent~\cite{liu2025infiguiagent} strengthens planning by integrating hierarchical task structuring into a multi-modal backbone, allowing agents to organize GUI procedures at multiple levels of abstraction. MobA~\cite{zhu2025moba} and PC Agent~\cite{liu2025pc} employ hierarchical architectures that separate high-level planning from low-level execution—on mobile and desktop respectively. GUI foundation models such as OS-ATLAS~\cite{wu2024atlas}, OSCAR~\cite{wang2024oscar} and UItron~\cite{zeng2025uitron} further emphasize robust cross-application planning: OS-ATLAS offers a platform-agnostic action model for consistent control, OSCAR maintains state-aware plans that adapt as execution unfolds and UItron unifies offline and online planning within a single general-purpose GUI agent.

Likewise, reinforcement learning has become a central way to endow GUI agents with planning over long action sequences. End-to-end frameworks such as ARPO~\cite{fanbin2025arpoendtoend} and ComputerRL~\cite{hanyu2025computerrl} directly optimize multi-step GUI trajectories with replay buffers or large-scale online interaction, replacing hand-crafted scripts with learned policies for general desktop control. R1-style and semi-online methods, including UI-R1~\cite{zhengxi2025uir1}, GUI-R1~\cite{run2025guir1}, InfiGUI-R1~\cite{yuhang2025infiguir1} and UI-S1~\cite{zhengxi2025uis1}, start from strong vision–language backbones and then use RL to sharpen action prediction and long-horizon reasoning. A complementary line focuses on where to act by improving visual grounding: GUI-Bee~\cite{yue2025guibee}, SE-GUI~\cite{xinbin2025enhancing}, UIShift/GUI-Shift~\cite{gao2025guishiftenhancingvlmbasedgui} and UI-AGILE~\cite{shuquan2025uiagile} develop RL-based grounding frameworks to help agents reliably localize target elements before executing actions. ZeroGUI~\cite{chenyu2025zerogui} pushes toward fully automated online RL loops, where the agent generates its own tasks and trajectories and improves with zero human annotation, while ComputerRL~\cite{hanyu2025computerrl} scales end-to-end online RL in large distributed desktop environments. AgentCPM-GUI~\cite{zhong2025agentcpmgui} couples supervised pre-training with reinforcement fine-tuning to strengthen decision quality on mobile apps. Finally, foundation-style GUI agents such as AutoGLM~\cite{xiao2024autoglm} and Mobile-Agent-v3~\cite{jiabo2025mobileagentv3} serve as general backbones that unify perception, grounding and action, and are trained or fine-tuned with scalable RL frameworks to align long-horizon GUI planning with real-world success signals.

% research agents
For autonomous research agents, planning modules translate abstract goals into actionable research itineraries.  For example, Agent Laboratory~\cite{schmidgall2025agentlaboratoryusingllm} organizes work into three structured stages, namely literature review, experimentation and report writing, and supports the workflow with tool-hooks that automate code execution, experiment runs and documentation. GPT Researcher~\cite{Elovic2023gptresearcher} uses a plan → research → write cycle, where a dedicated planner drafts the outline, retrieval/analysis agents gather evidence and a writer compiles the final report. Chain of Ideas~\cite{li2024chainideasrevolutionizingresearch} retrieves literature into a chain structure to reflect domain progression and support ideation via experiment design, whereas IRIS~\cite{garikaparthi2025iris} performs hypothesis exploration via Monte Carlo Tree Search to expand promising branches before committing to downstream tasks. Broader variants include ARIA~\cite{ramirezmedina2025acceleratingscientificresearchmultillm} and NovelSeek~\cite{novelseekteam2025novelseekagentscientist} that automate the research workflow with a complete literature search, hypothesis generation and experiment planning cycle.

\paragraph{Tool-Use.}

% web agents
For web agents, tool-use abilities underpins execute plans in realistic, dynamic environments. For example, WebVoyager~\cite{he2024webvoyager} systematizes multi-modal execution by building an end-to-end agent that operates on real websites. On the interaction side, BrowserAgent~\cite{zhang2025browseragent} makes the action space more human-like, defining a compact set of browser primitives (e.g., click, scroll, type) and coupling them with an explicit memory mechanism to maintain key conclusions across steps, yielding strong gains on multi-hop QA benchmarks. Finally, methods like WALT~\cite{prabhu2025walt} and pipeline-oriented systems such as WebDancer~\cite{wu2025webdancer} and WebShaper~\cite{tao2025webshaper} push tool use from mere execution toward tool discovery and data-centric interaction. Specifically, WALT teaches agents to reverse-engineer reusable tools from website functionality, while WebDancer and WebShaper embed web actions inside multi-turn information-seeking and dataset-synthesis loops, respectively.

% gui agent tool-use
Tool use is another core capability for GUI agents, enabling them to invoke system functions and application features as structured tools. As pioneering systems, AutoDroid~\cite{wen2024autodroid} automatically analyzes Android apps to construct functionality-aware UI abstractions that LLM agents can reason over as capabilities rather than raw layouts, while its successor AutoDroid-V2~\cite{wen2025autodroid} re-frames mobile UI automation as LLM-driven code generation, with an on-device small language model emitting executable scripts for a local interpreter. MobileExperts~\cite{zhang2024mobileexperts} models each expert as a tool-capable specialist and uses a dual-layer controller to select which expert and its associated tool-set to invoke at different stages of a mobile workflow. AgentStore~\cite{jia2025agentstore} pushes this idea to the platform level by treating heterogeneous agents themselves as tools: a MetaAgent uses AgentTokens to route operating-system subtasks to the most suitable specialized “tool-agent” through a unified interface. OS-Copilot~\cite{wu2024copilot} and OSCAR~\cite{wang2024oscar} integrate rich system-level tools into unified computer-control frameworks, so that complex desktop tasks are expressed as sequences of tool calls. OS-ATLAS~\cite{wu2024atlas} complements these systems with a foundation action model that offers robust cross-platform GUI grounding, serving as a reliable actuator layer for downstream tool-using agents. Finally, SeeClick~\cite{cheng2024seeclick} strengthens the execution stack by pre-training a visual GUI agent for GUI grounding, improving the ability to locate the correct on-screen elements from instructions.

Specialized tools can expand an autonomous research agent’s capabilities beyond pure text, allowing more fine-grained ability. For instance, Agentic Reasoning~\cite{wu2025agentic} automatically routes queries to appropriate tool modules like code execution, web search and structured memory agents when the main LLM detects a gap in reasoning; while Webthinker~\cite{li2025webthinkerempoweringlargereasoning} empowers autonomous web exploration and page navigation during long-horizon investigations, by interleaving reasoning, search and draft-writing with a web explorer module. PaperQA~\cite{lála2023paperqaretrievalaugmentedgenerativeagent} and its follow-up synthesis agent~\cite{skarlinski2024languageagentsachievesuperhuman} integrate PDF parsing and citation-level grounding to produce verifiable answers and literature syntheses, while Scideator~\cite{radensky2025scideatorhumanllmscientificidea} provides an IDE-style tool-chain that combines paper facets with novelty checks for real-time brainstorming. In addition, DeepResearcher~\cite{zheng2025deepresearcher} shows that reinforcement learning over real-web interactions improves deep-research efficiency and quality, with emergent behaviors such as plan refinement and cross-source corroboration.

Execution components ground high-level reasoning in code, simulations or laboratory protocols to produce verifiable scientific outcomes. Agent Laboratory~\cite{schmidgall2025agentlaboratoryusingllm} executes experiments specified in declarative configuration files by orchestrating external toolchains, while Agentic Reasoning~\cite{wu2025agentic} integrates a coding agent that executes Python alongside web search and structured memory, feeding the results back into the reasoning process. MLR-Copilot~\cite{li2024mlrcopilotautonomousmachinelearning} turns research plans into runnable implementations via an ExperimentAgent that leverages retrieved prototype code, runs experiments, and iteratively debugs implementations. Dolphin~\cite{yuan2025dolphin} closes the loop by generating ideas, implementing them through code templates with traceback-guided debugging, executing experiments, and using the analyzed results to steer the next research cycle. The AI Scientist~\cite{chris2024the} automates end-to-end ML experiments, i.e. generating ideas, writing code, executing experiments and visualizing results, so that observed outcomes can guide subsequent runs, while The AI Scientist-v2~\cite{yamada2025ai} adds a dedicated experiment manager and progressive agentic tree search to prioritize and schedule experiment branches. Most recently, NovelSeek~\cite{novelseekteam2025novelseekagentscientist} introduces a unified closed-loop multi-agent framework that spans hypothesis generation, idea-to-methodology construction, and multi-round automated experiment execution with feedback across diverse scientific domains.

\paragraph{Search and Retrieval.}

Search and retrieval lie at the heart of what differentiates web agents from static language models: they must locate, synthesize and refactor web-scale information in dynamic environments. WebExplorer~\cite{liu2025webexplorer} tackles this by generating challenging information-seeking trajectories and training agents to interleave a search tool and a browse tool over many turns, resulting in improved multi-step retrieval policies on complex benchmarks. WebSailor~\cite{li2025websailor} likewise focuses on information-seeking under extreme uncertainty, constructing high-uncertainty search tasks and using a two-stage post-training pipeline  to instill uncertainty-reducing search strategies for long-horizon web tasks. INFOGENT~\cite{reddy2025infogent} also performs  multi-query search across diverse web sources, enabling comprehensive information retrieval beyond task completion. For retrieval-augmented generation applications, RaDA~\cite{kim2024rada} explicitly disentangles web-agent planning into Retrieval-augmented Task Decomposition and Retrieval-augmented Action Generation, so that each high-level subgoal and concrete action is conditioned on fresh search results while respecting context limits. In addition, GeAR~\cite{shen2025gear} advances retrieval itself by augmenting a base retriever with graph expansion and an agent framework, enabling multi-hop passage retrieval along graph-structured evidence chains. Finally, WebRAGent~\cite{anonymous2025webragent} exemplifies retrieval-augmented generation for web agents by retrieving past trajectories and external knowledge into a multi-modal RAG policy.

% gui agents
Several GUI agents use retrieval capability to inject external experience or knowledge at inference time. Synapse~\cite{zheng2023synapse} maintains an exemplar memory of abstracted trajectories and, for each new task, retrieves similar past trajectories as in-context plans, substantially improving multi-step decision-making. LearnAct~\cite{liu2025learnact} builds a three-agent pipeline that mines human demonstrations into a knowledge store and retrieves the most relevant instructions to guide mobile GUI execution on unseen and diverse tasks. MobileGPT’s Explore–Select–Derive–Recall framework~\cite{lee2023explore} equips a phone agent with human-like app memory, storing modular procedures that can be recalled and recomposed when similar tasks reappear. TongUI~\cite{zhang2025tonguiinternetscaletrajectoriesmultimodal} turns large-scale multi-modal web tutorials into the GUI-Net trajectory corpus, effectively giving agents a large offline memory of how humans operate hundreds of apps across multiple operating systems. RAG-GUI~\cite{xu2025retrieval} makes retrieval explicit at inference time by querying web tutorials and generating textual guidelines that are fed into any VLM-based GUI agent as step-by-step hints. WebRAGent~\cite{anonymous2025webragent} shows a related pattern in web automation, combining a multi-modal retriever with a web agent so that each action is conditioned on retrieved guidance.

% research agents
Search modules probe the research landscape to surface relevant papers and passages, enriching context and grounding subsequent reasoning. WebThinker~\cite{li2025webthinkerempoweringlargereasoning} equips large reasoning models with a Deep Web Explorer module for autonomous web search and page navigation, and uses an Autonomous Think–Search–and–Draft strategy with RL-based training to decide when to browse and what to extract during long-horizon tasks. DeepResearcher~\cite{zheng2025deepresearcher} scales end-to-end training on the real web via reinforcement learning in live search environments, optimizing the iterative think–search loop and exhibiting emergent behaviors such as plan formulation, cross-source corroboration, and self-correction over multi-step research trajectories. Retrieval-centric agents like PaperQA~\cite{lála2023paperqaretrievalaugmentedgenerativeagent} and its successor PaperQA2~\cite{skarlinski2024languageagentsachievesuperhuman} demonstrate that tightly coupling full-text retrieval with generation can substantially improve scientific QA accuracy while preserving cited provenance for literature synthesis.

In research settings, retrieval-augmented generation (RAG) grounds idea generation and analysis in freshly retrieved and citable passages. For example, GPT Researcher~\cite{Elovic2023gptresearcher} is an autonomous research-agent that retrieves sources and generates reports with citations, enabling traceability of claims to evidence. Chain of Ideas~\cite{li2024chainideasrevolutionizingresearch} organizes relevant literature into a chain-structured scaffold that mirrors a field’s progressive development, thereby guiding retrieval and ideation toward subsequent links in the argument. Meanwhile, Scideator~\cite{radensky2025scideatorhumanllmscientificidea} extracts key paper facets (e.g., purpose, mechanism, evaluation) and leverages them to drive targeted retrieval and recombination of ideas for identifying methodological or evidentiary gaps.

\subsubsection{Self-evolving agentic reasoning}
\label{sec:app-auto-evolving}

Effective self-evolving abilities enable these autonomous agents to adapt their behavior over time, retain crucial task context across interaction cycles and incrementally refine planning and execution strategies. The following paragraphs review how memory, feedback and self-reflection mechanisms support this continual improvement across these agent families, turning interaction from a one-shot pipeline into an iterative learning loop.

% web agents
\paragraph{Memory.}

Memory modules transform brittle, single-pass web interactions into reusable experience. For example, Agent Workflow Memory (AWM)~\cite{Wang2024AgentWM} induces reusable workflows from successful trajectories and retrieves them to guide future tasks, while ICAL~\cite{sarch2024vlm} distills noisy trajectories into high-level verbal and visual abstractions that are stored as a memory of multimodal experience and later injected into prompts. Control-oriented designs such as BrowserAgent~\cite{zhang2025browseragent} maintain explicit histories of past actions and intermediate conclusions in the agent’s context, instead of only re-encoding the current page view. GLM-based agents like AutoWebGLM~\cite{lai2024autowebglm} and AgentOccam~\cite{yang2024agentoccam} emphasize compressed page representations, using HTML simplification and carefully tuned observation spaces so that the agent’s prompt contains a shorter, more informative view of the state, with past steps preserved through the usual action–observation history. More integrated frameworks like LiteWebAgent~\cite{zhang2025litewebagent} expose planning, memory and tree search as modular components, and can plug in workflow memories together with search traces for long-horizon reuse.

% gui agents
Recent GUI agents adopt explicit memory modules that store and retrieve task-relevant information during long-horizon execution. Earlier work such as MobileGPT~\cite{lee2024mobilegpt} equips a mobile assistant with human-like app memory: it decomposes procedures into modular sub-tasks that are explored, selected, derived, and then stored so they can be recalled and reused instead of being re-discovered from scratch. Chain-of-Memory (CoM)~\cite{gao2025chain} incorporates short- and long-term memory by recording action descriptions and task-relevant screen information in a dedicated memory module, enabling cross-application navigation to track task state. More recent systems build increasingly structured memories: MobA’s multifaceted memory module~\cite{zhu2025moba} maintains environment- and user-level traces that an adaptive planner retrieves when refining mobile task plans, while MGA~\cite{cheng2025mga} represents each step as a triad of current screenshot, spatial layout, and a dynamically updated structured memory that summarizes past transitions, mitigating error accumulation in long chains of actions. Mobile-Agent-E~\cite{wang2025mobileagente} adds a persistent long-term store of tips and shortcuts distilled from prior trajectories, so later plans can call reusable guidance and subroutines instead of relearning them. Mirage-1~\cite{xie2025mirage} similarly organizes experience into a hierarchical skill memory that a planner can retrieve as reusable building blocks for new GUI tasks.

% research agents
Long-term memory is crucial for autonomous research agents because it enables accumulation and reuse of prior knowledge, fostering continuity across research cycles. For example, Agent Laboratory~\cite{schmidgall2025agentlaboratoryusingllm} retains prior experiment code, results, and interpretation across its multi-phase workflow, enabling later stages to build on earlier work. GPT Researcher~\cite{Elovic2023gptresearcher} generates reports with embedded citations and provides context for planning and extension of research topics. Chain of Ideas~\cite{li2024chainideasrevolutionizingresearch} structures relevant literature into a chain scaffold that reflects a field’s progression and can be revisited as new evidence arises. The AI Scientist-v2~\cite{yamada2025ai} incorporates a progressive agentic tree-search approach that enables branching, backtracking and follow-up experimentation across iterations.

\paragraph{Agentic Feedback and Reflection.}

Modern web agents treat interaction as a continual learning process, using feedback signals and reflection modules to refine their reasoning and recover from failures over time. Agent Q~\cite{putta2024agent} combines guided Monte Carlo tree search with a self-critique stage, so that rollouts provide not only action sequences but also preference-style supervision. \textsc{ReAP}~\cite{azam2025reflection} makes reflection explicit by treating it as a retrieval problem: it stores task–reflection key–value pairs summarizing what was learned from past trajectories, then, at inference time, retrieves the most relevant reflections and appends them to the agent’s prompt to guide planning on new web-navigation tasks. Agent-E~\cite{azam2024multimodal} introduces an automatic validation pipeline that detects execution errors across text and vision, and then triggers self-refinement, enabling agents to iteratively correct their own workflows. Recon-Act~\cite{he2025recon} uses a dual-team architecture in which a Reconnaissance team extracts generalized tools from successful and failed trajectories, and an Action team applies these tools to re-plan tasks, forming a closed feedback loop. INFOGENT~\cite{reddy2025infogent}, on the other hand, leverages aggregator feedback to iteratively refine navigation and search strategies based on identified information gaps. And WINELL~\cite{reddy2025winell}, as a updating web agent, relies on feedback from the aggregation process to adapt subsequent searches and update selection during continuous operation. Finally, self-reflective search agents such as WebSeer~\cite{he2025webseer} integrate explicit self-reflection signals into reinforcement learning, constructing reflection-annotated trajectories and a two-stage training framework so that mis-solved or uncertain cases become targeted feedback that deepens future search and reasoning.

GUI agents also integrate explicit reflection so they can critique and repair their own plans. Early computer-control systems with structured reflection, for example, a zero-shot desktop control agent with structured self-reflection loops~\cite{li2023zero}, provides conceptual templates that later GUI agents adapt to visual, multi-application settings. GUI-Reflection~\cite{wu2025gui} instantiates this idea end-to-end: it builds a reflection-oriented task suite, automatically synthesizes error scenarios from existing successful trajectories, and adds an online reflection-tuning stage so multi-modal GUI models learn to detect failures, reason about causes, and generate corrective actions without human annotation. History-Aware Reasoning (HAR)~\cite{wang2025history} treats long-horizon GUI automation as a reflective learning problem, constructing reflective learning scenarios, synthesizing tailored correction guidelines, and designing a hybrid RL reward so the agent acquires episodic reasoning knowledge from its own errors and shifts from history-agnostic to history-aware reasoning. MobileUse~\cite{li2025mobileuse} introduces hierarchical reflection on mobile devices, where the agent self-monitors at the action, subtask, and task level and triggers reflection on demand, pausing only when needed to diagnose and recover. InfiGUIAgent~\cite{liu2025infiguiagent} integrates hierarchical and expectation–reflection reasoning in a second training stage, enabling the agent to run expectation–reflection cycles that compare expected and actual outcomes and revise multi-step plans when they diverge. Mobile-Agent-E~\cite{wang2025mobileagente} embeds an Action Reflector and Notetaker that evaluate executed steps and write refined Tips and Shortcuts back into persistent long-term memory, forming a self-evolution loop where the agent’s behavior is progressively refined from accumulated experience.

For autonomous research agents, learning from outcomes is essential to improve reasoning and experimental reliability over time. CycleResearcher~\cite{weng2025cycleresearcherimprovingautomatedresearch} couples a research agent with a reviewer agent that provides automated peer-review feedback, and uses an iterative preference-training loop so the research agent can refine future drafts and decisions. MLR-Copilot~\cite{li2024mlrcopilotautonomousmachinelearning} monitors execution results and human comments during experiment implementation and execution, using these signals to iteratively refine code, configurations and even upstream hypotheses. Dolphin~\cite{yuan2025dolphin} implements a closed-loop auto-research framework in which generated code is run on benchmarks and exception-guided debugging plus outcome analysis feed back into idea generation and implementation, pruning unproductive paths. At the search–reasoning interface, DeepResearcher~\cite{zheng2025deepresearcher} optimizes query, browsing, and answering policies via reinforcement learning on real-web trajectories, with outcome rewards inducing behaviors such as planning, cross-validation, and self-reflection. Agentic Deep Research~\cite{weizhi2025from} further emphasizes reward design for reasoning-driven search, arguing that principled incentives over answer quality and reasoning traces provide structured signals that improve downstream synthesis in deep-research agents.

\subsubsection{Collective multi-agent reasoning}
\label{sec:app-auto-multi-agent}

% web agent
Collective multi-agent reasoning for web agents reframes browser use as cooperation among specialized roles rather than a single monolithic policy. WebPilot~\cite{zhang2025webpilot} models web task execution as a multi-agent system with a global planning agent that decomposes tasks and local MCTS-based executors that solve subtasks, jointly steering search in complex web environments. INFOGENT~\cite{reddy2025infogent} organizes web information aggregation into a Navigator, Extractor, and Aggregator, so exploration, evidence extraction, and synthesis are handled by distinct cooperating agents with feedback from the Aggregator to guide future navigation; WINELL~\cite{reddy2025winell} leverages agentic web search to plan and execute iterative information gathering for discovering timely factual updates relevant to a target Wikipedia article.. Recon-Act~\cite{he2025recon} adopts a Reconnaissance–Action paradigm in which a Recon team analyzes successful and failed trajectories to derive generalized tools or hints, and an Action team re-plans and executes with this evolving toolset. PAE~\cite{zhou2025proposer} uses three roles, namely a task proposer, an acting web agent and a VLM-based evaluator, to autonomously generate vision-based web tasks and feed success signals back into the policy via RL. Hierarchical web agents such as Agent-E~\cite{abuelsaad2024agent} and Plan-and-Act~\cite{wang2023plan} similarly separate a high-level planner from a browser-navigation agent, enabling structured plan–execute cooperation. At a more conceptual level, Agentic Web~\cite{yang2025agentic} envisions the internet as an agentic web of interacting agents and analyzes how coordination, communication protocols and economic incentives shape such ecosystems, while Agentic Deep Research~\cite{weizhi2025from} frames information seeking as iterative feedback loops of reasoning, retrieval, and synthesis that can be instantiated by single- or multi-agent web research systems.

% gui agent
Multi-agent designs for GUI agents typically decompose “using a computer” into cooperating roles that plan, perceive, decide and execute. COLA~\cite{zhao2025cola} instantiates a scenario-aware task scheduler, a planner, a decision-agent pool, an executor, and a reviewer, so UI tasks are split into basic capability units and routed to domain-specialized agents rather than a single monolith. On mobile, Mobile-Agent-v2~\cite{wang2024mobile} adopts a tri-role pattern with planner, decision, and reflection agents for progress navigation, local action selection, and error correction, while Mobile-Agent-E~\cite{wang2025mobileagente} further builds a hierarchical stack with a Manager and four subordinate agents (i.e. Perceptor, Operator, Action Reflector, Notetaker) plus a self-evolution module that learns long-term Tips and Shortcuts from experience. Mobile-Agent-V~\cite{wang2025mobileagentvvideoguidedapproacheffortless} similarly employs a video agent, decision agent, and reflection agent to coordinate multi-modal perception and execution, and MobileExperts~\cite{zhang2024mobileexperts} dynamically forms teams of expert agents with a dual-layer planner that allocates subtasks to tool-specialized experts. SWIRL~\cite{lu2025swirl} makes this structure explicit for RL, training a Navigator that converts language and screen context into structured plans and an Interactor that grounds those plans into atomic GUI actions within a multi-agent RL workflow. PC Agent~\cite{liu2025pc} uses separate planning and grounding agents in a two-stage pipeline for desktop automation, illustrating how multi-agent decomposition can improve long-horizon PC control. 

% research agent
To facilitate autonomous research agents, multi-agent collaboration enables a single model’s linear workflow to become a coordinated research group: specialized agents operate in parallel, exchange intermediate artifacts through explicit interfaces, and provide adversarial or complementary feedback to improve both creativity and rigor. For example, AgentRxiv~\cite{schmidgall2025agentrxivcollaborativeautonomousresearch} coordinates author, reviewer, and editor agents that iteratively refine manuscripts and share evolving artifacts across virtual “labs.” ARIA~\cite{ramirezmedina2025acceleratingscientificresearchmultillm} instantiates a role-structured multi-LLM team that searches, filters, and synthesizes scientific literature into actionable experimental procedures. Earlier multi-agent designs such as CAMEL~\cite{qi2023largelanguagemodelszero} demonstrate how cooperative role-play with tool access can enhance hypothesis generation and task decomposition. In experimental sciences, Coscientist~\cite{boiko2023autonomous} integrates planning, robotic instrument control, and analysis into a multi-agent closed loop that autonomously designs and executes wet-lab experiments. Finally, TAIS~\cite{liu2024toward} defines a hierarchical team, namely project manager, data engineer and domain expert, that jointly discovers disease-predictive genes from expression data through coordinated division of labor.

\addtocontents{toc}{\protect\setcounter{tocdepth}{3}}

\section{Benchmarks}
\label{sec:benchmarks}

Agentic reasoning has been evaluated through a rapidly growing set of benchmarks, but existing suites often differ in what they treat as the core capability, such as tool invocation accuracy, memory retention under long contexts, or coordination quality in multi-agent settings. To provide a coherent view, we organize benchmarks from two complementary perspectives. We first summarize benchmarks that isolate core mechanisms of agentic reasoning, which helps pinpoint where systems succeed or fail at the capability level. We then review application-level benchmarks that evaluate end-to-end agent behavior in realistic domains, capturing the combined effects of perception, planning, tool use, memory, and coordination.

\subsection{Core Mechanisms of Agentic Reasoning}

We begin with benchmarks that target mechanism-level capabilities, aiming to evaluate agentic reasoning in a more controlled and interpretable manner. Concretely, these benchmarks decompose agentic behavior into a small set of recurring primitives, including tool use, search, memory and planning, and multi-agent coordination. Such mechanism-centric evaluations make it easier to attribute performance changes to specific components, and they complement end-to-end benchmarks that may conflate multiple sources of errors.

\begin{figure}
    \centering
    \includegraphics[width=\linewidth]{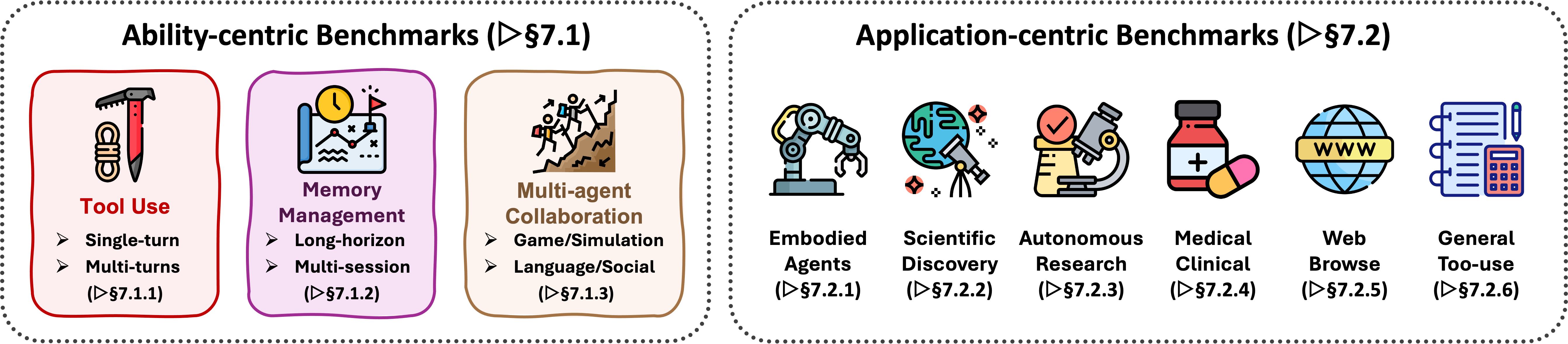}
    \caption{An overview of the benchmarks on agentic reasoning.}
    \label{fig:benchmark}
\end{figure}

\subsubsection{Tool Use}
Evaluating tool-using models remains an open challenge due to the diversity of tasks, tools, and usage scenarios involved \cite{DBLP:journals/csur/QinHLCDCZZHXHFSWQTZLSXZ25}. The key difficulties arise from the wide range of available tools, varying levels of scenario complexity, and the prevalence requirements specifically for the task domain.

\paragraph{Single-Turn Tool Use.}
While agentic reasoning often focuses on multi-turn or long-horizon interactions, single-turn tool use remains a foundational capability for evaluating LLMs' basic tool invocation skills. ToolQA \cite{DBLP:conf/nips/ZhuangYWSZ23} constructs a dataset of 1,530 dialogues involving 13 specialized tools, designed to assess LLMs' ability to interface with external knowledge sources in a question-answering context. APIBench~\cite{patil2024gorilla} introduces a large-scale benchmark grounded in real-world APIs from HuggingFace, TorchHub, and TensorHub, comprising 1,645 unique APIs and 16,450 instruction–API pairs. It is used to train and evaluate Gorilla, an LLM capable of invoking a broad range of APIs, emphasizing generalization across diverse tool interfaces. ToolLLM-ToolBench~\cite{DBLP:conf/iclr/QinLYZYLLCTQZHT24} curates 16,464 real-world APIs across 49 categories from the RapidAPI Hub, and uses ChatGPT to generate diverse, instruction-style prompts for these APIs. The benchmark is used to train ToolLLaMA, a model that demonstrates strong tool-use capabilities and exhibits promising generalization to unseen APIs. MetaTool \cite{DBLP:conf/iclr/HuangSLFWZ000G024} introduces the TOOLE dataset, containing over 20,000 entries and a benchmark comprising approximately 200 tools across diverse scenarios, including software engineering, finance, and art design. It splits tool selection tasks to tool selection with similar choices, tool selection in specific scenarios , tool selection with possible reliability issues, and multi-tool selection. T-Eval \cite{DBLP:conf/acl/ChenDZLLZZZLCZ24} decomposes tool utilization into a series of sub-processes: instruction following, planning, reasoning, retrieval, understanding, and review, and evaluates each step individually to provide a fine-grained assessment of tool-use capabilities. The benchmark includes a total of 23,305 test cases spanning 15 different tools. GTA (General Tool Agents)~\cite{jize2024gta} targets realistic tool-use scenarios by emphasizing real user queries, real-world deployed tools, and multimodal inputs. It introduces 229 challenging tasks grounded in practical applications, spanning 14 tools across diverse domains. ToolRet \cite{DBLP:journals/corr/abs-2503-01763} focuses specifically on the task of tool retrieval, introducing a heterogeneous benchmark consisting of 7.6K diverse retrieval tasks and a corpus of 43K tools.

\paragraph{Multi-Turn Tool Use.}
Multi-turn tool use offers a more realistic simulation of real-world applications, where agents autonomously select and sequence tools to solve complex tasks. ToolAlpaca \cite{DBLP:journals/corr/abs-2306-05301} is one of the earliest efforts in this direction, using multi-agent simulations to generate 3,938 tool-use instances from over 400 real-world APIs across 50 distinct categories. SambaNova-ToolBench \cite{DBLP:journals/corr/abs-2305-16504} introduces a benchmark centered on software tool manipulation for real-world tasks, with varying levels of API complexity to test agent capabilities. API-Bank \cite{DBLP:conf/emnlp/LiZ000YLHL23} provides a dataset of 1,888 tool-use dialogues from 2,138 APIs, along with a runnable evaluation system containing 73 APIs and 314 tool-use test cases.  UltraTool \cite{DBLP:conf/acl/HuangZLZGLHZWSJ24} evaluates tool-use capabilities across six dimensions: planning awareness, planning ability, creation, tool-use awareness, tool selection, and tool usage. The benchmark spans 22 domains, includes 2,032 tools, and provides 5,824 evaluation samples. ToolFlow distinguishes itself from prior benchmarks by emphasizing long-term planning. It features 224 expert-curated tasks involving 107 real-world tools, highlighting challenges in goal decomposition and multi-step decision-making. More recently, MTU-Bench \cite{DBLP:conf/iclr/WangWWLSPDZWPZG25} presents a multi-granularity benchmark for multi-turn, multi-tool scenarios, and releases MTU-Instruct, a large-scale instruction dataset containing 54,798 dialogues involving 136 tools. m \& m's introduces a benchmark with over 4,000 multi-step, multimodal tasks involving 33 tools, including multimodal models, public APIs, and image processing modules. It also provides a high-quality subset of 1,565 task plans that are human-verified and executable end-to-end.

\subsubsection{Search}

To systematically assess an agent’s ability to acquire information through interaction, recent benchmarks cast search as a sequential reasoning problem and can be broadly categorized into unimodal and multimodal settings, differing in the nature of evidence sources, interaction spaces, and grounding requirements.

\paragraph{Unimodal Search.} Recent benchmarks for single-modal agentic search increasingly frame information seeking as a sequential, decision-driven process, emphasizing planning, interaction, and evidence synthesis. For example, WebWalker~\cite{wu2025webwalker} emphasizes structured website traversal, explicitly modeling search as coordinated horizontal exploration and vertical drilling across interconnected pages. To reflect realistic open-world information seeking, InfoDeepSeek~\cite{xi2025infodeepseek} introduces a dynamic Web setting with verifiable yet non-curated answers, highlighting robustness to noise and distributional shift. Several benchmarks scale search along temporal and informational dimensions: Mind2Web 2~\cite{gou2025mind2web2} focuses on long-horizon browsing and citation-grounded synthesis, whereas RAVine~\cite{xu2025ravine} augments answer quality with process-level efficiency and interaction fidelity. Complementarily, WideSearch~\cite{wong2025widesearch} and DeepWideSearch~\cite{lan2025deepwidesearch} distinguish between breadth-oriented large-scale fact aggregation and depth-oriented multi-hop reasoning, revealing the difficulty of jointly optimizing coverage and reasoning coherence. Domain-specific benchmarks further stress reliability under strict correctness constraints: MedBrowseComp~\cite{chen2025medbrowsecomp} targets clinical decision support by requiring agents to integrate heterogeneous and potentially conflicting medical evidence, while FinAgentBench~\cite{choi2025finagentbench} evaluates retrieval-centric reasoning in financial analysis through document-type selection and fine-grained passage localization. Finally, LocalSearchBench~\cite{he2025localsearchbench} grounds agentic search in real-world local services, evaluating multi-constraint, multi-entity reasoning over large structured databases. Collectively, these benchmarks redefine agentic search evaluation around planning depth, interaction quality, evidence integration, and real-world fidelity, providing a more holistic assessment of search-centric reasoning in language-based agents.

\paragraph{Multimodal Search.} Recent benchmarks on multimodal agentic search move beyond static multimodal question answering to systematically evaluate an agent’s ability to actively retrieve, browse, and reason over heterogeneous information sources under realistic constraints. Benchmarks such as MMSearch~\cite{jiang2024mmsearch} and its extension MMSearch-Plus~\cite{tao2025mmsearch} frame multimodal search as an end-to-end process, where agents must interpret multimodal queries and synthesize answers by jointly leveraging textual and visual evidence, explicitly modeling different input–output modality configurations. Complementing this setting, MM-BrowseComp~\cite{li2025mm} adapts the “hard-to-find, easy-to-verify” paradigm to multimodal web environments, enforcing mandatory image dependence to prevent text-only shortcuts and to stress-test multimodal evidence grounding during open-web browsing. BEARCUBS~\cite{song2025bearcubs} further emphasizes computer-using agents in live web scenarios, requiring explicit interaction trajectories and multimodal manipulation (e.g., videos or 3D navigation), thereby evaluating not only retrieval accuracy but also procedural competence. Moving into domain-specific and tool-augmented regimes, PaperArena~\cite{wang2025paperarena} evaluates multimodal agentic search in scientific workflows, where agents must coordinate PDF parsing, figure understanding, database queries, and web search to answer research-level questions. Finally, Video-BrowseComp~\cite{liang2025video} and VideoDR~\cite{liu2026videodr} extend agentic search to video-centric settings, requiring agents to extract visual-temporal cues from videos and iteratively validate hypotheses via open-web evidence, with carefully designed constraints to ensure dual dependence on video and external retrieval. Together, these benchmarks delineate a clear evolution toward evaluating multimodal agents as interactive researchers, highlighting planning, tool use, and multimodal evidence integration as first-class capabilities in agentic search.

\subsubsection{Memory and Planning}

A distinctive advantage of agents lies in their ability to leverage memory to achieve accurate long-term performance and strong reasoning capabilities. This ability can be assessed from two complementary perspectives. The first concerns memory management, which reflects how effectively an agent integrates, organizes, and retrieves long-term memories. The second concerns memory utilization, which captures how well an agent exploits historical information to support planning and informed feedback. In this section, we separately discuss benchmarks from these two aspects.

From the perspective of memory management, existing benchmarks can be broadly categorized into \textit{Long-Horizon Episodic Memory} and \textit{Multi-session Recall}, depending on whether the textual context consists of a single continuous long-form input or multiple discontinuous conversational sessions.

\paragraph{Long-Horizon Episodic Memory.}
This category targets single-episode tasks with partial observability and delayed rewards, requiring agents to store and retrieve information over extended time spans. Benchmarks in this space evaluate memory retention, retrieval, and reasoning across long contexts. PerLTQA \cite{du2024perltqa} simulates personalized dialogue, where agents answer questions using long-term persona and event memories. It includes 8.5K QA pairs and evaluates memory classification, retrieval ranking, and synthesis fidelity. ELITR-Bench \cite{thonet2024elitr} tests QA on noisy meeting transcripts, where relevant evidence may appear far earlier than the query. Models are scored via GPT-4 across various ASR noise levels and dialogue settings. In the meanwhile, Multi-IF \cite{he2024multi} and MultiChallenge \cite{sirdeshmukh2025multichallenge} focus on multi-turn instruction following. Multi-IF \cite{he2024multi} spans 4.5K tri-turn conversations in 8 languages, with evaluation based on strict and relaxed instruction accuracy. MultiChallenge \cite{sirdeshmukh2025multichallenge} tests four memory-intensive phenomena: retention, inference, editing, and coherence, using 273 curated dialogues with binary pass/fail evaluation. TurnBench-MS \cite{zhang2025turnbench} evaluates multi-step reasoning across 540 symbolic logic games, tracking win rate, round-level accuracy, and verifier usage. StoryBench \cite{wan2025storybench} casts memory as decision-making in interactive narratives, where agents must remember prior choices to progress. It assesses decision accuracy, retry counts, and runtime efficiency. MemBench \cite{tan2025membench} tests factual and reflective memory across 60K episodes in participatory and observational settings, with metrics for accuracy, recall, capacity, and retrieval speed. MMRC~\cite{xue2025mmrc} develops a multimodal memory benchmark focused on single-round multimodal conversations. Together, these benchmarks emphasize structured memory demands, with metrics capturing not just task success but also memory precision, synthesis quality, and robustness under long-context stress.

\paragraph{Multi-session Recall.}
Multi-session Recall focuses on multi-episode tasks where agents must retain and integrate knowledge across separate sessions, supporting lifelong adaptation and mitigating catastrophic forgetting. A range of recent benchmarks systematically probe this capability under realistic, long-term interaction scenarios. LOCOMO \cite{maharana2024evaluating} evaluates LLM agents on sustained conversational memory across 19-session dialogues, using tasks such as multi-hop QA, event summarization, and multi-modal response generation. MemSim \cite{zhang2024memsim} introduces a simulator-based framework with over 2,900 synthetic trajectories in daily life domains, assessing fact retention across sessions via accuracy, diversity, and rationality scores. LONGMEMEVAL \cite{wu2024longmemeval} benchmarks assistants on five sub-tasks: information extraction, multi-session reasoning, temporal inference, knowledge updating, and abstention, over dialogue histories spanning up to 1.5M tokens, with GPT-4 judged accuracy and retrieval recall. REALTALK \cite{lee2025realtalk} presents 21-day real human conversations with 17K tokens per dyad, enabling evaluation of memory probing and persona simulation through multi-hop QA and emotional grounding metrics. Furthermore, MemoryAgentBench \cite{hu2025evaluating} unifies diverse memory tasks such as test-time learning, conflict resolution, and long-range understanding across multiple datasets, with task-specific metrics including classification accuracy, partial-match F1, and ROUGE. Mem-Gallery~\cite{bei2026mem} introduces a multimodal long-term memory evaluation benchmark that systematically covers a wide range of memory management and utilization scenarios. Lastly, Evo-Memory \cite{wei2025evo} introduces a benchmark and a unified evaluation protocol for measuring experience reuse in test-time learning. Collectively, these benchmarks underscore the importance of dynamic memory integration across sessions and provide comprehensive evaluations across factual recall, adaptation, and reasoning. 

From the perspective of memory utilization, we provide a detailed discussion of benchmarks that evaluate an agent’s ability to support planning and feedback using historical information.

\paragraph{Planning and Feedback.} Benchmarks targeting planning and feedback primarily assess whether agents can effectively utilize memory to support multi-step planning based on environmental feedback, and maintain coherent internal state over extended interactions. First, ALFWorld~\cite{shridhar2020alfworld} employs interactive environments to evaluate the consistency of multi-step planning, requiring agents to accumulate observations across actions and maintain latent internal states throughout execution. Moreover, formal planning benchmarks such as PlanBench~\cite{valmeekam2023planbench} and ACPBench~\cite{kokel2025acpbench} assess planning capabilities in explicitly defined dynamic environments, testing whether agents can correctly reason about action preconditions, effects, reachability, and overall plan validity. TEXT2WORLD~\cite{hu2025text2world} integrate fragmented textual descriptions into a coherent and executable world model, evaluating the capacity to continuously consolidate historical facts into structured planning representations.
More recent benchmarks place greater emphasis on feedback integration and planning under non-stationary conditions. For example, REALM-Bench~\cite{geng2025realm} introduces dynamic disturbances in real-world manufacturing scenarios, requiring agents to remember prior commitments and replan when underlying assumptions are violated, while TravelPlanner~\cite{xie2024travelplanner} focuses on accurate itinerary construction under constrained and evolving information. Finally, FlowBench~\cite{xiao2024flowbench} and UrbanPlanBench~\cite{zheng2025urbanplanbench} assess planning performance in procedural and domain-specific settings, respectively, where agents must preserve conversational or policy context and apply it consistently across decision steps. Together, these benchmarks go beyond one-shot plan generation and systematically investigate whether agents can leverage historical information to support sustained planning, adaptive feedback integration, and iterative decision revision over time.

\subsubsection{Multi-Agent System}

To evaluate coordination, competition, and decision making beyond isolated reasoning, recent benchmarks situate multi-agent systems in interactive environments. These works broadly span game-based evaluations, simulation-centric real-world scenarios, and language-driven social reasoning tasks.

\paragraph{Game-based reinforcement learning evaluation.}
Game-based reinforcement learning evaluation benchmarks leverage classical and novel gaming environments to systematically compare the performance of multi-agent RL algorithms under cooperative and adversarial settings.
MAgent \citep{zheng2018magent} facilitates massive-scale multi-agent scenarios such as pursuit and resource competition within customizable grid-worlds, evaluating individual cumulative rewards and competitive metrics like resource occupancy rates. Pommerman \citep{resnick2018pommerman} adapts the classic Bomberman game for cooperative and adversarial interactions, quantifying performance through win rates, survival duration, and kill-to-suicide ratios. SMAC \citep{samvelyan2019starcraft} centers on decentralized micromanagement challenges in StarCraft II scenarios, evaluating team success via win rates, average damage output, and formation dispersion.
MineLand \citep{yu2024mineland} utilizes Minecraft as a realistic ecological simulation for large-scale multi-agent coordination, with up to 64 agents cooperating to meet physical needs under partial observability. TeamCraft \citep{long2024teamcraft} also employs Minecraft to benchmark embodied multi-modal agents tasked with interpreting visual, textual, and environmental prompts to collaboratively achieve 55,000 procedurally generated task instances. Melting Pot \citep{leibo2021scalable} assesses agents’ zero-shot generalization capabilities in diverse social dilemma environments, utilizing metrics such as per-capita return, social welfare, and inequality indices. BenchMARL \citep{bettini2024benchmarl} provides standardized algorithm comparisons across multiple scenarios (e.g., SMACv2, VMAS, MPE), measuring convergence rates, final performance, and hyperparameter sensitivity. Finally, Arena \citep{song2020arena} encompasses a comprehensive suite of cooperative and adversarial games across various complexities, evaluating individual returns, collective social welfare, and emergent communication protocols.

\paragraph{Simulation-centric real-world assessment.}
Simulation-centric real-world benchmarks simulate realistic or pseudo-realistic environments, emphasizing scalability, partial observability, and dynamic planning. SMARTS \citep{zhou2020smarts} offers a scalable multi-agent driving platform for real-world traffic scenarios like merges and intersections, with evaluation based on collision rates, task completion, and agent behavior distributions. Nocturne \citep{vinitsky2022nocturne} provides high-throughput, partially observable driving simulations using Waymo trajectories, testing coordination and human-like behavior in tasks such as intersections and roundabouts. MABIM \citep{yang2023versatile} benchmarks multi-echelon inventory management, simulating cooperative and competitive retail dynamics, evaluated via profit metrics across diverse inventory settings. IMP-MARL \citep{leroy2023imp} addresses infrastructure inspection and maintenance scheduling, measuring risk reduction and cost efficiency in large-scale systems. POGEMA \citep{skrynnik2024pogema} focuses on decentralized multi-agent pathfinding in grids, tracking success rate, path efficiency, and large-scale coordination. INTERSECTIONZOO \citep{jayawardana2024intersectionzoo} studies contextual RL for cooperative eco-driving at intersections, using traffic simulations to evaluate emissions and travel-time performance. REALM-Bench \citep{geng2025realm} introduces real-world planning tasks from logistics to disaster relief, with dynamic disruptions, multi-threaded dependencies, and evaluation via planning quality, adaptability, and constraint satisfaction. Together, these benchmarks reflect challenges in scaling, uncertainty, coordination, and dynamic adaptation, offering rigorous testbeds for real-world multi-agent systems.

\paragraph{Language, Communication, and Social Reasoning.}
Benchmarks in Language, Communication, and Social Reasoning explore multi-agent communication protocols, Theory-of-Mind reasoning, game-theoretic interactions, and language-driven coordination. LLM-Coordination \citep{agashe2023llm} examines collaborative reasoning and joint-planning abilities of LLM agents through cooperative gameplay (e.g., Hanabi, Overcooked-AI), measured by holistic scores and fine-grained coordination question accuracy. AVALONBENCH \citep{jonathan2023avalonbench} leverages the social deduction game Avalon to assess role-conditioned language-based reasoning, with datasets of thousands of five-player dialogues and metrics on win-rate, role accuracy, and voting dynamics. Welfare Diplomacy \citep{mukobi2023welfare} extends the classic game Diplomacy to general-sum welfare negotiation, using 50-game datasets to quantify coalition stability and welfare-oriented strategic reasoning. MAgIC \citep{xu2023magic} covers social deduction and classic dilemmas (e.g., Chameleon, Prisoner's Dilemma), employing handcrafted scenario datasets to benchmark reasoning, deception, coordination, and rationality. BattleAgentBench \citep{wang2024battleagentbench} assesses language-based cooperative and competitive dynamics in strategic gameplay environments, scoring navigation accuracy, agent interactions, and exploitability across diverse map datasets. COMMA \citep{ossowski2024comma} evaluates multimodal communicative reasoning through collaborative puzzle-solving tasks involving visual-language coordination, measured by grounding accuracy, privacy compliance, and dialogue effectiveness across thousands of scenarios. IntellAgent \citep{levi2025intellagent} introduces synthetic conversational AI tasks in retail and airline domains, generating extensive policy-constrained dialogue datasets evaluated by conversational success, mistake frequency, and policy adherence. Finally, MultiAgentBench \citep{zhu2025multiagentbench} provides a comprehensive assessment across tasks such as Minecraft building, coding, and bargaining, employing dynamic key-performance indicators and LLM-scored communication quality across various multi-agent topologies and scenarios.

\subsection{Applications of Agentic Reasoning}
\label{sec:app-benchmark}

While mechanism-centric benchmarks help isolate individual capabilities, real-world deployments require these capabilities to work together under realistic constraints, such as partial observability, long-horizon dependencies, and safety-critical decisions. We therefore next review application-level benchmarks that evaluate end-to-end agent performance across representative environments, with tasks that jointly stress perception, reasoning, action execution, and coordination.

In this subsection, we review benchmarks designed to evaluate the application-level performance of agentic reasoning systems across various domains. These benchmarks assess agents' ability to perceive, reason, and act in realistic or high-impact task settings. We organize the discussion into six categories based on the application environment: \textit{Embodied Agents}, \textit{Scientific Discovery Agents}, \textit{Autonomous Research Agents}, \textit{Medical and Clinical Agents}, \textit{Web Agents}, and \textit{Tool-Use Agents}. Each subsubsection introduces representative benchmarks and describes their design motivation, task format, and evaluation metrics.

\subsubsection{Embodied Agents}
Benchmarks under this category evaluate agents that interact with physical or simulated environments, requiring grounding, perception, and action planning. AgentX~\cite{tajamul2025agentx} provides a diverse suite of vision-language embodied tasks in driving and sports, where agents must make decisions using multimodal information from videos. It emphasizes reasoning across scenes with occlusions, temporal gaps, or distractors. BALROG~\cite{davide2024balrog} builds a reinforcement learning-centric framework for benchmarking agentic planning in game environments, focusing on instruction-following, temporal abstraction, and error correction. ALFWorld~\cite{shridhar2020alfworld} links language instructions to object interactions in a text-based 3D environment, evaluating perception-grounded execution. AndroidArena~\cite{mingzhe2024understanding} targets GUI-based mobile tasks, where agents must perform actions like form-filling and app navigation using vision-language understanding. StarDojo~\cite{weihao2025stardojo} leverages the open-ended Stardew Valley game to study social planning and role-based coordination. MindAgent~\cite{ran2023mindagent} and NetPlay~\cite{dominik2024playing} create multiplayer gaming testbeds to benchmark emergent social reasoning and negotiation under uncertainty. OSWorld~\cite{tianbao2024osworld} offers a simulated desktop environment with diverse cross-app productivity tasks, such as opening files, converting formats, and modifying documents. These environments challenge agents to coordinate between perception, planning, and symbolic action in dynamic and often partially observable scenarios.

\subsubsection{Scientific Discovery Agents}
Scientific benchmarks aim to test agents' capabilities in knowledge acquisition, hypothesis generation, and experimental automation. DISCOVERYWORLD~\cite{jansen2024discoveryworld} introduces a virtual lab where agents explore scientific phenomena in biology, chemistry, and physics through simulated tools and instruments. ScienceWorld~\cite{ruoyao2022scienceworld} focuses on elementary science experiments using textual instructions and environment interactions, requiring step-by-step hypothesis testing. ScienceAgentBench~\cite{ziru2024scienceagentbench} builds a benchmark from real-world scientific papers, translating tasks like code implementation, figure generation, and variable extraction into executable subtasks, assessing agents’ ability to automate the research process. The AI Scientist~\cite{chris2024the} simulates a full end-to-end research pipeline, where agents perform literature review, method writing, experiment execution, and peer-review simulation. LAB-Bench~\cite{jon2024labbench} evaluates biology-specific agents on tasks involving genetic sequence reasoning and experiment planning. MLAgentBench~\cite{qian2023mlagentbench} benchmarks agents’ ability to autonomously train, evaluate, and tune machine learning models, offering realistic experimentation workflows. These benchmarks collectively probe open-ended reasoning, long-horizon planning, and scientific grounding in semi-structured data settings.

\subsubsection{Autonomous Research Agents}
This category benchmarks agents designed for long-horizon workflows across general-purpose research, office, or planning tasks. WorkArena~\cite{alexandre2024workarena} and its extension WorkArena++~\cite{leo2024workarena} propose enterprise task benchmarks where agents must complete ticket-based workflows involving retrieval, summarization, and coordination across documents. OfficeBench~\cite{zilong2024officebench} simulates a productivity software suite environment with tasks such as creating meeting memos, modifying spreadsheets, and replying to emails, emphasizing goal decomposition and tool selection. PlanBench~\cite{valmeekam2023planbench} and FlowBench~\cite{xiao2024flowbench} test general workflow planning skills with abstracted task graphs and structured dependencies. ACPBench~\cite{kokel2025acpbench} evaluates agents in assistant–collaborator–planner triads, tracking performance in a hybrid role hierarchy. TRAIL~\cite{darshan2025trail} focuses on multi-agent trace debugging and error attribution~\cite{zhang2024knowledge} in LLM-based systems, providing dense annotations for reasoning chains. CLIN~\cite{bodhisattwa2023clin} introduces lifelong few-shot learning benchmarks where agents adapt to distribution shift and task evolution. Agent-as-a-Judge~\cite{mingchen2024agentasajudge} studies peer-review style evaluation with agents grading reasoning chains and correctness of other agents’ outputs. InfoDeepSeek~\cite{xi2025infodeepseek} measures information-seeking abilities in open-domain QA and synthesis tasks. Together, these benchmarks capture the growing demand for agentic reasoning in complex knowledge workflows that involve abstraction, iteration, and evaluation.

\subsubsection{Medical and Clinical Agents}
These benchmarks test agents’ abilities to reason with clinical knowledge, patient data, and multimodal biomedical sources. AgentClinic~\cite{samuel2024agentclinic} introduces a virtual hospital environment where agents make diagnostic decisions based on patient symptoms and medical imaging. MedAgentBench~\cite{yixing2025medagentbench} combines medical QA, patient simulation, and retrieval tasks in a multi-format benchmark grounded in standardized exams. MedAgentsBench~\cite{xiangru2025medagentsbench} evaluates multi-hop medical reasoning over structured and unstructured data, scoring agents on correctness and evidence alignment. EHRAgent~\cite{wenqi2024ehragent} benchmarks agents working over structured electronic health record (EHR) tables and clinical notes to complete tasks like diagnosis code prediction and medication reasoning. MedBrowseComp~\cite{chen2025medbrowsecomp} focuses on browsing-based medical QA, where agents must retrieve and verify information across web pages. ACC~\cite{thakrar2025architectingclinicalcollaborationmultiagent} explores trustworthy medical agents with retrieval, hallucination detection, and citation-based support evaluation. MedAgents~\cite{xiangru2023medagents} uses a collaborative multi-agent dialogue setup to simulate patient–doctor–nurse interactions, scoring fluency and factual accuracy. GuardAgent~\cite{zhen2024guardagent} proposes a clinical privacy safeguard agent with structured risk detection benchmarks on EHR and website forms. These datasets emphasize correctness, trustworthiness, and safety in real-world clinical deployment contexts.

\subsubsection{Web Agents}
Web agents operate in realistic browsing environments and are benchmarked on their ability to parse layouts, execute actions, and handle dynamic content. WebArena~\cite{shuyan2023webarena} introduces a browser-based benchmark suite containing 90+ realistic websites across domains like shopping and booking, where agents complete tasks with structured goals and click-based APIs. VisualWebArena~\cite{jing2024visualwebarena} extends this with visual rendering, requiring agents to parse webpage images and align instructions with rendered components. WebVoyager~\cite{he2024webvoyager} proposes goal-driven navigation with long-horizon tasks involving multi-page traversal and backtracking. Mind2Web~\cite{gou2025mind2web2} targets cross-domain web automation with multi-task datasets and rich grounding annotations. WebCanvas~\cite{yichen2024webcanvas} supports fine-grained layout manipulation, such as drag-drop and resize actions. WebLINX~\cite{xing2024weblinx} simulates information gathering tasks with browsing, summarization, and answer synthesis. BrowseComp-ZH~\cite{peilin2025browsecompzh} brings language and infrastructure diversity with Chinese websites, challenging agents on multilingual understanding. LASER~\cite{kaixin2023laser}, WebWalker~\cite{wu2025webwalker}, and AutoWebBench~\cite{lai2024autowebglm} focus on structured page representation, real-time action execution, and policy learning in web navigation. These benchmarks highlight perception, grounding, and policy generalization challenges in web settings.

\subsubsection{General Tool-Use Agents}

This group of benchmarks emphasizes LLM agents' ability to invoke, coordinate, and reason over tools and APIs. GTA~\cite{jize2024gta} presents a realistic tool-use benchmark grounded in user queries and deployed software tools, spanning APIs from image generation to analytics dashboards. NESTFUL~\cite{kinjal2024nestful} evaluates nested API invocation tasks requiring compositional planning across toolchains. CodeAct~\cite{wang2024executable} simulates executable function calling and evaluates agents on parsing, composition, and runtime accuracy. RestGPT~\cite{song2023restgpt} connects LLMs with RESTful APIs via coarse-to-fine planning pipelines, tested on 60+ tool types. Search-o1~\cite{li2025search} frames tool use as sequential retrieval, with benchmarks spanning code search, PDF querying, and scientific tool usage. Agentic RL~\cite{joykirat2025agentic} proposes a reinforcement learning agent with access to tool interfaces and evaluation tasks such as calendar scheduling and translation. ActionReasoningBench~\cite{divij2024actionreasoningbench} benchmarks agents’ ability to reason about action side effects and downstream consequences using a structured action grammar. R-Judge~\cite{tongxin2024rjudge} introduces safety judgment benchmarks where agents assess risky plans involving tools. These datasets jointly reflect the increasing complexity and compositionality of tool-augmented agent environments.

\section{Open Problems}
\label{sec:openproblems}

In this section, we highlight open problems arising from user-centric personalization, long-horizon interaction and credit assignment, world-model-based reasoning, multi-agent collaboration and training, latent internal reasoning, and the governance of agentic systems operating autonomously in real-world environments.

\subsection{User-centric Agentic Reasoning and Personalization}
User-centric agentic reasoning \cite{qian2025userrl,li2025hello} refers to an agent’s ability to tailor its reasoning and actions to a specific individual user by modeling user characteristics, preferences, and interaction history over time. Rather than optimizing a fixed, task-defined objective, a user-centric agent treats the user as part of the environment and continuously adapts its strategy through extended, multi-turn interaction. This requires the agent to dynamically infer evolving user intent, accommodate changes in goals and behavior styles, and adjust decisions based on explicit or implicit user feedback as the dialogue progresses. Crucially, user-centric agentic reasoning involves balancing short-term task rewards with long-term user experience, satisfaction, and trust, which introduces non-stationary objectives and long-horizon credit assignment challenges beyond conventional agentic reasoning settings.

\subsection{Long-horizon Agentic Reasoning from Extended Interaction}
A central open challenge in agentic reasoning is robust long-horizon planning and credit assignment across extended interactions. While methods such as ReAct and Tree of Thought improve short-horizon reasoning \citep{yao2023react,yao2023tree}, errors still compound rapidly in long tasks, as illustrated by embodied agents like Voyager \citep{wang2023voyager}. RL-trained agents such as WebRL and Agent-R1 improve performance in realistic environments but rely on heavily engineered, domain-specific rewards and largely treat episodes independently \citep{qi2024webrl,wei2025webagent}. More recent process-aware approaches attempt to construct finer-grained credit signals \citep{xiang2024retrospex,yan2025memoryr1,ji2025tree}, yet remain environment-specific. A core open problem is how to assign credit across tokens, tool calls, skills, and memory updates, and to generalize such learning across a long sequence of episodes and tasks.

\subsection{Agentic Reasoning with World Models}
World-model-based agents \cite{carbonneaux2025cwm, zhang2025agent2} aim to mitigate myopic reasoning by enabling internal simulation and lookahead. Model-based RL systems such as DreamerV3 demonstrate the effectiveness of imagined rollouts for long-horizon control \citep{hafner2023mastering}, while recent LLM-based agents adapt world models to web, code, and GUI environments \citep{tang2024worldcoder,carbonneaux2025cwm,chae2024web,luo2025vimo}. However, current designs rely on ad hoc representations and are typically trained on short-horizon or environment-specific data, raising concerns about calibration and generalization. Only a few works explore co-evolving world models and agents over long time scales \citep{fang2025webevolver,deng2025simura}. An open problem is how to jointly train, update, and evaluate world models in non-stationary environments, and how to assess their causal impact on downstream planning reliability.

\subsection{Multi-agent Collaborative Reasoning and Training}
Multi-agent collaboration has emerged as a powerful paradigm for scaling agentic reasoning through role specialization and division of labor \citep{li2023camel,chen2023agentverse,wu2024autogen}. While debate- and role-based systems often outperform single agents, most collaboration structures are still manually designed. Recent multi-agent RL approaches begin to treat collaboration itself as a trainable skill \citep{liu2025llm,park2025maporl,ma2024coevolving}, but credit assignment at the group level remains poorly understood. Scaling to larger agent populations further introduces challenges in topology adaptation, coordination overhead, and safety \citep{qian2024scaling,florian2025agentsnet, zhang2024knowledge}. A key open problem is how to learn adaptive, interpretable collaboration policies that remain robust under partial observability and adversarial conditions.

\subsection{Latent Agentic Reasoning}
Latent agentic reasoning \cite{du2025enabling,fu2025cache,latentmas} explores performing planning, decision-making and collaboration in internal latent spaces rather than explicit natural language or symbolic traces. Recent work suggests that latent reasoning can improve efficiency and scalability, but at the cost of reduced interpretability and controllability. In agentic settings, this raises additional challenges, including how to align latent reasoning with external objectives, tools, agents and memory systems. Diagnosing failures becomes particularly difficult when intermediate reasoning steps are not externally observable. An open problem is how to design learning objectives, probing methods, and evaluation benchmarks that make latent agentic reasoning both effective and auditable.

\subsection{Governance of Agentic Reasoning}
Governance is a cross-cutting challenge for agentic reasoning systems that act autonomously over tools, environments, and other agents. Beyond standard LLM safety issues, agentic systems introduce new risks due to long-horizon planning, persistent memory, and real-world action execution \citep{wang2025comprehensive}. Failures may arise from interactions across time and components, making attribution and auditing difficult. Existing benchmarks and guardrails mainly focus on short-horizon behaviors \citep{zhen2024guardagent,tongxin2024rjudge}, leaving planning-time failures and multi-agent dynamics underexplored. A central open problem is to develop governance frameworks that jointly address model-level alignment, agent-level policies, and ecosystem-level interactions under realistic deployment conditions.

\bibliographystyle{unsrtnat}
\bibliography{reference}

\end{document}